# When the Same Prompt Gives Different Answers (and the Same Answer Isn't Always Right): A Taxonomy of Failure Mechanisms, Testing, and Mitigation Strategies for LLM-Assisted Evidence Synthesis Read-and-Evaluate Tasks

Carole Lunny, Ian Goodman, Stefano Berrone, Connor T. A. Brenna, Emma K Reid, Iván Pérez-Neri, Shailesh Kolekar, Nityanand Jain, Lucas HCC Santos, Yulia Otmakhova, Ashwin Kamath, Heba Hussein, Diana M. Céspedes, Gioia Dahdal, Salmaan Kanji

Corresponding author: Carole Lunny. Cochrane Hypertension Review Group, University of British Columbia. carole.lunny@ubc.ca. ORCID 0000-0002-7825-6765.

LLM read-and-evaluate tasks include study screening, data extraction, risk-of-bias assessment, and critical appraisal in the evidence synthesis field. These task differ in the amount of LLM interpretation they require. Some tasks involve retrieving relatively simple information, such as an author name, sample size, or publication year. Others require the LLM to read source material, interpret it against predefined criteria, and make a judgement. Researchers should assess whether LLM judgements are accurate, supported by appropriate source evidence, and reproducible under specified conditions. A correct judgement does not necessarily indicate that the underlying evaluation process is reliable. We developed a taxonomy for identifying mechanisms that may compromise these properties and for informing the selection and reporting of appropriate evaluation methods. The taxonomy outlines potential ways that LLM judgments from read-and-evaluate tasks can fail, called TRAILS (Taxonomy of Reliability And Instability in LLM-assisted evidence Synthesis). The TRAILS taxonomy addresses two complementary questions. First, what mechanisms can cause the same read-and-evaluate task to produce different judgements across repeated LLM runs? Second, what mechanisms can systematically affect the judgement or required output in a way that is not apparent from the judgement itself? This study presents 38 mechanisms and identifies the deployment settings in which each can operate, approaches for testing them, and mitigation strategies. We also show why repeatability or accuracy does not rule out a hidden judgement failure and why human evaluation should assess whether a judgement is grounded in the supplied evidence, independent of inappropriate external influence, and produced through a process that can be examined and verified. The TRAILS taxonomy is primarily intended for researchers, methodologists, and developers who use or evaluate LLMs for evidence synthesis read-and-evaluate tasks. It can help users identify potential mechanisms of repeat-run variability or hidden judgement failure, select appropriate tests, and mitigation strategies, and interpret accuracy and repeatability results in relation to the conditions under which the LLM was evaluated.

**Summary points**

- Large language models (LLMs) are increasingly used for evidence synthesis tasks, but LLM read-and-evaluate studies usually emphasise accuracy or agreement from a single prompt run.
- LLM judgements can vary across repeated runs, prompt formulations, model versions, and technical settings. A prompt is the instruction and other input supplied to the model.
- TRAILS (Taxonomy of Reliability And Instability in LLM-assisted evidence Synthesis) uses two prespecified eligibility criteria to distinguish repeat-run variability from hidden judgement failure.
- We identified 38 mechanisms organised into ten categories and mapped each mechanism to the deployment settings in which it can operate, a method for testing it, and mitigation strategies.
- We demonstrate why repeatability does not rule out hidden judgement failure and why researchers should assess whether the judgement is grounded in the supplied evidence, independent of inappropriate external influence, and produced using a transparent process that can be examined and verified.

- Researchers should first identify and test relevant mechanisms and apply needed mitigations. LLM judgements can then be compared with the human reference judgements for that evaluation dataset to estimate accuracy.
- High accuracy is not sufficient evidence that an LLM can reliably perform a read-and-evaluate task, because an LLM's judgement may show repeat-run variability or a hidden judgement failure.

## INTRODUCTION

Large language models (LLMs) such as Chat-GPT, Claude, and Gemini are large-scale neural language models that generate natural language judgements, summaries, and evaluations from user- provided prompts. These technologies have seen rapid adoption as assistant tools across different research workflows given their ability to rapidly process information at scale. Large language models (LLMs) are increasingly being used across an evidence synthesis process, including study screening, data extraction, and assessment of study quality or bias.[1] These specific LLM tasks differ in the amount of interpretation they require. Some tasks involve retrieving relatively simple information, such as an author name or publication year, while others require the model to read source material, interpret it against predefined criteria, and make a substantive judgment. We refer to the latter as read-and-evaluate tasks.[2, 3] Examples include deciding whether an article meets a systematic review's inclusion criteria, determining whether participant blinding was adequate when assessing risk of bias of a randomised control trial, judging whether review authors used appropriate methods to investigate heterogeneity, or deciding whether a systematic review addresses an AMSTAR 2[4] or ROBIS[5] criterion. These tasks overlap with the broader category of LLM-as-a-judge tasks, in which LLMs evaluate information, responses, or other objects using specified criteria (with or without requiring source evidence).[2, 3] Thus, we define read-and-evaluate tasks as an evidence-focused subset of LLM-as-a-judge tasks. Read-and-evaluate tasks create particular reliability concerns because the task requires the model to interpret information and apply evaluation criteria to reach a judgement, rather than simply retrieve information.[1] In this paper, "judgement" refers to the substantive decision or rating produced by the LLM, while "output" refers to the full model response, which may also include supporting evidence, quotations, rationale, references, explanations, or other required information. **Box 1** defines key LLM terminology used throughout this paper. Additional technical terms, deployment settings, and workflow architectures are defined in **Supplementary File 1**.

**Box 1. Key terminology used in LLM-assisted read-and-evaluate tasks**

**Application programming interface (API).** A way for researchers to send requests to a hosted LLM through software rather than a chat interface, with control over some supported settings.

**Agentic workflow.** A workflow in which an LLM can select actions, use tools, or carry out dependent steps to complete a task.

**Large language model (LLM).** An artificial intelligence model trained on large amounts of data, including text, that can interpret instructions and generate language-based outputs.

**Prompt.** The text or other instructions presented to the model to specify or guide how a task should be performed.

**Read-and-evaluate task.** A task in which the model interprets source information and applies specified criteria to make a judgement, classification, decision, or informed extraction.

**Judgement.** The substantive decision, classification, rating, or other task-level determination produced by the LLM.

**Output.** The full response returned by the model, which may include the judgement, quotations, rationale, references, explanations, and other required information.

**Supporting output.** Material accompanying or substantiating the judgement, such as quotations, evidence locations, references, or explanations.

**Specified conditions.** The source evidence, task, prompt and instructions, specified model variant and version, and user-accessible model settings held the same for a repeat-run comparison.

**Material difference.** A difference affecting interpretation, fulfilment of task requirements, or downstream use, rather than wording alone.

**Accuracy.** Agreement with a prespecified reference judgement for a specified evaluation dataset.

**Repeatability.** Agreement between repeated judgements under the same specified conditions.

**Reproducibility.** Agreement when an independent analyst repeats the evaluation using the same data, methods, and specified conditions.

**Judgement validity.** Whether the judgement was determined by information that should legitimately determine it.

**Robustness or sensitivity testing.** Deliberately changing a specified condition and examining whether the judgement or required output changes. This does not itself demonstrate IC1 or IC2.

**Mechanism.** A process or condition that can contribute to a specified failure.

**Failure mode.** An operationally defined pattern in which a problem appears; its underlying computational cause may not be established.

**Retrieval-supported workflow.** A workflow that selects relevant documents or passages and supplies them to the model to inform its judgement.

Conflating judgement accuracy with repeatability can lead to overconfident claims about LLM performance in evidence synthesis, and neither can detect hidden LLM failures. Accuracy measures agreement with a prespecified reference standard, whereas repeatability measures agreement across repeated runs under the same specified conditions. A model may produce different judgements across runs, some of which match the reference standard, or repeatedly produce the same incorrect judgement.[6-9] For example, Lai and colleagues assessed ChatGPT and Claude on risk-of-bias assessments of 30 randomised controlled trials, repeating each assessment twice. Although repeat-run agreement was 84% and 87%, respectively, 57% of incorrect judgments had correct supporting rationales, including cases where models repeated an incorrect judgement despite identifying the relevant evidence. [7] Hidden judgement failures may also occur when judgements match the reference standard, because the model may have relied on information that should not have influenced its decision. For example, Han and colleagues found that instead of following instructions and using only the provided evidence, LLM agents used the web to retrieve reference answers approximately 3% of the time, reducing accuracy by approximately 15% among those questions, suggesting that web access to reference answers had inflated measured performance.[10] These findings highlight the need to evaluate accuracy and repeatability separately, alongside hidden judgement failures, in which an LLM mechanism or failure mode affects the judgement or required output in a way that is not apparent from the judgement itself.

We recognized a need for a coherent framework for identifying and organising the LLM mechanisms and failure modes that can contribute to repeat-run variability and hidden judgment failures for read-and evaluate tasks. Although previous reviews and taxonomies have classified LLM evaluation problems according to different dimensions, including judgement biases, evaluation methods, and system-level failure modes, they use different terminology and do not organise mechanisms according to whether they can produce repeat-run variability, hidden judgement failures, or both.[2, 3, 11-13] Moreover, different failure mechanisms produce similar observed errors, which may require different testing methods and mitigation strategies. For example, an incorrect eligibility decision could arise because relevant evidence was omitted during document preparation, the LLM overlooked an exclusion criterion, or the criteria were misinterpreted. A mechanism-based taxonomy could provide a common terminology, help researchers identify which mechanisms are relevant to their study or evaluation design, and guide the selection of appropriate assessments and mitigations. It could also support help developers, reviewers, and evaluators identify potential hidden judgement failures that accuracy and repeatability may not detect, and promote transparent reporting of which mechanisms were applicable to the task and should have been investigated.

The objective of this paper is to describe the development, and provide a brief overview of, the taxonomy of repeat-run variability and hidden judgement failure mechanisms within LLM-assisted read-and-evaluate evidence synthesis tasks. The paper provides a description of the 38 mechanisms, their supporting evidence, proposed tests, and mitigation strategies, and accompanying user tools and guidance. TRAILS (Taxonomy of Reliability and Instability in LLM-assisted evidence Synthesis) focuses on read-and-evaluate tasks in which an LLM must interpret source material and apply criteria, rules, or substantive reasoning to reach a judgment. It addresses two questions. First, what mechanisms can cause the same read-and-evaluate task to produce different judgments across repeated runs? Second, what mechanisms can systematically affect the judgement or required output in a way that is not apparent from the judgement itself?

This is the second paper in a four-paper series on evaluating LLM-assisted read-and-evaluate tasks in evidence synthesis. The first paper is a tutorial that explains the core properties of LLM judgements and supporting outputs that should be evaluated, including accuracy, repeatability, hidden judgment failures, and judgement validity. This second paper introduces TRAILS (Taxonomy of Reliability And Instability in LLM-assisted evidence Synthesis), a taxonomy of mechanisms and failure modes that can produce repeat-run variability or hidden judgement failures and links those mechanisms to testing and mitigations. The third paper maps the evidence synthesis task, deployment settings and workflows, and technical features to the mechanisms that apply. The fourth paper provides a study-design framework for evaluation of LLM read-and-evaluate studies. It will specify how to measure accuracy, repeatability, sensitivity analyses, evidence verification, and occurrence and magnitude of mechanism effects.

**Scope of the taxonomy**

The taxonomy was limited to mechanisms relevant to LLM read-and-evaluate tasks. A read-and-evaluate task was defined as a task in which the LLM is required to interpret information and apply evaluation criteria to reach a judgement, classification, decision, or informed extraction, rather than simply retrieve information. The definition concerns what the model is asked to do, not how the task is executed. The same read-and-evaluate task may be performed in a single model call or within a retrieval-supported, tool-using, agentic, or multi-agent workflow. The scope was intentionally narrow because the aim was to identify mechanisms that affect tasks requiring interpretation and judgement. Simple factual extraction, such as identifying an author name or publication year, usually has one clear value and requires little interpretation. In contrast, evidence synthesis tasks such as risk of bias assessment or interpretation of a figure require the LLM to identify relevant information and apply criteria to that information. These tasks provide more opportunity for LLM judgements to vary across repeated runs or for a problem in the judgement or required output to remain unapparent from the judgement itself.

Mechanisms were included when they met inclusion criterion 1 (IC1), inclusion criterion 2 (IC2), or both (Figure 1):

- IC1 repeat-run variability: The mechanism can produce materially different judgements, extractions, or required supporting outputs across repeated runs of the same read-and-evaluate task under the same specified conditions. Specified conditions include the source evidence, task, prompt and instructions, specified model variant and version, and model settings.
- IC2 hidden judgement failure: The mechanism can systematically affect the judgement or required output in a way that is not apparent from the judgement itself. Repeatability and accuracy are not part of this definition and do not rule out IC2.
- Both: A mechanism meets both criteria when it can cause repeat-run variability and can also cause an IC2 problem.

A material/substantive difference was defined as a change in the judgement, classification, decision, required task completion, or downstream use, rather than wording or formatting alone.

**Figure 1. Defining and investigating repeat-run variability (IC1) and hidden judgement failure (IC2) in LLM read-and-evaluate tasks**

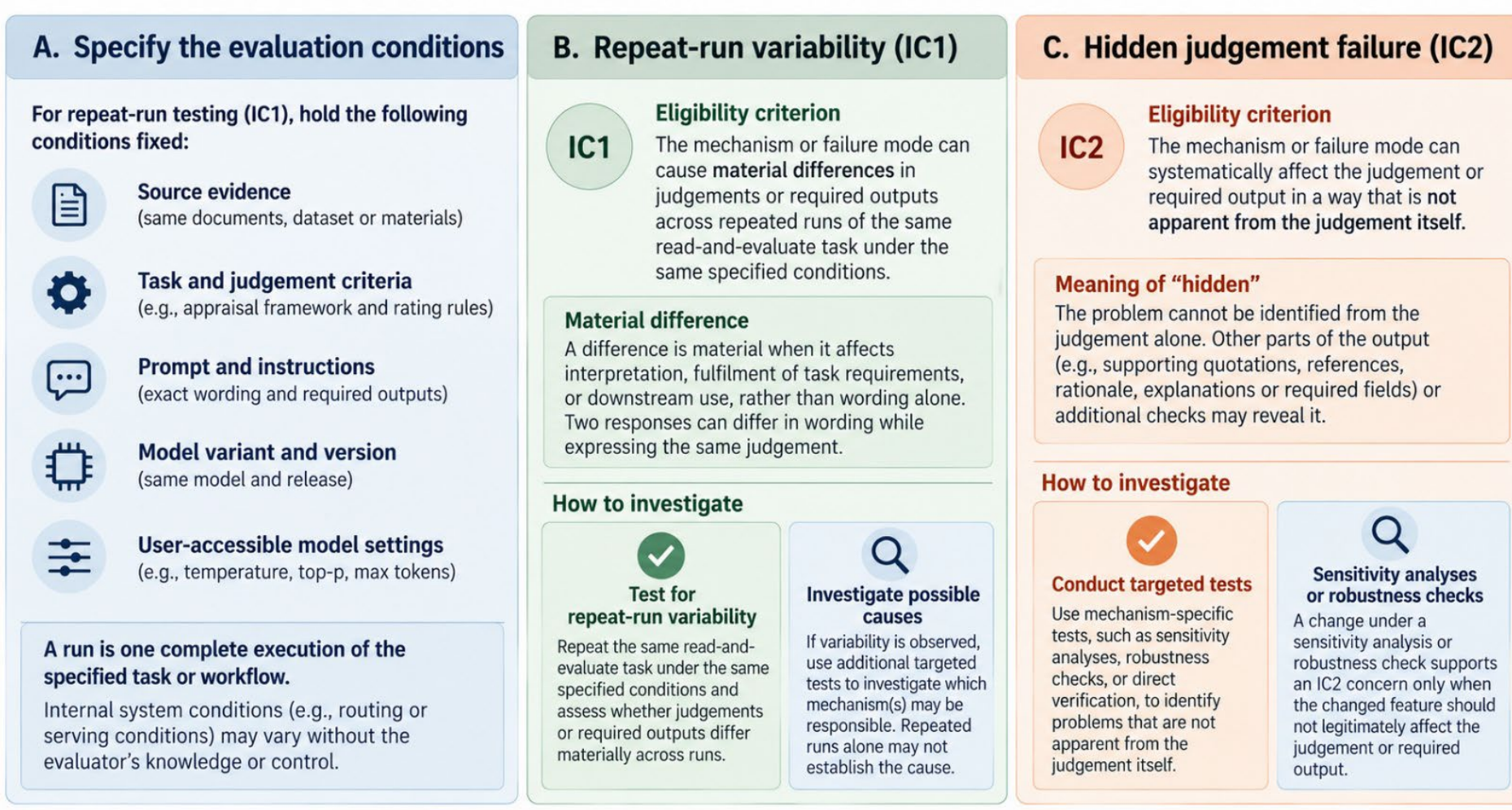

**Note:** IC1, repeat-run variability: Materially different judgements, extractions, or required supporting outputs can occur across repeated runs under the same specified conditions. IC2, hidden judgement failure: A mechanism can systematically affect the judgement or required output in a way that is not apparent from the judgement itself.

Each included mechanism or operationally defined failure mode met three requirements. First, published research, vendor documentation, or other credible technical evidence documented the relevant failure, behaviour, or process. Second, this evidence supported a plausible pathway to repeat-run variability (IC1), hidden judgement failure (IC2), or both. Third, the proposed pathway had to be testable in practice, even if it had not yet been tested in published studies. Third, inclusion did not require empirical evidence that a mechanism had caused repeat-run variability or a hidden judgement failure. Instead, each mechanism or operationally defined failure mode required a testable pathway to IC1, IC2, or both. The taxonomy therefore identifies potential sources of these failures and provides a basis for targeted evaluation, rather than establishing that the mechanisms caused particular observed failures.

We excluded mechanisms that were outside the applicable read-and-evaluate task or workflow (e.g., function-calling failures when no external tools were used), lacked a testable pathway to IC1 or IC2 (e.g., verbatim memorisation that did not affect the judgement or required output), or described broad performance outcomes without a distinct mechanism or operationally defined failure mode (e.g., aggregate calibration failure). We also excluded research and reporting practices that could affect the interpretation or reproducibility of an evaluation study but did not themselves cause IC1 or IC2 (e.g., incomplete reporting of model settings or omission of evaluation results). Failures that only prevented the workflow from starting, without affecting an attempted evaluation, were outside the scope of the taxonomy. Exclusion decisions and search, extraction, and synthesis methods are reported in **Supplement 2**. We merged mechanisms describing essentially the same failure when they lacked distinct definitions.

We coded deployment setting and workflow architecture separately because they determine which mechanisms can operate and what the researcher can observe or control. Deployment referred to how the model was accessed, and workflow described how model calls, retrieval, tools, and processing steps were arranged. We distinguished four deployment settings: a managed chat or application interface (e.g., ChatGPT, Claude.ai); a vendor-managed agentic environment (e.g., Claude Code, Codex CLI); a direct model API; and a user-controlled model runtime using locally run open weights and inference software. Control over instructions, routing, context, settings, and hardware differs across these settings. Self-hosting gives more control over execution but does not necessarily reveal a model's training or development history. We distinguished four workflow types including single-call (i.e., the LLM completes the task in one model call), retrieval-supported (i.e., relevant evidence is retrieved and provided to the LLM to inform its judgement), agentic (i.e., an LLM system plans and executes multiple steps, potentially using tools and checking intermediate results), and multi-agent (i.e., two or more LLM agents perform separate or coordinated tasks to produce a judgement) (a tutorial on deployment settings and workflow is found in **Supplementary File 3**). These workflow types are not mutually exclusive and determine which failure mechanisms may operate.

The ten categories (A–J) grouped mechanisms by their defining features. We also coded each mechanism according to five aspects of the read-and-evaluate task that it could affect including framing (presentation of task and criteria), evidence basis (available and used evidence), judgement formation (application of evidence and criteria), expression (recording of judgements and required outputs), and substantiation (supporting quotations, references, and explanations). A mechanism could affect several aspects.

## OVERVIEW OF THE TAXONOMY OF MECHANISMS

The taxonomy contains 38 mechanisms across ten categories (**Table 1**). Four mechanisms met IC1 only, 20 met IC2 only, and 14 met both (sub-mechanisms were not included in these totals). **Table 1** outlines the mechanisms in plain language and indicates under which deployment setting they can operate and whether the researcher can directly control or test it. These deployment headings indicate the level of access generally needed to investigate a mechanism, not where the mechanism can occur. The final column describes what researchers can observe, test, or control within each deployment setting. Detecting an effect does not necessarily establish its underlying cause.

The ability to investigate a mechanism depends on the deployment setting and the controls available to the researcher. Some mechanisms, including training contamination (D.1), hidden model switching (B.1), numerical variation during model serving (B.3), and unfaithful reasoning (H.1), are particularly difficult to verify directly because the relevant training data, internal routing, numerical processes, or model reasoning are generally inaccessible. In contrast, vendor-managed agentic environments may provide greater access than commercial chat interfaces to document processing (E.1, E.2), retrieval (E.3), agent execution and verification steps (G.1, G.2), and context management (J.1–J.3). This additional access may allow researchers to examine intermediate steps and investigate potential causes of judgement failures, although the availability of these controls varies across systems and does not guarantee that a mechanism can be identified or prevented.

**Supplement 4, Table S1** groups all mechanisms into ten categories based on related features of the LLM system, inputs, evaluation task, and workflow: two in Category A (User-Level Instability), three in Category B (Infrastructure Instability), one in Category C (Sampling Instability), four in Category D (Contamination), seven in Category E (Input Processing and Context Effects), five in Category F (Prompt & Task Design Instability), two in Category G (Workflow Instability), nine in Category H (Judgement, Evidence Use, and Reporting Failures), two in Category I (Source Presentation Effects), and three in Category J (Context Management). Letters A–J identify the ten mechanism categories, and the number identifies the individual mechanism within that category (for example, A.1 or H.4). **Supplement 4, Table S1** also presents the strength of the evidence for each mechanism, their testing and mitigation strategies.

We developed a simplified checklist to help researchers apply the taxonomy to LLM read-and-evaluate studies (**Supplement 5**). The checklist lists all 38 parent mechanisms, grouped by deployment setting, with two H.7 sub-mechanisms shown separately. For each mechanism, researchers can record whether it was tested, whether a required feature was disabled, whether the setting, procedure, results, and limitations were reported, and whether a mitigation was implemented. The checklist is intended as a practical reference for identifying and documenting relevant mechanisms, rather than a comprehensive evaluation or reporting tool. It has not been formally validated.

Some mechanisms in this taxonomy are *conditional mechanisms*. These mechanisms are only applicable when a particular system feature or capability is enabled or used, either by the model provider or the researcher (e.g., web access or retrieval). An observed judgement failure does not establish which mechanism caused it, because several mechanisms may produce similar failures. Where possible, researchers should confirm that the relevant condition occurred and assess whether it contributed to the failure, using system logs, API outputs, or controlled experiments.

**Table 1: Taxonomy of mechanisms affecting repeatability and hidden judgement failure in LLM-assisted read-and-evaluate tasks**

| Mechanism (code) | Description | IC1/IC2 | Can you do anything about it (by deployment setting)? |
|---|---|---|---|
| **All deployments** | | | |
| 1. Past chats or memory (A.1) | Saved memory, prior conversation history or user profiles can alter what the model uses beyond the supplied evidence. | Both | Controllable/testable in any deployment that carries prior user, session, or memory state into the task: Repeat the task in a new session with no prior history and with memory or personalisation disabled where possible and compare the judgement. Avoid supplying prior history or memory unless it is needed. |
| 2. Hidden instructions (A.2) | Persistent account-level instructions outside the prompt can add assumptions, priorities, or decision rules. IC2 applies only when those instructions change the substantive evaluation rather than merely tone or formatting. | Both | Controllable/testable: Disable user-controlled custom instructions for evaluation sessions or standardise them explicitly and report their content/state. |
| 3. The model looked it up (D.2) | The model may use information from outside the supplied material, e.g., benchmark judgements, earlier appraisals, or other online sources. This can make its performance look better than it is on the intended task. | IC2 | Controllable/testable in all deployments: Check and record accessible retrieved sources. Disable web access where possible when the model should judge only the supplied material, or restrict retrieval to approved sources. |
| 4. The expected judgement | The expected judgement is included in the information given to the model (e.g., a PDF | IC2 | Partly controllable/testable in managed interfaces, vendor-managed agentic or work environments, and APIs, |
| travelled with the question (D.3) | contains invisible white text that reveals the expected judgement). The model may then rely on that cue instead of doing the task independently. The cue may appear in the prompt or instructions, filenames or metadata, or in visible or hidden text within a document. | | and more directly controllable in user-controlled model runtimes: Remove known reference judgements from prompts and files before uploading them, and check documents for hidden text where possible. Hidden system instructions or platform processing may not be visible to the user. In self-hosted: Inspect and clean all model inputs and document-processing steps, including hidden text and metadata, before evaluation. |
| 5. Hidden instructions enter with the evidence (D.4) | A document, image, Unicode, metadata, or other source may contain hidden or unintended instructions. The LLM may follow these instructions instead of only using the source as evidence. | IC2 | Controllable/testable: Check whether the evidence contains instructions humans cannot see, using a hidden-text detector where appropriate. Remove detected instructions before the evidence reaches the LLM. Clearly separate task instructions from document content and tell the model not to follow instructions found inside the document. |
| 6. Prompt sensitivity (F.1) | Prompts that mean the same thing can still produce different judgments when wording or formatting changes. | IC2 | Controllable/testable: Test meaning-preserving prompt versions on a development set when available. Prespecify and version the final prompt before the main evaluation, and test reasonable equivalent prompt variants as sensitivity analyses. |
| 7. Writing and format effects (I.1) | The model may give different ratings when the same information is written in a different style or format. | IC2 | Controllable/testable: Compare matched versions that contain the same information but use different writing styles or formats. Use clear criteria that focus the judgement on the methods or evidence. |
| 8. Who produced the material (I.2) | Information about who or what produced the material can influence the model's judgement even when the content is unchanged. This includes labels e.g., Human, AI, Expert, an author, or an institution. | IC2 | Controllable/testable: Remove source, author, model, or institutional labels when they are not needed for the task. If labels are required, keep the content unchanged and vary or remove the label to test whether the judgement changes. |
| 9. Role playing (F.5) | When the evidence and task stay the same, giving the model a role or persona, e.g., "You are a clinical expert," changes the model's judgement even when the evidence and task are unchanged. | IC2 | Controllable/testable: Avoid persona prompts unless they are needed for the task. If a persona is introduced as a task-specific intervention, compare it with a no-persona condition and reasonable alternatives before fixing it for the main evaluation. Do not treat persona prompting as a general mitigation without evidence that it improves performance for the specific task. |
| 10. Worked examples can steer the judgment (F.3) | When worked examples are included, the judgment can depend on which examples are used and the order in which they appear. In general classification tasks, models sometimes favoured labels that appeared more often in the worked examples or appeared in examples near the end of the prompt. Reordering examples changes the declared prompt condition and therefore is not IC1. | IC2 | Controllable/testable: Where feasible, use development data to select and refine worked examples, then prespecify and version the exact examples and their order. before the main evaluation Test other reasonable example sets and orders as sensitivity analyses. |
| 11. User influence on the judgement (H.6) | The model may change its judgement to agree with what the user thinks or wants the judgement to be, even when the source evidence does not support that position. This does not include appropriate changes when the user provides new relevant evidence. | IC2 | Controllable/testable: Run the same task with no stated user opinion, then with the user suggesting the correct judgement and an incorrect judgement. Do not tell the model the user's preferred conclusion unless it is needed for the task. Ask the model to identify and assess the source evidence before making the judgement, and require any later change in judgement to be supported by evidence from the source. |
| 12. Order of the options (F.7) | The model can favour a response position or label independently of the evidence, e.g., the first option or a middle category, even though their meaning is unchanged. | IC2 | Controllable/testable: Present the same choices in different balanced orders and compare the judgments. If order effects occur, rotate or balance label positions across items. Prespecify and keep each category unchanged across versions. |
| 13. Several items or tasks in the same model call (E.6) | When several items, tasks or studies share one prompt, its judgement on one item can be affected by the other items being evaluated at the same time. The size and direction of this effect depend on the model, task, number of items, and complexity of the material. | IC2 | Controllable/testable: Compare the same item alone and with other items. For important decisions, consider separate calls for each item or task, particularly when validation shows that combined calls reduce performance If several are combined, keep them clearly labelled and test the chosen setup. |
| 14. Forcing structured output (F.4) | Requiring an LLM to return its output in a fixed format, e.g., JSON, XML, or a set response structure, can sometimes change the quality of its judgement or required output. The effect depends on the task, the instructions, the format, and the model. This is more likely when the format is complicated and the task requires reasoning. | IC2 | Controllable/testable: Compare the required structured format with a minimally constrained or free-text output and assess whether substantive judgements change. Use the simplest schema needed and validate it before the main evaluation. Do not assume structured output improves substantive accuracy simply because it reduces formatting errors. |
| 15. The output quietly leaves things out (H.5) | The model may leave requested components out of its output, e.g., prompt instructions, output values, or supporting evidence. | Both | Controllable/testable: Require one output for every request and check for missing or duplicate items. Repeat identical runs to test whether omissions change judgments across runs. |
| 16. Document distorted (E.1) | PDF parsing, optical character recognition (OCR), or image-to-text conversion can incorrectly recognise, omit, reorder, merge, split, or associate text, numbers, labels, or table content before the LLM evaluates the source. OCR software converts text in images or scanned documents into machine-readable text. The LLM may therefore make a judgement using a representation that differs from the original document. | IC2 | Controllable/testable in all deployments: Compare converted content with the original source and, where feasible, compare alternative parsers or OCR methods. Where externally controlled, set the parser or OCR software and settings; in managed chat, internal conversion may remain hidden. Where possible, archive the converted input used for evaluation, and escalate failed or uncertain conversions for human review. |
| 17. Converting images or other media (E.4) | Resizing, compression, resampling, image conversion, or other preprocessing steps can result in distortion or loss of evidence before the model receives it. | IC2 | Partly controllable/testable in all deployments: In chats, users can preprocess media before upload, but some processing is vendor-controlled. API users can preprocess media, although some vendor handling may still occur. For testing, fix preprocessing settings and the original media where possible. Test whether different preprocessing settings change the judgement. |
| 18. Evidence buried mid-document gets under-used (E.7) | With a long prompt, evidence in particular positions, especially in the middle, may be under-used even when it is present. Repeated runs can agree and still miss the same evidence. | IC2 | Controllable/testable: Keep inputs within the context range; use targeted excerpts where justified; test-controlled evidence relocation or matched shorter versions as sensitivity analyses. If segmentation or retrieval is introduced to manage long inputs, validate the new pipeline and evaluate retrieval completeness and variability (E.3) separately. |
| 19. Quoted or extracted evidence (H.4) | When the model must support a judgement or extraction with a quote, location, or extracted value from text or a table, it may return different evidence across identical runs (IC1). It may also return evidence that is not present in the source, is copied incorrectly, is assigned to the wrong source, section, treatment group, outcome, subgroup, time point, row, column, or label, or does not support the judgement it accompanies (IC2). | Both | Controllable/testable: Repeat the extraction and compare it with human-checked evidence. Check separately whether the quoted or extracted evidence exists in the source, is attributed to the correct source, and location in the text, and actually supports the judgement. |
| 20. Hallucinated evidence (H.3) | The model cites hallucinated evidence or attributes a claim to evidence that does not exist or does not support it. | IC2 | Controllable/testable: Verify every model-generated citation or restrict the model from searching the internet. Use automated resolution plus human review. |
| 21. Accepting unchecked | The LLM accepts an assumption in the prompt without checking it against the source evidence that could verify or | Both | Controllable/testable: Keep factual claims separate from instructions and ask the model to check important assumptions against the source. Test by comparing |
| assumptions (H.2) | contradict the assumption. It then allows the assumption to influence its judgment. | | matched prompts with correct, incorrect, and no added assumptions. |
| 22. The stated reasoning is not the real reasoning (H.1) | A model-generated rationale does not reflect or logically link to the evidence or reasoning. This can be shown when the evidence is not acknowledged in the rationale, or when the rationale changes but the judgment remains the same. The mechanism concerns whether the stated rationale is linked to the judgment, not whether statements in the rationale are factually correct or supported by the evidence. | IC2 | Controllable/testable: Compare the rationale with source evidence or a reference rationale. Change relevant information and check whether the judgement and explanation respond in the expected way. Do not treat a generated rationale as proof of how the model reached its judgement; internal reasoning generally cannot be directly verified. |
| 23. Failure to integrate task components (H.7) | The model may assess individual parts of a task but fail to connect them or apply the required rule when reaching the overall judgement. | IC2 | Applicable to all deployments: Test each task component and the final integration step separately. Provide an explicit decision rule, check whether all relevant components were considered, and independently verify the combined judgement. |
| 23a. Failure to connect task components (H.7a) | The model fails to connect related evidence, instructions, or criterion components that must be considered together. | IC2 | Applicable to all deployments: Supply cases requiring evidence from multiple parts, verify whether required connections were made, and review the resulting judgement against the source and criteria. |
| 23b. Incorrect integration-rule application (H.7b) | The model identifies individual components but combines them using a rule that differs from the prespecified decision rule. | IC2 | Applicable to all deployments: Record the component-level results and apply the prespecified integration rule independently. Compare the calculated judgement with the model's final rating. |
| 24. Source-embedded instruction interference (H.8) | Instructions contained in the supplied evidence may be treated as directions to the model rather than as material to evaluate, changing its judgement or required outputs. | IC2 | Applicable to all deployments: Keep source content separate from task instructions. Test matched evidence with and without embedded directives, inspect whether the model follows them, and verify ratings against the source. This is a specific source-instruction failure pathway. |
| 25. Surface-cue substitution in criterion judgement (H.9) | The model may rely on superficial wording, layout or rubric cues instead of evaluating the evidence required by the criterion. | IC2 | Applicable to all deployments: Compare matched cases that preserve criterion-relevant evidence but vary superficial cues. Check whether ratings respond to the actual evidence rather than those cues. |
| 26. Different agent runs can reach different judgements (G.1) | Even when an agent follows the same multi-step workflow, repeated runs can use different tools, evidence, or intermediate steps and reach different final judgements. For example, a ROBIS appraisal agent may retrieve different passages across runs and assign a different final judgement. | IC1 | Controllable/testable in all deployments using agentic workflows: Repeat the same workflow several times and compare final judgements; inspect intermediate agent actions where accessible. Structure intermediate outputs and use consensus across generated retrieval queries where applicable. |
| 27. Verification material changes before checking (G.2)† | The agent or verifier checks a summary, selected excerpt, or other representation that changes or leaves out information from the original evidence. The verification may therefore miss a problem in the judgement. | IC2 | In retrieval-supported or agentic workflows, compare the verification input against the original evidence where the actual verifier input is accessible. Preserve the complete source and audit what was selected or transformed when these steps are visible. Test whether verification decisions change when the original evidence is used. |
| 28. Model updates can change the judgement (B.2)† | A vendor model or service can change over time, so the same task may produce a different judgement even when the user repeats the same setup. | IC1 | Testable in vendor-managed interfaces, agentic or work environments, and APIs, and directly controlled in user-controlled model runtimes: Record the model identifier and access date. After a known model or version update, re-run a fixed reference set and compare both overall performance and item-level substantive judgements before continued use. In self-hosted systems, fix the model and serving environment and test new versions before use. |
| **API, vendor-managed agentic environments, or self-hosted** | | | |
| 29. The model learned it during training (D.1) | The model may have encountered the reference judgement, a published appraisal, or the same question with its answer during training. It may therefore provide an judgement partly because it has seen the reference judgement before, rather than because it independently appraised the attached evidence. Apparent accuracy can be high and highly repeatable, so repeated agreement does not rule out contamination. | IC2 | Usually not directly controllable or reliably verifiable in closed vendor-managed models; API access alone does not reveal training data. More directly investigable when training data and provenance are available, including some researcher-developed models: When training data or source are available, compare materials with the available training corpus using lexical and semantic overlap checks, and add model-based contamination tests. |
| 30. Randomness in token selection (C.1) | When sampling is used, the model chooses each new token (i.e., word) from a set of possible tokens based on their probabilities. This means that the same evidence, prompt, model, and declared settings can produce different judgements across repeated runs. The amount of randomness depends on the decoding settings. The random seed also affects which tokens are selected when sampling is used. | IC1 | Controllable/testable: Where sampling controls are exposed, use the least stochastic decoding appropriate for the task and fix and report the decoding settings and seed when available. Lower or zero temperature may reduce sampling variation but does not establish determinism or remove the need for repeated-run testing. |
| 31. Splitting a document can separate related information (E.2) | Internal segmentation of an uploaded document into smaller parts can separate information that needs to be read together. This can change the model's judgement. | IC2 | Controllable/testable where segmentation controls are exposed, including some APIs, vendor-managed agentic or work environments with accessible file tools or logs, and user-controlled model runtimes: Keep document segmentation unchanged for the main evaluation. Test alternative segmentation methods as a sensitivity analysis and, when a judgement depends on evidence across segments, test the complete segmentation-to-aggregation workflow. Use human review when relevant links between segments are uncertain. |
| 32. Different evidence may be retrieved (E.3) | The retrieval step may supply different, incomplete, or misleading evidence, which can change the model's judgement. | Both | Partly controllable/testable where retrieval can be configured or logged, including some APIs, vendor-managed agentic or work environments with query and tool logs, and user-controlled model runtimes: Repeat retrieval with the same setup and compare the retrieved evidence. Keep the retrieval setup fixed and test alternative query wording separately as a sensitivity analysis. |
| 33. Loss or distortion during context compaction (J.1)† | Earlier task context is shortened into a summary or other representation, and required evidence, instructions, or qualifications may be lost or changed before a later judgement. | Both | Where compaction is enabled: Where logs or compaction controls are exposed, verify that compaction actually occurred and compare the original and compacted context; otherwise assess observable effects, and check required evidence and instructions. Repeat identical runs for IC1. Preserve original context and reinsert critical details. |
| 34. Removal of task-relevant context (truncation or pruning) (J.2)† | Evidence, instructions or other task-relevant material previously available to the model is removed rather than preserved in a replacement summary. | Both | Managed chat interfaces and vendor-managed agentic environments where enabled: Inspect available removal logs, where exposed, and compare what was available before and after pruning; managed chat generally does not expose internal pruning logs. Retain necessary evidence and instructions; repeat identical workflows and independently verify judgements. |
| 35. Loss of required instructions across turns (J.3)† | Instructions set in an earlier turn are not included in the assembled input for a later assessment, so the model may no longer apply required criteria or steps. | Both | Multi-turn managed chat, APIs and agentic systems: Where assembled prompts or logs are accessible, log assembled prompts for every continuation, check that required instructions are present, and resend or version standing instructions where needed. Verify final ratings and required outputs. |
| 36. Hidden model switching (B.1)† | A vendor may re-route prompts to different model variants, versions or inference routes without changing the model identity reported to the user. Hidden request-to-request routing can change judgements across repeated runs (IC1); undisclosed substitution can also systematically affect the judgement in a way that is not apparent from the judgement itself (IC2). | Both | Controllable/testable: In vendor-managed environments and APIs, use fixed model versions where available and record available model metadata and assess unexpected changes; undisclosed routing generally cannot be directly verified. In user-controlled runtimes, fix the routing configuration and log which model serves each request. |
| **Self-hosted (user-controlled runtime)** | | | |
| 37. Tokenisation effects (E.5) | Models read text as small pieces called tokens. How text is split can affect performance or how much source material fits in the model's context limit. | IC2 | Controllable where the model, tokenizer, and input preprocessing are specified, most directly in user-controlled runtimes and some APIs with tokenizer information: Keep the model and tokenizer fixed. Compare models with different tokenizers and test each relevant language separately. |
| 38. Numerical variation during model serving (B.3) | The same model can make slightly different calculations when otherwise identical requests are processed. Usually this does not change the judgement, but a small difference can sometimes change the selected token and eventually the judgement. | IC1 | Directly investigable mainly in a user-controlled model runtime; in hosted systems, repeat runs can reveal variability but generally cannot establish its numerical cause: Change one serving condition at a time to test its effect. Keep the hardware, numerical precision, drivers, software libraries, and serving configuration fixed. Run requests separately. Use deterministic or batch-invariant kernels. |

**Notes:** The table presents 38 mechanisms and two H.7 sub-mechanisms (not counted separately). **Can you do anything about it (by deployment setting:** The deployment settings indicate the level of access generally needed to investigate a mechanism, not where it can occur. Researchers may be able to test a mechanism's effects without being able to identify its cause or prevent it. Vendor-managed agentic environments may provide access to files, retrieval steps, verification inputs, and context logs that are unavailable in commercial chat interfaces, depending on the system. †**Conditional mechanism:** A mechanism that can affect the judgement or required output only when a particular condition or event occurs. **IC1 (repeat-run variability):** A mechanism that can produce materially different judgements or required outputs across repeated runs of the same task under the same specified conditions. **IC2 (hidden judgement failure):** A mechanism that can affect the judgement or required output in a way that is not apparent from the judgement itself. **Both:** A mechanism that can produce either IC1 or IC2, although these effects need not occur under the same circumstances. The IC1/IC2 classifications indicate which types of failure a mechanism could cause, not whether those failures have been demonstrated in published studies. ***Abbreviations:*** *AI, artificial intelligence; API, application programming interface; IC1, inclusion criterion 1; IC2, inclusion criterion 2; LLM, large language model.*

Figure 2 illustrates the distribution of the 38 mechanisms by IC1 and IC2 eligibility criteria and deployment setting. It shows which mechanisms can produce repeat-run variability, hidden judgement failures, or both, and the deployment settings in which they can be investigated or controlled.

**Figure 2. Distribution of the 38 FAILS taxonomy mechanisms by eligibility criterion and deployment setting**

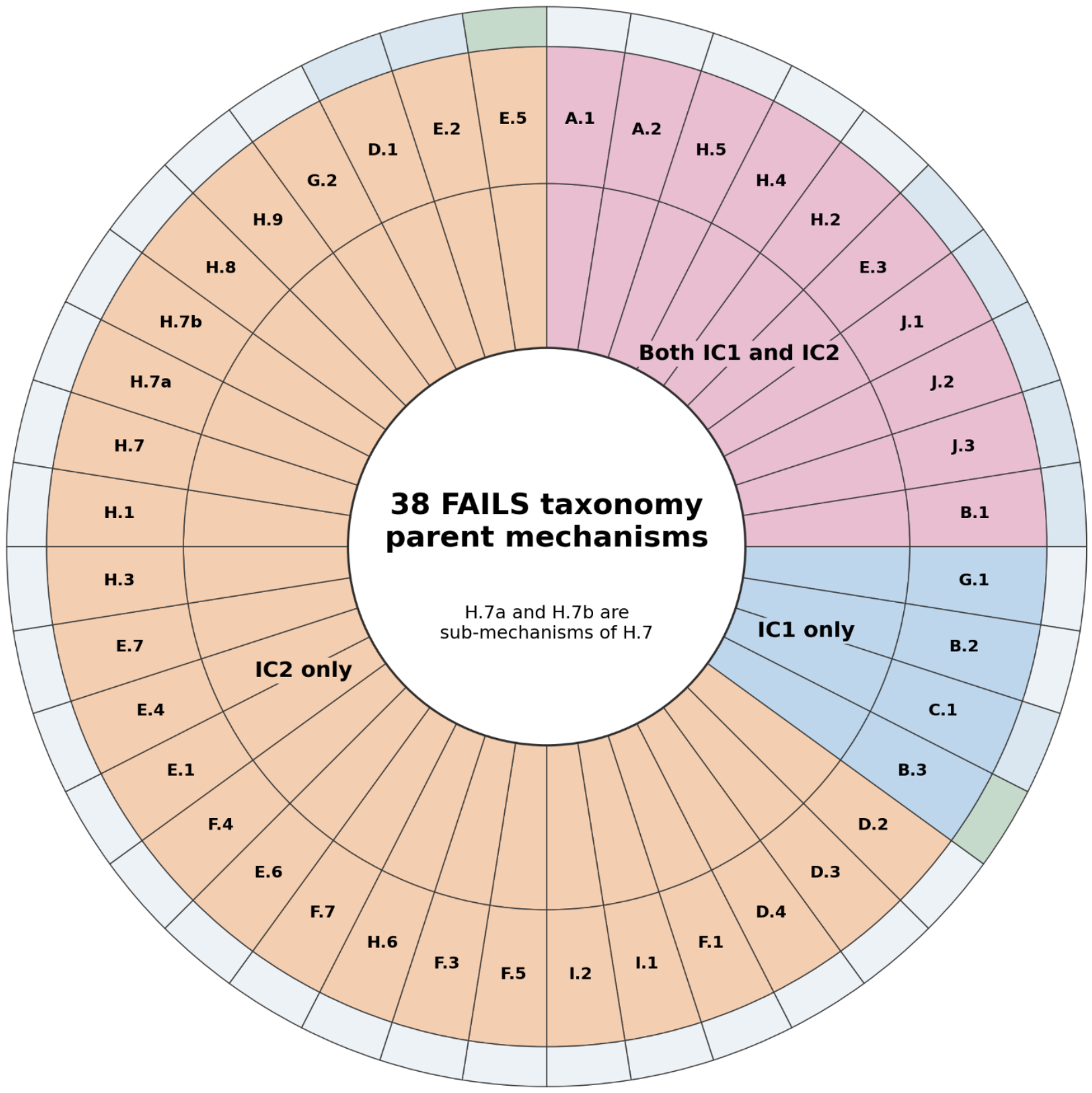

**Notes:** The outer band indicates deployment setting: light grey = all deployments; pale blue = commercial API, vendor-managed agentic environments, or self-hosted; green = self-hosted only. IC1 = repeat-run variability; IC2 = hidden judgement failure. A mechanism may meet both criteria, although the effects need not occur under the same circumstances.

**Evidence varied across mechanisms**

Thirteen mechanisms were rated as having strong overall evidence, 24 as having moderate evidence, and one as limited evidence (**Supplement 4, Table S1**). When considering evidence specific to read-and-evaluate tasks, including eligible LLM-as-a-judge evaluations, eight mechanisms were rated as having strong evidence, 18 as having moderate evidence, and 12 as having limited evidence. These descriptive ratings reflect the yield and directness of the available evidence. For several mechanisms, evidence from the broader LLM literature was stronger than evidence specific to read-and-evaluate tasks, including the LLM-as-judge literature. For example, personalisation and user history (A.1), model version drift (B.2), and agentic workflow instability (G.1) had strong overall evidence but moderate task-specific evidence, while routing variability (B.1) and serving-time numerical non-determinism (B.3) had limited direct evidence from these tasks. A detailed summary of evidence gaps across the taxonomy can be found in **Supplement 6**.

**Testing approaches**

A mechanism may have a testable pathway to IC1 or IC2 even when inspection, audits, or workflow checks are more appropriate than controlled experiments. IC1 is assessed by repeating the same task while holding the specified source evidence, prompt, model, and researcher-controlled settings fixed, and comparing the resulting judgements and required outputs. The number of repetitions and criteria for material differences should be prespecified. Repeat-run differences establish variability but do not necessarily identify the mechanism responsible.

IC2 can be assessed through controlled robustness or sensitivity analyses, direct inspection of the evaluation process, or verification of the evidence and required supporting outputs. In sensitivity analyses, researchers change one potentially influential feature while keeping the intended task and evidence constant. Examples include changing prompt wording (F.1), context segmentation (E.2), evidence position (E.7), response-option order (F.7), source labels while retaining identical evidence (I.2), or the user's stated position (H.6). These comparisons may reveal an IC2 problem when the LLM's judgement changes in response to a feature that should not influence the decision. For example, changing the journal name while keeping the review text identical should not change an AMSTAR 2 judgement, because the appraisal should be based on the review's methods rather than the journal's reputation. Where several mechanisms could produce similar failures, targeted comparisons and process checks can help distinguish potential explanations, although observing a failure or confirming that a condition occurred does not necessarily establish that a particular mechanism caused it.

**Mitigation strategies**

A mitigation is intended to reduce the probability or magnitude of a mechanism or failure mode operating. Several similar approaches were suggested for the evidenced-based mitigation strategies. Recommendations generally involved removing unnecessary sources of influence, keeping important conditions the same, recording the versions used, and reporting the conditions used for evaluation. Examples included using fresh sessions (A.1), using the same model version where possible (B.2), keeping retrieval settings and components the same (E.3), prespecifying prompts and worked examples (F.1, F.3), and removing unnecessary source labels (I.2). Mitigations can also include automated checks for missing required outputs (H.5), whereas others required manual human verification, such as checking citations (H.3) or supporting quotations (H.4). Employing a mitigation should not be interpreted as evidence that the mechanism was prevented or its effects eliminated. Mitigations may reduce the likelihood or severity of a failure without preventing it entirely. For example, setting temperature to zero may reduce repeat-run variability but does not guarantee identical judgements. Researchers should therefore assess whether the mitigation was implemented as intended and, where feasible, test whether the mechanism or its effects remain.

Where a mechanism is relevant but evidence for a proposed mitigation is limited, test whether the mitigation works for the specific task, model, and workflow, and report that its effectiveness remains uncertain. Mitigation strategies aim to reduce the likelihood or effect of a mechanism, whereas human verification provides a later safeguard by checking for errors or judgement failures that remain. When a human reference standard includes both the judgement and supporting evidence, human verification is not needed during model evaluation. It may instead be relevant as a safeguard in real-world use, where reference-standard judgements are unavailable.

Mitigations can also introduce new mechanisms. For example, using retrieval to reduce the amount of long-context material presented to the model creates an opportunity for retrieval omission or variability (E.3), which should then be evaluated separately. Mitigation is more limited when the relevant process is controlled by a commercial vendor and cannot be directly observed or changed by the user.

**How to apply the taxonomy**

The taxonomy is intended for researchers designing or evaluating LLM-assisted read-and-evaluate tasks, as well as journal editors, peer reviewers, developers, and other users interpreting the resulting evidence. It can inform study protocols, the choice of tests and safeguards, and the appraisal of published results. It does not establish that a particular mechanism has occurred in a given study.

The starting point is a task requiring the LLM to interpret source material against specified criteria and produce a judgement or other required output (Figure 2). Examples include eligibility assessment, interpretive data extraction, risk-of-bias assessment, methodological appraisal, and certainty-of-evidence judgements. Not every mechanism in the taxonomy will be relevant to every task.

First, specify the task (i.e., methodological quality assessment), evaluation criteria (i.e. AMSTAR 2 items), source documents (i.e., PDF of the study, required outputs (i.e., high or low quality), and reference standard (i.e., human assessed systematic reviews). Then document the model and version, deployment setting (such as a commercial chat interface or direct API), instructions, model settings, and other relevant features of the workflow. Record which features the researcher can observe or control and which are managed by the commercial LLM vendor. The mechanisms that may affect an LLM judgement depend on the deployment setting and workflow, the task being performed, and how the LLM is configured and used. Relevant factors include:

- **Model:** The model and version used (e.g., Claude Sonnet 4.5).
- **Deployment setting:** How the model is accessed (e.g., through a commercial chat interface or direct API).
- **Instructions:** The system prompt, task instructions, and decision rules provided to the model.
- **Model settings:** The temperature and other settings that influence how the model generates responses.
- **Context:** The information provided to the model, such as a full systematic review or selected sections, and whether previous conversation history is included.
- **Technical environment:** How the model is routed and run on the provider's infrastructure, including processes that may not be visible or controllable by the researcher.

Some of these factors can be controlled by the researcher, while others are managed by the commercial LLM vendor. Identifying these factors helps researchers determine which mechanisms may be relevant to their evaluation.

Next, document the full workflow from source preparation to recorded judgement. Identify such features as PDF-to-text conversion, optical character recognition, selection or segmentation of evidence, retrieval, external web access, tools, multiple model calls, and agentic or multi-agent steps. These features may introduce mechanisms that would not arise in a simpler workflow. For example, PDF conversion may omit relevant methods text, retrieval may fail to select a relevant passage, and web access may expose the model to external answers that should not inform the appraisal. A full discussion of how mechanisms map to specific read-and-evaluate tasks, deployment setting, workflow and model features is outlined in Paper 3 of the series.

Map the task, deployment, and workflow features to individual taxonomy mechanisms, recording which are applicable, inapplicable, or uncertain and why. This is a prospective assessment of what could go wrong, not a diagnosis that IC1 or IC2 is already present. Prioritise testing and mitigation according to the likely consequence for the judgement, plausibility, feasibility of testing, and degree of user control. Applying every proposed mitigation is neither necessarily feasible nor demonstrably effective. For applicable mechanisms, specify targeted tests, safeguards, human checks where needed, and reporting requirements before the main evaluation. Repeated runs under the same specified conditions assess IC1, whereas deliberate changes to a specified feature test sensitivity and may provide evidence relevant to IC2 if that feature should not legitimately affect the judgement. Verify the required quotations and rationales, and compare judgements with a human reference standard, where appropriate. An observed difference or failure does not, by itself, establish which mechanism caused it. Evaluate whether mitigations reduced the targeted problem, because a mitigation may be ineffective or introduce another mechanism. Report remaining uncertainty, including conditions the researcher cannot observe or hold constant.

As an illustrative example, consider an LLM applying an AMSTAR 2 item to systematic-review PDFs. Researchers might identify PDF conversion distortion (E.1) and source-label influence (I.2) as relevant concerns. They could check the converted PDF against the original and verify quotations supporting the judgement. They could also repeat the same appraisal, with the same specified inputs and settings, to estimate IC1 repeat-run variability. In a separate sensitivity test, they could change only the journal name while keeping the review methods unchanged. If the judgement changes, this supports concern about inappropriate source-label sensitivity (IC2), but further checks are needed to determine the cause. Neither identical repeated judgements nor successful application of a mitigation proves that all relevant failures have been excluded.

Readers and evaluators of published LLM-assisted studies should ask which mechanisms were applicable, which were tested, and what evidence supports the authors' claims. Repeating the task under matching specified conditions can investigate IC1, although a reader may be unable to recreate hidden model or serving conditions. IC2 generally requires additional checks, such as evidence verification, comparisons against reference judgements, or targeted sensitivity tests; repeating the task alone is insufficient. If the paper does not report these tests or enough details to reproduce them, readers should describe the potential problem as unassessed rather than conclude that the mechanism occurred or was absent. Readers are not necessarily expected to replicate the whole study, but may be unable to independently establish whether particular failures affected the findings.

## DISCUSSION

### Principal findings

The TRAILS taxonomy provides a structured framework for identifying mechanisms that may produce repeat-run variability (IC1), hidden judgement failures (IC2), or both, and for selecting appropriate mechanism-specific tests and potential mitigations. Thirty-four mechanisms met the IC2 criteria and therefore cannot be identified by repeated-run testing alone. A model may repeatedly give the same judgement because, for example, it has seen the reference judgement during training (D.1), retrieves an existing risk of bias judgment from the internet (D.2), or is influenced by the writing style of the material being judged (I.1). High repeat-run agreement therefore does not show that the judgement was based only on the supplied evidence and appropriate decision criteria.

Testing IC1 generally requires repeated runs under the same specified conditions, whereas investigating IC2 may require robustness tests, sensitivity analyses, direct evidence verification, or other targeted assessments. The same observed failure may arise from different mechanisms, so identifying a plausible explanation does not establish that the mechanism occurred or contributed to the failure. The relevance of individual mechanisms, as well as researchers' ability to observe, control, or test them, depends on the specified task, deployment setting, and technical features of the workflow. The taxonomy may therefore help researchers design more comprehensive evaluations, identify limitations that conventional accuracy assessments may not reveal, and support efforts to improve the repeatability, reproducibility, and replicability of LLM-assisted tools and studies. However, the effectiveness of many proposed tests and mitigations remains to be established empirically.

The taxonomy is useful because it links an observed problem to mechanisms that can be tested directly. However, IC1 and IC2 describe the testing method, not which mechanism caused the problem. Evidence may be unavailable, overlooked or used inappropriately because text was lost during PDF conversion (E.1), separated during document segmentation (E.2), underused because of its position in a long document (E.7), omitted by retrieval (E.3), or inappropriately linked to the judgement by the model (H.4). These mechanisms can produce similar judgements but require different tests and mitigation strategies. Mitigation strategies may reduce but not remove the risk of these mechanisms contributing to unreliable findings. Testing is therefore recommended regardless of whether mitigation strategies are employed.

### Challenges in applying the taxonomy

Applying the taxonomy presents several practical challenges. Testing and mitigating potential failure mechanisms can require substantial resources, even when it is uncertain whether a mechanism will occur or materially affect the results. Implementing a mitigation does not establish whether it prevented the mechanism, reduced its occurrence, or reduced the magnitude of its effects. Knowledge users also face challenges when interpreting published LLM evaluations, as IC1 and IC2 cannot necessarily be identified from the reported judgements alone. These challenges highlight the need to prioritise mechanisms according to the task and evaluation setting, assess the effectiveness of mitigations, and report sufficient methodological details and results.

**Accuracy**

Most published studies evaluating LLMs have focused on accuracy.[14] However, a study may have high accuracy with low repeatability, high repeatability with low accuracy, or high values for both despite the presence of hidden judgement failures (IC2). Thus, accuracy and repeatability alone may not reveal all failures affecting LLM judgements or required outputs. Accuracy, repeatability, and judgement validity answer different evaluation questions, but none alone captures all potential failures in LLM read-and-evaluate tasks. IC2 may also require assessments of evidence support, output completeness, task completion, or other mechanism-specific outcomes.

**Implications on evidence synthesis findings**

Residual repeat-run variability matters when LLM judgements feed into later stages of an evidence synthesis. Where a mechanism cannot be prevented, or mitigation only partly reduces its effect, evaluation should measure both variation in the LLM judgement and whether that variation changes the review findings. This can be examined by carrying the complete judgements from repeated runs through the same prespecified downstream review steps. For example, if repeated screening decisions produce different sets of included studies, the meta-analysis can be repeated for each set and the resulting pooled estimates compared. If risk of bias judgements vary, the relevant sensitivity analyses or certainty assessments can be repeated to determine whether the interpretation of the evidence changes.[15] This distinguishes repeat-run variability at the point of the LLM judgement from variation that changes what the review ultimately reports. The consequence of the same mechanism can therefore differ according to the evidence synthesis task and where the LLM judgement enters the review workflow. Human verification can intercept a variable, unsupported, or inappropriately influenced judgement before it propagates to later review stages, but this is a workflow safeguard rather than evidence that the underlying mechanism cannot operate.

A related problem occurs if the model retrieves an existing appraisal rather than deriving the judgement from the supplied study, or if it encountered the reference judgement during training (D.1). The model may then repeatedly produce a judgement that matches the reference without demonstrating that it can perform the appraisal task. Contamination occurs when material used to evaluate a model, including information about the expected reference judgement, was already available to the model before it performed the evaluation.[16] Private or prospective reference sets reduce this risk, but they are difficult to create when the material being evaluated is already published. Reference sets can be kept temporarily unavailable, placed under controlled access, or refreshed over time, with materials released once they are no longer needed as unseen evaluation items. The aim is not secrecy, but preserving an uncontaminated test set at the time the model is evaluated.[16, 17]

**Relation to reporting guidance and evaluation frameworks**

The taxonomy addresses questions that are complementary to, but different from, those addressed by reporting checklists and guidance. Existing AI reporting and risk of bias frameworks were developed primarily to improve transparency, methodological reporting, responsible use, or assessment of study quality and risk of bias.[18-27] They were not designed to identify the mechanisms that can produce repeat-run variability or a hidden judgement failure. Reporting the model, prompt, and workflow does not establish whether a judgement was reproduced from training data (D.1), whether relevant evidence was underused because of its position in the input (E.7), or whether the order of the response options (F.7) influenced the judgement. Complete reporting remains necessary, but does not investigate potential pathways to IC1 or IC2 failures.

**Implications for LLM evaluation studies**

The taxonomy could complement AI evaluation frameworks such as Inspect AI[28] by helping researchers identify relevant failure mechanisms and select targeted tests. Inspect AI provides infrastructure to run evaluations, score outputs, and analyse results. For example, researchers could assess whether an LLM scorer produces materially different judgements across repeated runs or whether its judgements are supported by the available evidence. Integrating these tests into frameworks like Inspect AI could strengthen the evaluation of LLM-based scoring in AI research.

Evaluation design should reflect the distinction between IC1 and IC2. When relevant mechanisms are plausible, researchers should prespecify appropriate tests, including repeated runs and definitions of material differences for IC1, and mechanism-specific robustness tests, sensitivity analyses, or direct checks for IC2. Evaluations should assess the implemented workflow, including document processing and retrieval where applicable, rather than attributing performance to the LLM alone. Relevant validation and mechanism-specific tests should be repeated after material changes to the model, instructions, inputs, or workflow.

**Reproducibility and human oversight**

The taxonomy can help identify where human verification is most valuable. Some mechanisms point to checking document conversion or retrieved evidence, while others point to verifying supporting quotes, assumptions, or final judgements. For web platforms, users may be able to verify returned quotes against the original source, but internal document conversion and retrieval are often not visible. For API-based workflows, researchers can usually save and inspect converted text and retrieved passages. This allows human oversight to be targeted to parts of the workflow where a relevant mechanism has an opportunity to affect the result.
In addition, identifying, testing, and controlling relevant mechanisms helps create the conditions needed for reproducible LLM evaluation.

**Relevance to current and evolving LLM systems**

Although some mechanisms were identified using earlier generations of LLMs, evidence indicates that several failure pathways remain relevant to current models (e.g., GPT-6 Astra, Claude Fable 5.1, Gemini 3.1 Pro).[11, 29, 30] Three studies examining candidate position effects, long-context position effects, and rubric-related effects found that model judgements continued to be influenced by the position of candidate responses, the location of information within long documents, and the wording of evaluation criteria, respectively.[11, 29, 30] Improvements in model capabilities and technical controls may also address specific aspects of model performance without preventing other failures. For example, OpenAI reported that a new 2024 API-level technical feature called Structured Outputs, prevented output-format failures but acknowledged that the information contained within those outputs could still contain failures.[31]

The relevance of the mechanisms may extend beyond conventional LLMs as new AI systems for classification and judgement emerge.[32] Jev, for example, may prevent some technical output failures, such as returning deformed structured output, because its responses return predefined structured judgements with probabilities (e.g., high risk: 0.72, low risk: 0.21).[32] One study of 41,500 responses demonstrated that irrelevant numerical information and the order of response options could influence Jev's decisions.[33] OpenAI's Decisions API similarly returns predefined choices, probabilities or scores, but operates as a specialised interface to an underlying model rather than a separate decision-model architecture.[34] Although these systems differ in design, neither structured output format establishes whether the resulting judgement is correct or appropriately supported by the evidence. The TRAILS mechanisms are relevant to both, although which ones apply, and their effects on judgements, require empirical investigation.

The TRAILS taxonomy should be treated as an evolving framework. As LLM systems develop, some mechanisms may become less common or have smaller effects, while others may become more important or arise in new forms. Improvements in document processing, long-context capabilities, and inference reproducibility may reduce certain failures, although recent research shows that long-context position effects and numerical non-determinism remain concerns.[29, 35] At the same time, increasing use of retrieval, multimodal inputs, personalisation, and agentic workflows may introduce additional opportunities for failures involving evidence selection, context management, and interactions between workflow steps.[36, 37]

**Relationship to other failure mode frameworks**
Our taxonomy differs from earlier frameworks in its purpose and inclusion criteria. Previous studies have classified sources of irreproducibility in machine learning, system-level failures in LLM applications, and problems affecting LLM-as-a-judge evaluations.[2, 13, 38] Other studies have examined repeatability, agreement with human assessors, and sensitivity to changes in input presentation (Norman et al., 2026; Tiwari, 2026).[39-43] Building on this work, we required each mechanism to have a testable connection to IC1, IC2 or both. This provides a common basis for assessing mechanisms described using different terminology and across research settings, and distinguishes repeat-run variability from sensitivity to deliberately changed conditions.

**Research priorities**
No standard measurement framework has yet been established and validated for quantifying repeatability and mechanism-specific sensitivity across evidence synthesis read-and-evaluate judgements.[6] The appropriate measure depends on the output and mechanism being tested. For example, numerical variation during model serving (B.3) can be investigated by repeating the same task under specified conditions, whereas sensitivity to prompt wording (F.1) or response-label order (F.7) can be assessed by deliberately varying those features. Gartlehner et al.[44] assessed test–retest reliability by repeating 160 data extractions across three runs and compared accuracy and error patterns. Other studies used percentage agreement and Cohen K,[7, 44, 45] for repeated judgements, and assessed sensitivity tests using differences in accuracy, recall, sensitivity, or other measures.[46-48] However, future research should determine which mechanisms materially affect different tasks and deployment settings, and whether proposed mitigations reduce failures without introducing new sources of error. The fourth paper in this series addresses measurement by matching tasks, output types, units of analysis, comparison designs, and downstream consequences to the appropriate metrics.

**Limitations**
This study has several limitations. First, the taxonomy identifies mechanisms that can affect LLM read-and-evaluate tasks but does not estimate how frequently each mechanism occurs or the magnitude of their impact, if any. Their opportunity to occur differs across deployment settings, tasks, and workflows. Some mechanisms are broadly applicable, whereas others require specific features such as retrieval, long-context input, worked examples, multimodal processing, or agentic workflows. This will be explored in the fourth paper in our series. Second, the evidence base was identified using unstandardised terminology and a structured, iterative search, including grey literature such as preprints, conference proceedings, and therefore technical documentation and relevant studies or mechanisms may have been missed. Thus, the taxonomy should not be interpreted as exhaustive. Third, the taxonomy was developed and refined by the author group rather than through formal external consensus or validation, such as a Delphi process. Finally, the taxonomy is intentionally limited to read-and-evaluate tasks and is not a comprehensive taxonomy of LLM or machine learning failures.

## CONCLUSIONS
In conclusion, a single LLM judgement provides limited information about whether the model would produce the same result on repeated runs or whether that judgement was reached using the intended evidence and evaluation criteria. Although repeated testing can identify variability, consistent judgements may still conceal failures in evidence selection, interpretation, or substantiation. Similarly, high agreement with human reference judgements does not establish that the LLM performed the intended evaluation process. The TRAILS taxonomy provides a framework for identifying plausible mechanisms underlying these problems and selecting targeted tests and potential mitigations. Researchers should therefore evaluate accuracy, repeatability, evidence support, and relevant failure mechanisms rather than relying on the accuracy or consistency of final judgements alone. When LLM judgements feed into later stages of an evidence synthesis, researchers should also assess whether failures could change review findings or conclusions. The taxonomy provides a framework for identifying relevant failure mechanisms; however, it does not establish that every mechanism occurs in every LLM system or that the proposed mitigations prevent those failures.

**Tables and Figures:**

| Type | Display | Function |
|---|---|---|
| **Box 1** | Glossary of important definitions | Definitions |
| **Figure 1** | Repeatability and hidden failures | Explains the core conceptual distinction |
| **Table 1** | Condensed and plain language 38-mechanism taxonomy | Shows what the mechanisms are |
| **Figure 2** | Distribution of the 38 FAILS taxonomy mechanisms by eligibility criterion and deployment setting | Matches deployment to mechanism |

**Supplementary material:**
Supplementary File 1. Glossary of LLM terms
Supplementary File 2. Exclusion decisions and search, extraction, and synthesis methods
Supplementary File 3. Tutorial on types of deployment settings with worked examples
Supplementary File 4. Full taxonomy, evidence base, and testing and mitigation strategies
Supplementary file 5. Taxonomy checklist for LLM read-and-evaluate tasks
Supplementary file 6. Evidence gaps

**Contributorship statement:** CL conceived and designed the study. Data collection, extraction, and synthesis: CL, AS, HH, CB, SK, IG, NJ, YO, IPN, LS, GD, SB, SK, DC; CL wrote the first draft of the manuscript. All authors edited subsequent drafts.
**Competing interests:** None declared.
**Funding:** No funding was received.

# Supplementary 1: Glossary of LLM terminology

## A. Models and technical features

**Artifact.** An artifact is a feature of the evaluation input or procedure that may influence the LLM's judgement without being relevant to the intended evaluation. A failure mechanism is the process through which such a feature, or another system condition, can produce repeat-run variability or hidden judgement failure.

**Large language model (LLM).** An artificial intelligence model trained on large amounts of data, including text, to interpret instructions and generate language-based outputs.

**Model.** The computational system used to process inputs and generate outputs; a model family may have multiple versions or variants.

**Model version or variant.** The specific release or variant used for an evaluation. Different versions may behave differently.

**Model weights.** Numerical parameters learned during training that influence how a model processes inputs and generates outputs.

**Open-weight model.** A model whose trained weights are made available for download and use, subject to its licence. Open weights do not necessarily mean that training data, code, or development history are available.

**Model provenance.** Who developed or modified the model and what is known about its training and post-training history.

**Sampling settings.** Settings affecting how the model selects generated text; availability and user control vary across models and interfaces.

**Temperature.** A sampling parameter that can affect variation in generated responses, where exposed by the model provider.

**Context window.** The amount of input and generated content a model can process within a given context; practical handling may depend on the application.

**Model routing.** Selection of a model, variant, or serving pathway for a request, potentially without direct user control.

**Safeguards.** Provider- or researcher-implemented measures intended to prevent, detect, or reduce specified undesirable behaviour or outputs. Their presence does not establish effectiveness.

**Vendor-controlled.** A relevant parameter or process is controlled by the provider and cannot be directly fixed or inspected by the researcher.

## B. Evidence and document processing

**Source evidence.** The documents or other information that should be used to perform the read-and-evaluate task.

**Parser.** Software that converts a document or model output into text or structured data for subsequent processing.

**Optical character recognition (OCR).** Software that converts text in scanned pages or images into machine-readable text. Recognition errors may alter supplied evidence.

**Chunking or context segmentation.** Dividing a document into smaller sections for separate processing or presentation to a model; for example, splitting a systematic review into sections.

**Retrieval.** Selecting documents or passages relevant to a request or evaluation.

**Retrieval-supported workflow.** A workflow that retrieves selected evidence and supplies it to a model when it produces a judgement.

**Multimodal processing.** Processing more than one form of information, such as text, figures, images, or tables.

**External internet or web access.** A capability allowing a model or its workflow to obtain information from websites beyond the supplied task material.

**Evidence traceability.** The ability to identify and check the source evidence cited or used to support a judgement.

## C. Deployment settings and workflow architecture

**Deployment setting.** How a model is accessed and which parts of its runtime a researcher can specify, observe, or control.

**Managed interface.** A vendor-hosted chat service or application where users provide prompts and files but many serving and context-handling processes remain provider-controlled.

**Vendor-managed agentic or work environment.** A provider-managed environment in which users can configure files, tools, permissions, or instructions while some orchestration and model settings remain provider-controlled.

**Direct model API.** A developer-facing hosted service through which research software sends requests and receives outputs; some model settings can be specified, while the provider controls serving infrastructure.

**User-controlled model runtime.** A model run by the researcher or organisation, with direct control over model weights, inference software, hardware, sampling settings, and context configuration.

**Workflow architecture.** How model calls, retrieval, tools, and processing steps are combined to produce an output.

**Single-call workflow.** A workflow in which the final output is produced from one direct model call.

**Agentic workflow.** A workflow containing dependent steps in which a model can select actions, use tools, choose arguments, assemble context, or decide when to stop.

**Multi-agent workflow.** A workflow in which more than one model instance or assigned model role contributes to, critiques, or combines the final output.

**Tool.** An external capability used by the workflow, such as document retrieval, code execution, or web access.

**Persona.** An instruction asking the model to respond from a specified role or perspective.

## D. Task and evaluation terminology

**Task.** The full activity the model is instructed to perform, including the judgement and any required supporting steps or outputs.

**Prompt.** The text or other instructions presented to the model to specify or guide how the task should be performed.

**Run.** One complete performance of the specified task or workflow.

**Judgement.** The substantive decision, rating, classification, inclusion/exclusion decision, or other task-level determination.

**Output.** The full response returned by the model, potentially including the judgement, quotations, rationale, references, and required fields.

**Supporting output.** Material accompanying or substantiating the judgement, such as quotations, evidence locations, references, or explanations.

**Specified conditions.** The source evidence, task, prompt and instructions, specified model variant and version, and user-accessible model settings held the same for a repeat-run comparison.

**Material difference.** A difference affecting interpretation, fulfilment of task requirements, or downstream use, rather than wording alone.

**Accuracy.** Agreement with a prespecified reference judgement for a specified evaluation dataset.

**Repeatability.** Agreement between repeated judgements under the same specified conditions.

**Reproducibility.** Agreement when an independent analyst repeats the evaluation using the same data, methods, and specified conditions.

**Judgement validity.** Whether the judgement was determined by information that should legitimately determine it.

**Robustness or sensitivity testing.** Deliberately changing a specified condition and examining whether the judgement or required output changes. This does not itself demonstrate IC1 or IC2.

**Mechanism.** A process or condition that can contribute to a specified failure.

**Failure mode.** An operationally defined pattern in which a problem appears; its underlying computational cause may not be established.

**Detection.** Identifying that a mechanism or failure occurred.

**Prevention.** Stopping a mechanism or failure from occurring.

**Mitigation.** Reducing the likelihood or impact of a mechanism or failure; applying a mitigation does not establish its effectiveness.

**Controllable or testable.** A condition can be held fixed or a targeted diagnostic comparison can be performed to investigate its effect.

**Unverifiable.** The researcher cannot directly establish from an individual judgement whether a mechanism occurred, although risk-reduction or indirect testing may still be possible.

## Taxonomy inclusion criteria

**IC1, repeat-run variability.** A mechanism can produce materially different judgements, extractions, or required supporting outputs across repeated runs of the same read-and-evaluate task under the same specified conditions.

**IC2, hidden judgement failure.** A mechanism can systematically affect the judgement or required output in a way that is not apparent from the judgement itself. Accuracy and repeatability do not rule it out.

# Supplement 2. Exclusion decisions, search and synthesis methods

This supplement reports the study- and mechanism-level eligibility criteria, reasons for excluding or combining candidate mechanisms, and methods used to search for, extract, assess and synthesise evidence.

## 1.0 Eligibility criteria

Eligibility was assessed in two stages: first for the sources used to identify and describe mechanisms, and then for each proposed mechanism.

### Phase 1. Study eligibility criteria

#### Inclusion criteria

- Eligible sources described or investigated factors that could change LLM outputs or affect the evidence, reasoning, judgement or required supporting output in read-and-evaluate tasks. A source could help identify a plausible mechanism without directly demonstrating IC1 or IC2.
- Sources were limited to English-language publications and documentation. This may have excluded relevant work published in other languages, particularly technical reports from non-English-speaking research groups.
- We considered peer-reviewed journal articles, conference papers and preprints with enough methodological detail to assess their relevance. We also used vendor and laboratory documentation to establish technical processes or available system features. Documentation alone was not treated as proof that a judgement failure had occurred.

#### Exclusion criteria

- We excluded studies concerned only with researcher decisions or reporting outside the LLM evaluation workflow, such as selective reporting, p-hacking or hypothesising after results were known (HARKing). These practices can bias research but are not, by themselves, mechanisms within an LLM read-and-evaluate task.
- We excluded sources concerned only with barriers to reproducing a published study, such as missing code, inaccessible data or broken links, unless those problems also affected the LLM judgement or required output.

### Phase 2. Mechanism eligibility criteria

#### Inclusion criteria

- IC1 (repeat-run variability): Can the mechanism produce materially different judgements, extractions or required supporting outputs when the same read-and-evaluate task is repeated under the same specified conditions? These conditions include the evidence, task, prompt, instructions, stated model version and user-accessible settings. Hidden routing or serving conditions may still vary. Changes in wording alone do not establish IC1.
- IC2 (hidden judgement failure): Can the mechanism affect the judgement, its supporting evidence, or completion of a required task in a way that cannot be ruled out by reading the final judgement or observing agreement across repeated runs?

A material difference is a change in the judgement, classification, decision, required task completion or supporting output that could affect interpretation or later use. Differences in wording or formatting alone are not material. Deliberately changing a prompt, parser, label order or other specified condition is a sensitivity or robustness test, not an IC1 repeat-run test. Researchers can repeat calls within each condition to assess IC1 and compare conditions to examine potential IC2 failures.

We also set boundaries for individual mechanisms. For example, accepting unchecked assumptions (H.2) applies when the model relies on a claim or assumption without checking available evidence that could confirm or contradict it, although the task requires an evidence-based judgement. It does not apply when the task asks the model to accept a hypothetical premise as true. An excluded example may still involve another mechanism.

## Illustrative examples of IC1 and IC2

These hypothetical examples show the difference between IC1, IC2 and a situation in which both could occur. They are illustrations, not findings from the studies included in this review.

- IC1-only illustration: The same trial PDF and risk-of-bias criterion are assessed in repeated independent runs with the same specified model and settings. The model gives materially different ratings. This establishes repeat-run variability, but does not by itself identify whether sampling, routing or another mechanism caused it.
- IC2-only illustration: The model is asked to assess a risk-of-bias item using a supplied trial PDF. It gives the same judgement across repeated runs, but its tool log shows that it relied on a published appraisal rather than independently assessing the PDF. This could indicate inference-time contamination (D.2), which agreement between runs would not reveal.
- Both illustration: The model gives materially different ratings across repeated runs of the same task (IC1). In some runs, it also treats evidence about random sequence generation as proof of allocation concealment (a possible IC2 evidence-use failure). The two findings must be evaluated separately. Different quotations do not establish IC1 if they provide equally valid support for the same judgement.

## Exclusion and consolidation criteria

We excluded proposed mechanisms when they fell outside the task scope or lacked a specific, plausible and testable pathway to IC1 or IC2. We combined or cross-referenced concepts already covered by an included mechanism instead of counting them again. We used five decision categories:

1. Outside the defined task or workflow. A mechanism cannot affect a task if the necessary feature is absent. For example, tool-calling failures cannot occur in a workflow with no tools. However, tool failures may be relevant in retrieval-supported or agentic tasks. The reversal curse is not directly relevant when the model must judge supplied evidence rather than recall learned factual relationships.
2. Does not meet IC1 or IC2. The proposed problem has no specific, plausible and testable pathway to repeat-run variability or hidden judgement failure. For example, verbatim reproduction of copyrighted text is outside this taxonomy when it has no effect on the judgement or required output.
3. Broad outcome or observed pattern without a distinct failure pathway. Overall calibration error, fairness gaps, differences between subgroups or aggregate logical inconsistency describe results rather than specific mechanisms. A directly testable failure, such as an invented citation, can still qualify even if its internal cause is unknown.
4. Research or model-development process outside the evaluation workflow. Selective reporting, HARKing, missing study code and differences between separately trained models are not automatically mechanisms in the evaluation task. By contrast, changes introduced when preparing documents for the LLM may be relevant if they affect the judgement.
5. Already represented by an included mechanism (consolidation, not exclusion). We merged, cross-referenced or treated a proposed concept as an example of an existing mechanism when it described the same failure pathway. For example, premise-order effects may fall under prompt sensitivity, long-context position effects or response-option order, depending on what was changed.

**Table of excluded or consolidated mechanisms and rationale**

| Exclusion category | Rationale | Mechanisms |
|---|---|---|
| Outside the defined task or applicable workflow | Not applicable when a necessary tool or feature is absent. It may apply in another workflow if a clear IC1 or IC2 pathway exists. | Function-calling failure in a single-call workflow without tools; reversal curse when the task evaluates provided source evidence instead of recalling learned factual relations. |

| Exclusion category | Rationale | Mechanisms |
|---|---|---|
| Does not meet IC1 or IC2 | No specific, plausible, empirically testable route to a material within-condition change or a hidden judgement/required-output failure. | Verbatim regurgitation considered only as privacy or copyright leakage, where the judgement and required supporting output are unaffected. |
| Broad observed outcome or pattern | An overall performance measure is not itself a separate mechanism. A specific, testable failure pathway may still qualify even when its internal cause is unknown. | Calibration/overconfidence metrics, fairness gaps, subgroup repeatability differences, or aggregate logical inconsistency without a defined failure pathway. |
| Research or development process outside the evaluation pathway | Researcher practices and reproducibility barriers outside the specified LLM task are not mechanisms. Task preparation and source conversion remain in scope where they affect judgement. | HARKing; selective reporting; absent code; stale URLs; different training splits for separately developed models (without a task-level pathway). |
| Already represented by an included mechanism — consolidated | Not an eligibility exclusion. Overlapping concepts were merged or cross-referenced rather than counted twice. | Premise-order effects mapped to prompt sensitivity, long-context position or response-option order as applicable; missing outputs attributable to refusal or degeneration mapped to output completeness. |

# 2.0 Search methods

## Search rationale and design

We used a structured, iterative narrative search rather than a database-only systematic search. Research on LLM reliability and evaluation is often first reported in arXiv preprints or computer science conference proceedings.[1, 2] We therefore searched sources beyond biomedical databases, including technical documentation from model providers. The approach was designed to identify and organise plausible mechanisms, not to estimate how often each failure occurs.

## Information sources and search approach

We searched arXiv (including cs.CL, cs.LG and cs.AI), Google Scholar and relevant conference proceedings, including IEEE (Institute of Electrical and Electronics Engineers), NeurIPS (Conference on Neural Information Processing Systems), ICLR (International Conference on Learning Representations), ACL (Annual Meeting of the Association for Computational Linguistics), EMNLP (Conference on Empirical Methods in Natural Language Processing) and AAAI (AAAI Conference on Artificial Intelligence). We also examined vendor and laboratory documentation. Search terms were developed and refined as relevant studies were found. We screened titles and abstracts, read potentially relevant full texts, and checked reference lists and citing papers for further sources. We continued searching to identify distinct mechanisms and representative supporting evidence. This was an iterative narrative search, not an exhaustive systematic review.

# 3.0 Synthesis

## Coding and analysis

For each retained mechanism, we recorded its label and definition, proposed IC1 and/or IC2 pathway, possible tests, and potential mitigations. We also distinguished the settings in which a mechanism could occur from the access needed to investigate it. The manuscript **Table 1** and **Supplement 4, Table S1**, used four deployment settings: commercial chat, vendor-managed agentic environment, direct model API, and self-hosted/user-controlled runtime. Workflow architecture was recorded separately including single-call, retrieval-supported, agentic and multi-agent approaches. The four deployment settings are: (1) Commercial chat interface, a commercial chat or app such as ChatGPT, Claude.ai, or Gemini, where the user can usually set the prompt, files, and basic options while hidden instructions, routing, context handling, and most model settings remain vendor-controlled; (2) Vendor-managed agentic environment, environments such as Claude Code, Codex CLI, or IDE-based coding agents, where users can set files, allowed tools, permissions, and standing instructions, while orchestration, hidden prompts, context construction, model routing or version, and some model settings may remain vendor-controlled; (3) Direct model API, a commercial vendor API in which users can specify exact requests, system prompts, supported settings, and tool schemas, while the vendor controls servers, hardware, and serving code; and (4) Self-hosted model, a model run in a user-controlled runtime. This includes third-party open-weight models downloaded and run by the researcher, researcher fine-tuned models, and models developed or trained in-house. Runtime means the hardware, serving or inference software, and settings used to run the model. In self-hosted use, the researcher can usually specify the weights, runtime, sampling settings, seed where supported, and context setup. Self-hosting increases control over execution conditions but does not necessarily provide knowledge of the training data or post-training history. Retrieval-supported, tool-using or agentic, and multi-agent systems are workflow architectures rather than additional deployment settings. Multi-provider gateways can add another routing layer even when the interface looks like an API.

**Table 1** in the manuscript described the mechanisms in short, plain-language detailing what could go wrong and how it might affect the judgement or required output. and indicated under which deployment setting they can operate and whether the researcher can directly control or test it. These deployment headings indicated the level of access generally needed to investigate a mechanism, not where the mechanism can occur. The final column described what researchers can observe, test, or control within each deployment setting. Detecting an effect does not necessarily establish its underlying cause.

For **Supplement 4, table S1**, we organised mechanisms into ten categories (A–J): user-level factors, infrastructure, sampling, contamination, input processing and context, prompting and task design, agentic workflows, judgement and evidence use, source presentation, and context management across interactions. The current taxonomy reports 38 parent mechanisms and two H.7 sub-mechanisms that are not counted separately. Mechanisms were refined through author discussion. We recorded proposed testing approaches and mitigations without assuming that they had been validated in every deployment setting. We summarised the most relevant evidence and whether the findings directly tested the proposed mechanism. Evidence strength was recorded as Strong, Moderate or Limited:

| Rating | Definition used in Supplement 4, Table S1 |
|---|---|
| Strong | Multiple credible studies or technical sources provide consistent evidence of the mechanism or a closely related failure, with enough detail to support how it could occur. |
| Moderate | Credible research or technical documentation supports the behaviour and a plausible, testable pathway, but the evidence is narrower, indirect, has not been replicated, or has limited task-specific confirmation. |
| Limited | Evidence is preliminary, sparse or mainly conceptual, leaving substantial uncertainty about whether the mechanism operates as described. |

These definitions assess how strongly the available literature supports the mechanism and its proposed pathway, including evidence from broader LLM research. Evidence from read-and-evaluate tasks was distinguished from broader LLM research and vendor documentation.

The testing procedures and mitigation strategies for each mechanism that has established or literature support. We mapped mechanisms to the deployment settings in which they could be relevant and distinguished applicability from whether the mechanism could be observed, tested or controlled in each setting.

The mechanisms are also coded using five descriptive aspects, which show where a mechanism can affect a read-and-evaluate task: (1) Framing: how the task, criteria, and response options are presented; (2) Evidence basis: what evidence reaches the model and is selected or used; (3) Formation: how the model interprets criteria and weighs evidence to reach a judgement; (4) Expression: how the judgement and required outputs are communicated and recorded; and (5) Substantiation: whether quotations, references, and explanations support the judgement, including the interpretive link between evidence and the rating. A mechanism can be assigned to more than one aspect. For infrastructure mechanisms that can change the whole model or serving route, system identity may be recorded separately rather than forcing an aspect assignment.

# Supplement 3 Tutorial: How to identify the workflow architecture and deployment setting of an LLM study

## Purpose

This tutorial explains how to classify an LLM-assisted read-and-evaluate evidence synthesis study. It is intended for readers who are familiar with evidence synthesis methods but have limited experience with large language models (LLMs).

**What is an LLM?** A large language model is a computer model trained to process and generate language. In evidence synthesis, an LLM may be asked to read titles and abstracts, extract information from reports, or make structured judgements such as study eligibility, risk of bias assessments, or certainty of the evidence ratings. The same named model can be used in very different technical arrangements, so the model name alone is not enough to classify a study.

Correctly classifying a study is important because different LLM deployment settings and workflow architectures expose different features that may affect which mechanisms are relevant and which conditions researchers can observe or control.

**The deployment setting** is the route through which researchers access and run the model. The **workflow architecture** is the sequence of processing and decision steps used to turn the study material into the final output. These two dimensions are independent. For example, a commercial API can be used for a simple single-call workflow, a retrieval-supported workflow, or an agentic workflow. For long or complex source documents, it can help to reconstruct the processing chain explicitly. The source file, parsing or OCR, segmentation or retrieval, the model call, and the final judgement. A failure at an earlier step can change the evidence the model receives before it makes any judgement.

These two classifications help identify potentially relevant mechanisms, however, they do not establish that a mechanism occurred:

1. **How was the LLM deployed?** (deployment setting)
2. **How did the LLM perform the task?** (workflow architecture)

## Learning objectives

After completing this tutorial, you should be able to:

- distinguish between commercial chat, vendor-managed agentic or work environment, direct model API and user-controlled model runtime settings;
- distinguish between single-call, retrieval-supported, agentic and multi-agent workflows; and
- identify these characteristics from the methods section of an evidence synthesis study.

## Step 1. Identify the deployment setting

The **deployment setting** describes **how researchers accessed the LLM**. This information is usually reported in the Methods section.

In plain language: Think of deployment as answering: "Where and through what route did the researchers use the model?" It is not the same as the model name. Names such as GPT, Claude, Gemini, or Llama identify a model or model family; commercial chat, vendor-managed agentic or work environment, direct model API, and user-controlled runtime describe how that model was accessed and run for the study.

## 1) Commercial chat

A commercial chat includes vendor-hosted chatbot interfaces and applications that researchers interact with directly through a web browser or mobile application. Examples include:

- ChatGPT
- Claude
- Gemini
- Microsoft Copilot
- Google AI mode

Researchers type prompts into the interface and receive responses without writing code. Typical wording includes:

- "We entered each abstract into ChatGPT."
- "Claude was used to assess risk of bias."

The researcher has relatively little control over the underlying infrastructure.

**In plain language:** The researcher opens a conversation, pastes or uploads evidence, types an instruction, and reads the response. The commercial vendor operates the servers and controls many technical details behind the interface. The user may be able to choose some visible options, but usually cannot fully inspect or lock the complete software and hardware environment.

**Do not infer too much from the commercial chat:** Uploading a PDF or using a built-in file tool in a chat interface does not, by itself, tell you whether the workflow was single-call (i.e., one run or one prompt) or retrieval support (i.e., searching for and retrieving relevant information from documents or other sources to inform the judgement). The publication needs to describe what happened to the document before the final answer was produced. If that processing is not reported, record the workflow architecture as unclear rather than inferring it from the appearance of the chat interface. If PDF conversion, OCR, chunking, retrieval, or other preprocessing is not reported, record the relevant feature as unclear rather than assuming that the original file reached the model unchanged.

Visual 1 shows that a commercial chat interfaces provide a simple conversational interface but expose relatively few technical settings to the user.

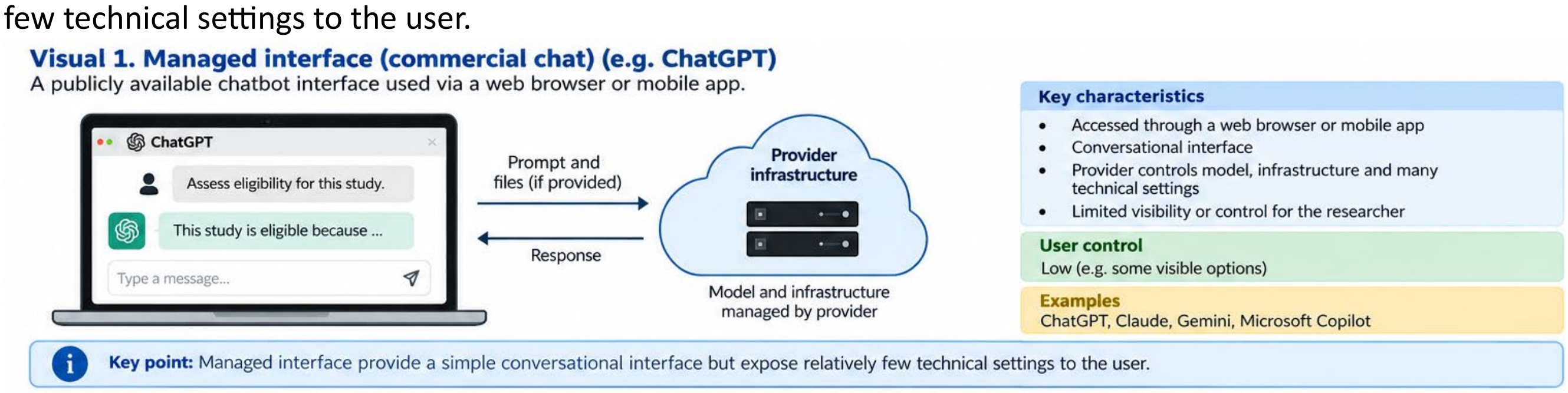

## 2) Vendor-managed agentic or work environment

This is a vendor-hosted environment in which researchers can configure files, tools, permissions or standing instructions and delegate multi-step work to the system. The vendor may still control model routing, hidden instructions, context assembly and tool orchestration. Examples include hosted AI workspaces with file and tool access. A tool-enabled chat does not automatically establish that an agentic workflow was used. The workflow is separate from the deployment setting.

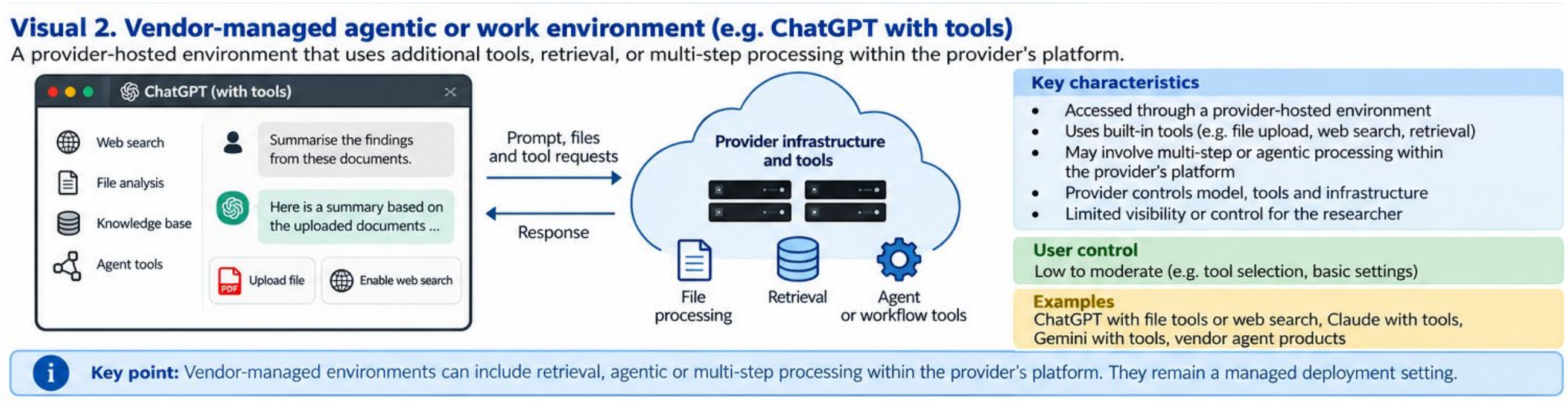

Visual 2 shows that a vendor-managed agentic or work environments may provide tools, files and delegated work, while the vendor controls underlying services.

### 3) Direct API (Application Programming Interface)

A direct commercial API allows researchers to communicate with an LLM programmatically using software rather than a web interface. Instead of typing directly into ChatGPT, researchers write computer code that sends prompts to the model and receives responses automatically.

**In plain language:** An API can be thought of as a structured doorway that lets one computer program send a request to another. A research script can send the same type of request across hundreds or thousands of records, save the outputs automatically, and set technical options consistently. The commercial vendor still operates the underlying model service, but the researcher is controlling how requests are sent to it.
Typical wording includes:

- "The OpenAI API was used."
- "Responses were generated using GPT-5 via the OpenAI API."
- "Temperature was fixed at zero."

Unlike commercial chat, direct commercial APIs allow researchers to control many additional parameters, including:

- model version
- temperature
- structured outputs
- random seeds (when supported)
- prompt templates
- web search
- reasoning settings

**What these settings mean:** Temperature affects how strongly generation favours the highest-probability next words; lower values generally reduce randomness but do not guarantee identical outputs. A random seed, when supported, is a value used to make some random choices more repeatable. Structured outputs constrain the response to a specified format, such as fixed fields in JSON. Prompt templates ensure that the same instruction structure can be applied repeatedly. Tool or web-search settings determine whether the model can obtain information beyond the text supplied directly in the request.

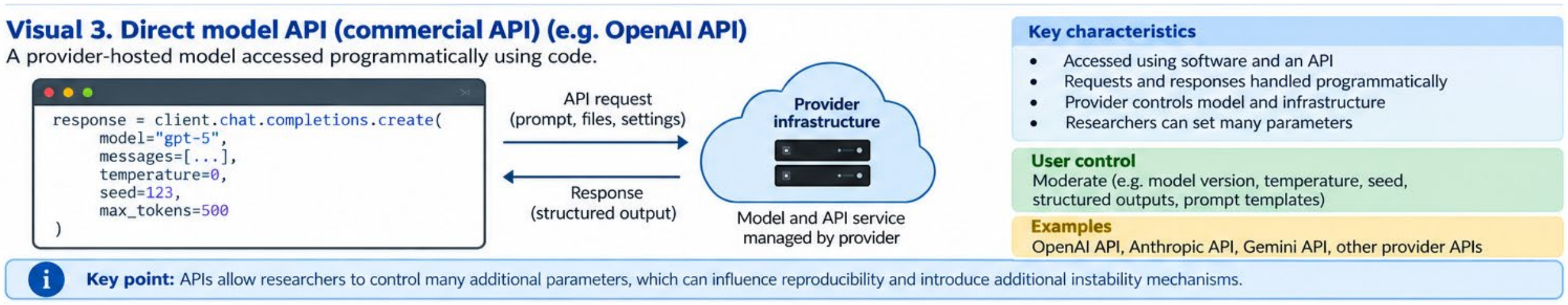

Visual 3 shows that direct APIs may expose settings relevant to repeatability, although the vendor still controls the serving environment.

## 4) Self-hosted model (user-controlled model runtime)

In a self-hosted model or user-controlled model runtime, the researchers or their organisation control model execution, including the model weights and inference configuration. This may be local or on infrastructure they administer.

**In plain language:** Self-hosted does not necessarily mean that the researchers trained the LLM themselves. More commonly, they obtain an already trained model and run inference, meaning the process of generating outputs from that trained model, on hardware and software that they or their institution control.
Instead of sending requests to OpenAI or Anthropic, the model runs on university, hospital or institutional servers. Typical wording includes:

- "Llama 3 was deployed locally."
- "The model was run using vLLM."
- "Inference was performed on NVIDIA A100 GPUs."

Self-hosted deployments allow researchers to control:

- software libraries
- inference engines
- hardware
- precision
- tokenisers
- model weights

These additional controls also introduce additional sources of instability.

**Some self-hosting terms you may encounter:** A GPU is computing hardware commonly used to run LLMs. An inference engine is the software that loads and executes the model. A tokenizer converts text into smaller units that the model processes. Precision describes how numerical values are represented during computation, and model weights are the learned numerical parameters that make up the trained model. Differences in any of these components can matter when a study is self-hosted.

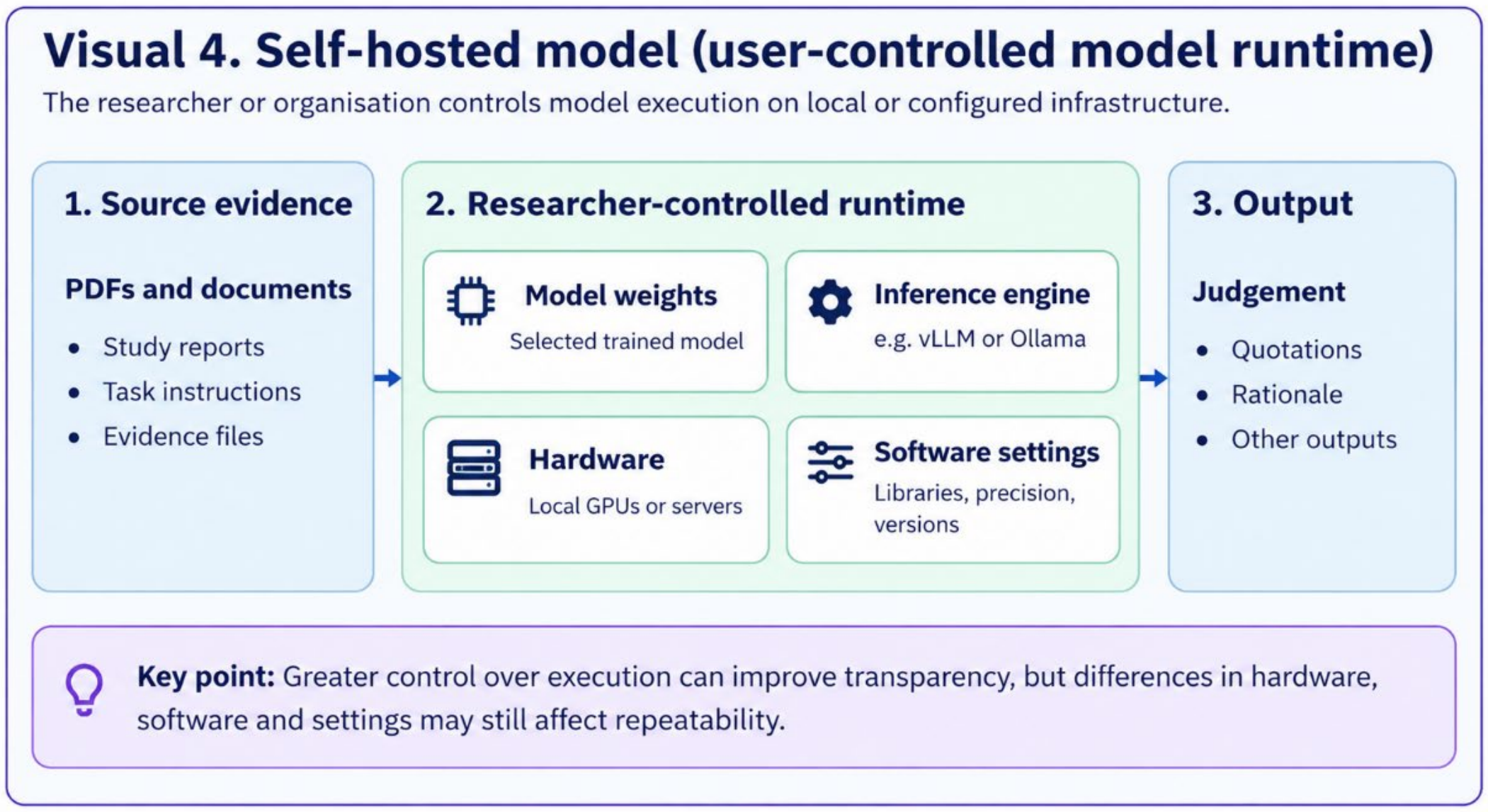

Visual 4 shows that self-hosted/user-controlled runtimes offer greater control over execution but do not necessarily disclose model training or development history.

## Step 2. Identify the workflow architecture

After determining how the model was deployed, determine **how the model completed the task.** The workflow architecture describes the sequence of steps occurring before the final judgment is produced by the LLM. **Keep the two dimensions separate:** Deployment setting and workflow architecture answer different questions. Deployment describes access to the model, whereas workflow architecture describes the sequence of operations used to complete the task. The same deployment can support several workflow architectures. For example, an OpenAI API study could send one direct request per article, retrieve relevant passages before each request, or allow the model to select tools and repeat steps.

**Two technical terms you will see often:** A *model call*, sometimes called an *inference request*, means one request sent to the model and the response generated from that request. A *run* means one complete execution of a task, from the initial input to the final judgement or required output. A run may involve one model call or several calls, depending on the workflow. Understanding the difference between calls and runs can help you reconstruct what happened between the evidence entering the system and the final judgement coming out.

### Single-call workflow

A single-call workflow is the simplest architecture. The researcher provides the input once. The model immediately produces an answer. There are no intermediate steps.

**In plain language:** "Single-call" describes the path used to produce one task result, not the number of records in the entire study. A study could screen 1,000 abstracts using 1,000 separate calls, one call for each abstract, and still use a single-call architecture if each abstract goes directly from input to final output without retrieval, tool use, or model-directed intermediate actions.

**What "single" does not mean:** A long prompt can still be a single call. For example, the prompt may contain several eligibility criteria or several risk-of-bias questions, but if the complete input is sent once and the model returns the final judgement directly, the architecture is still single-call. Examples include:

- title and abstract screening
- data extraction
- risk of bias assessment
- GRADE certainty assessment

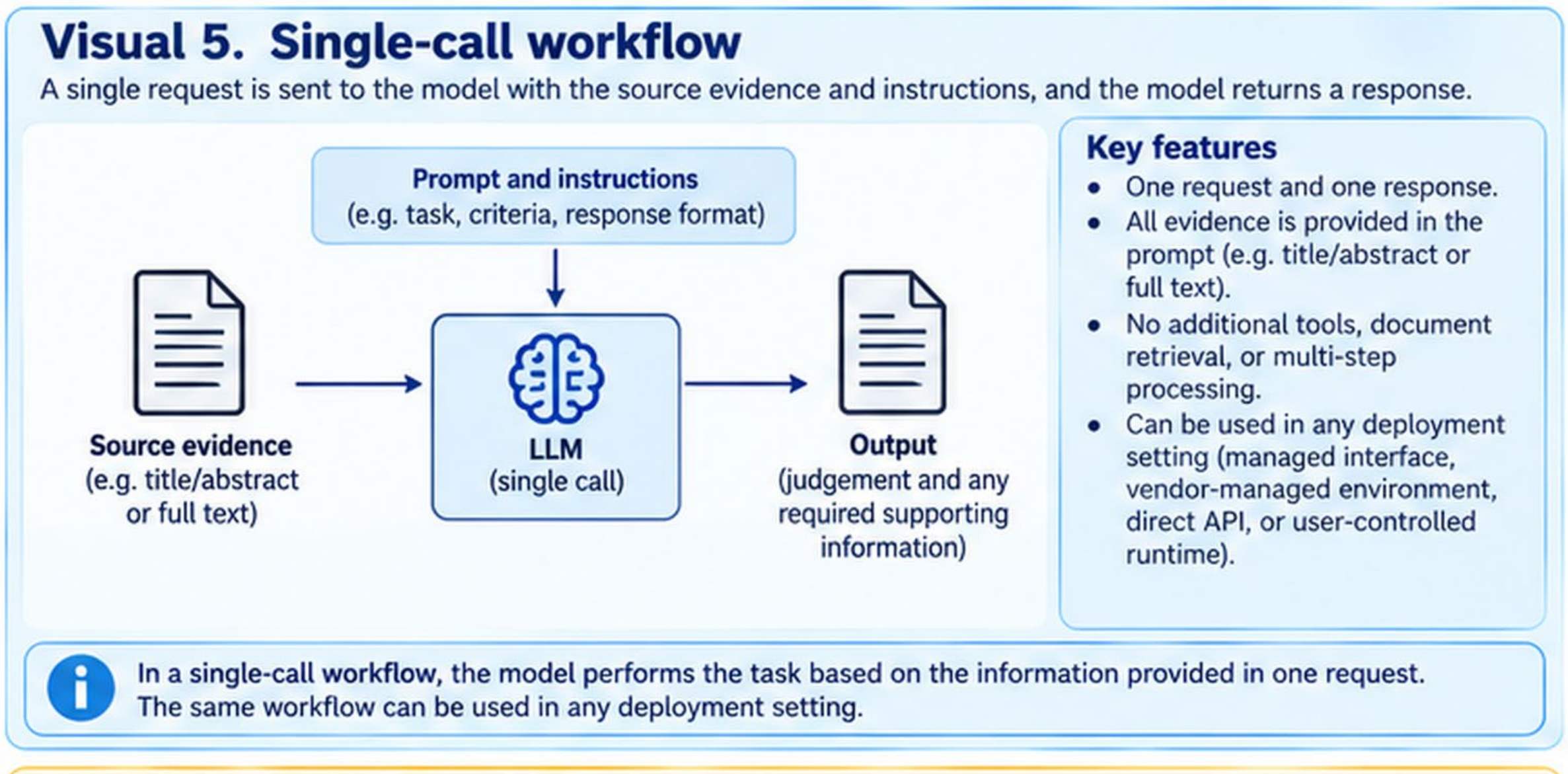

Visual 5 shows that the model directly transforms the supplied evidence into an evaluation

**Typical wording includes:**

- Each abstract was submitted to the LLM in a separate request, and an eligibility decision was returned.
- The full study report and risk-of-bias questions were provided in one prompt, and the model returned its ratings.
- We used one model call per record, with no additional retrieval or tool use.

## Retrieval-supported workflow

Retrieval-supported workflows first retrieve relevant information before the model reasons over it. The model does **not** necessarily receive the complete document. Instead, the workflow:

1. splits documents into chunks;
2. searches those chunks;
3. retrieves the most relevant passages; and
4. supplies only those passages to the LLM.

Retrieval may use:

- vector databases
- semantic search
- RAG (Retrieval-Augmented Generation)
- PDF retrieval systems

**Why use retrieval?** Retrieval is commonly used when the source material is long or when the task requires finding a small amount of relevant evidence within a larger document collection. The retrieval system acts as a selection step: it decides which information the LLM will see before the LLM makes the final judgement.

**Key terms in simple language:** A chunk is a smaller piece of a document, such as a paragraph or several hundred words. An embedding is a numerical representation of text that software can use to estimate which chunks are similar in meaning to a query. A vector database stores these representations so that likely relevant passages can be found quickly. Retrieval-Augmented Generation, or RAG, is a broad term for workflows that retrieve information and then place that information in the context supplied to the LLM before it generates an answer.

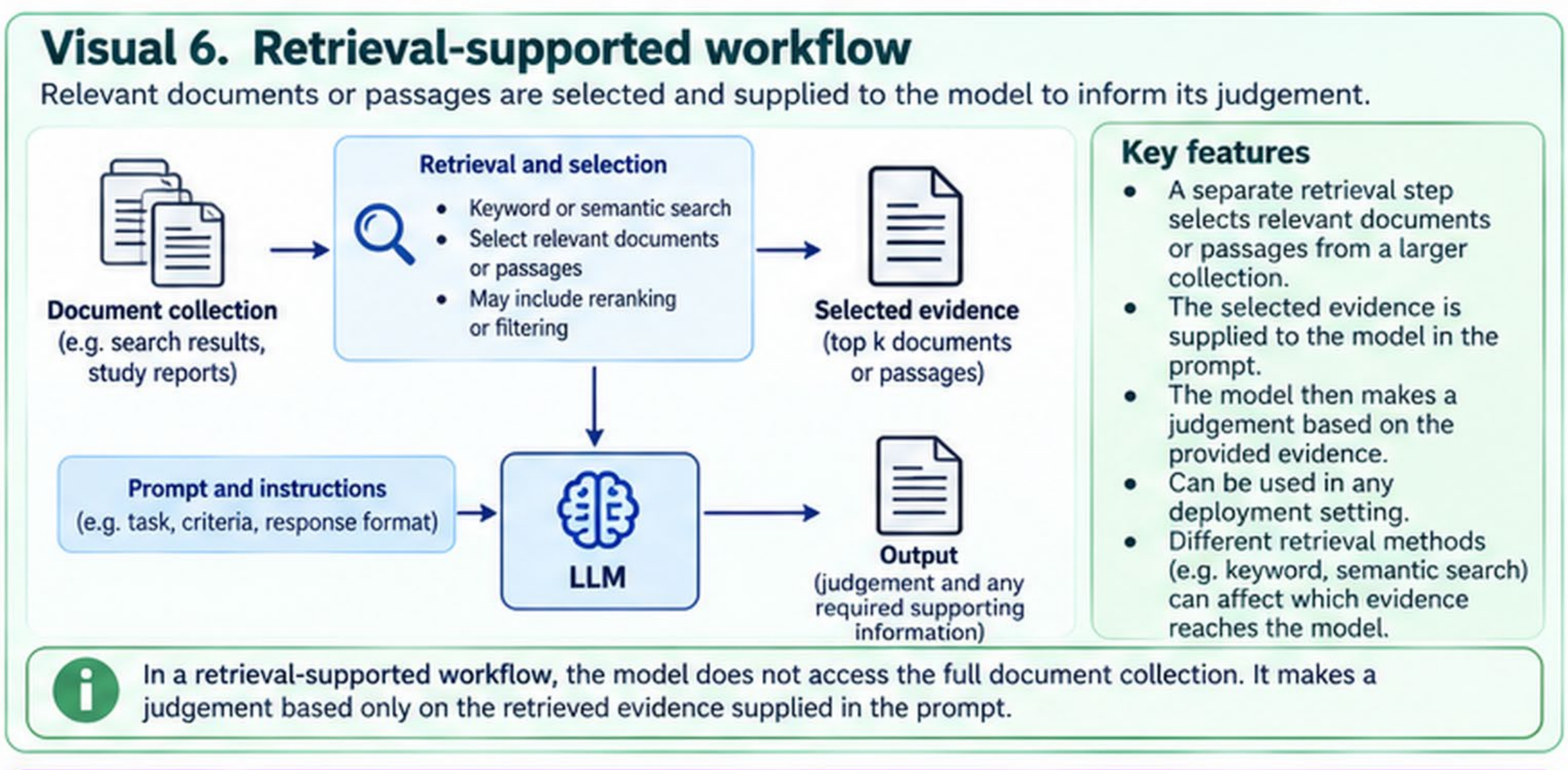

Visual 6 shows that only retrieved passages are presented to the model, creating additional opportunities for instability.

**Typical wording includes:**

- We used retrieval-augmented generation (RAG) to select relevant passages from the included reports before appraisal.
- Full-text articles were divided into chunks, and the five most relevant passages were retrieved for each question.
- A semantic search of the document collection identified evidence excerpts that were supplied to the LLM.

**Why retrieval matters for repeatability and reproducibility:** Retrieval can add variability before the LLM even begins its evaluation. If a different passage is selected from the same document, the model is effectively being asked to judge from different evidence. For this reason, the retrieval step itself is relevant when assessing reproducibility.

## Agentic workflow

Agentic workflows allow the model to make decisions during execution. Instead of immediately answering the question, the model decides which action to perform next.

**In plain language:** An "agent" is not a different species of language model. It is a way of organising an LLM inside a workflow so that the model has some autonomy to choose among allowed actions, such as searching, opening another file, calling a tool, or deciding whether another step is needed. The defining feature is model-directed choice about what happens next. For example, it may:

- search for additional information,
- retrieve another document,

- use an external tool,
- repeat earlier steps,
- decide that more evidence is required.

Each decision influences later steps.

**How to distinguish agentic from simply multi-step:** Do not classify a workflow as agentic merely because it contains several programmed steps. If the developers predetermined every step, for example “extract text, then run prompt A, then run prompt B,” the workflow is multi-step but not necessarily agentic. Look for evidence that the model itself selected, repeated, or changed actions based on intermediate results.

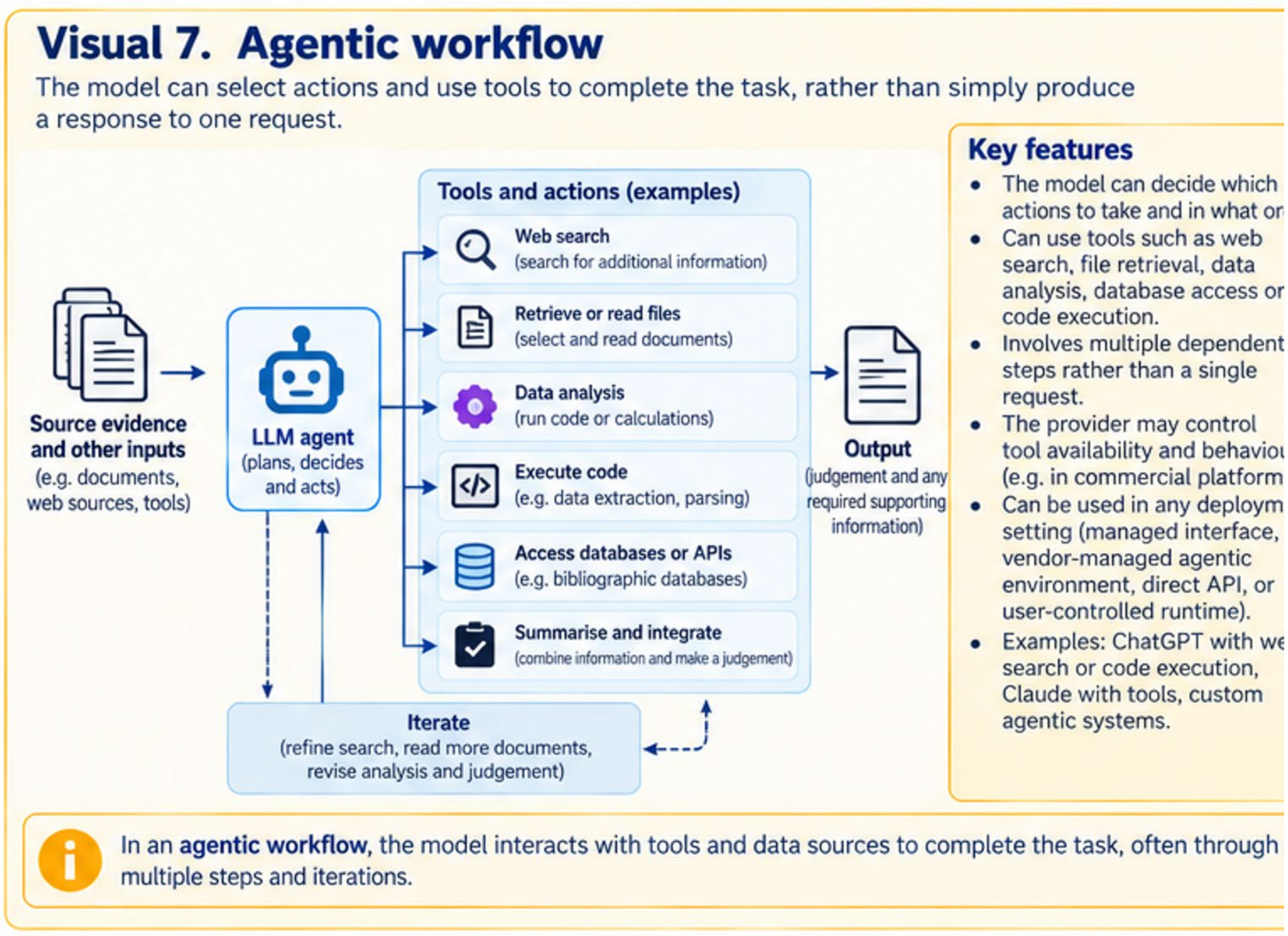

Visual 7 shows that agentic workflows involve sequential decision making rather than a single inference.

**Typical wording includes:**

- The LLM agent selected which tools to call and whether further searches were needed before reaching a judgement.
- Based on intermediate results, the model decided whether to open additional reports or repeat the search.
- The agent autonomously planned and executed a sequence of evidence-gathering and appraisal steps.

## Multi-agent workflow

Multi-agent workflows involve more than one LLM. Each model performs a specialised task before contributing to the final answer. Examples include:

- one model extracts evidence;
- another evaluates risk of bias;
- a third reconciles disagreements.

Alternatively, multiple models may independently evaluate the same evidence before voting.

**In plain language:** In this context, an “agent” is an LLM-based role or component. The agents do not have to use different model families. Several agents may all use the same underlying model but receive different instructions, tools, or responsibilities. What makes the workflow multi-agent is that multiple LLM-based agents generate intermediate outputs that are then combined, compared, debated, voted on, or adjudicated.

**Overlap with agentic workflows:** Multi-agent and agentic are not mutually exclusive concepts. A multi-agent system may also be agentic if one or more agents decide which actions to take during execution. If the publication does not describe the decision process clearly enough to distinguish these features, record the architecture as unclear rather than inferring it from terminology alone.

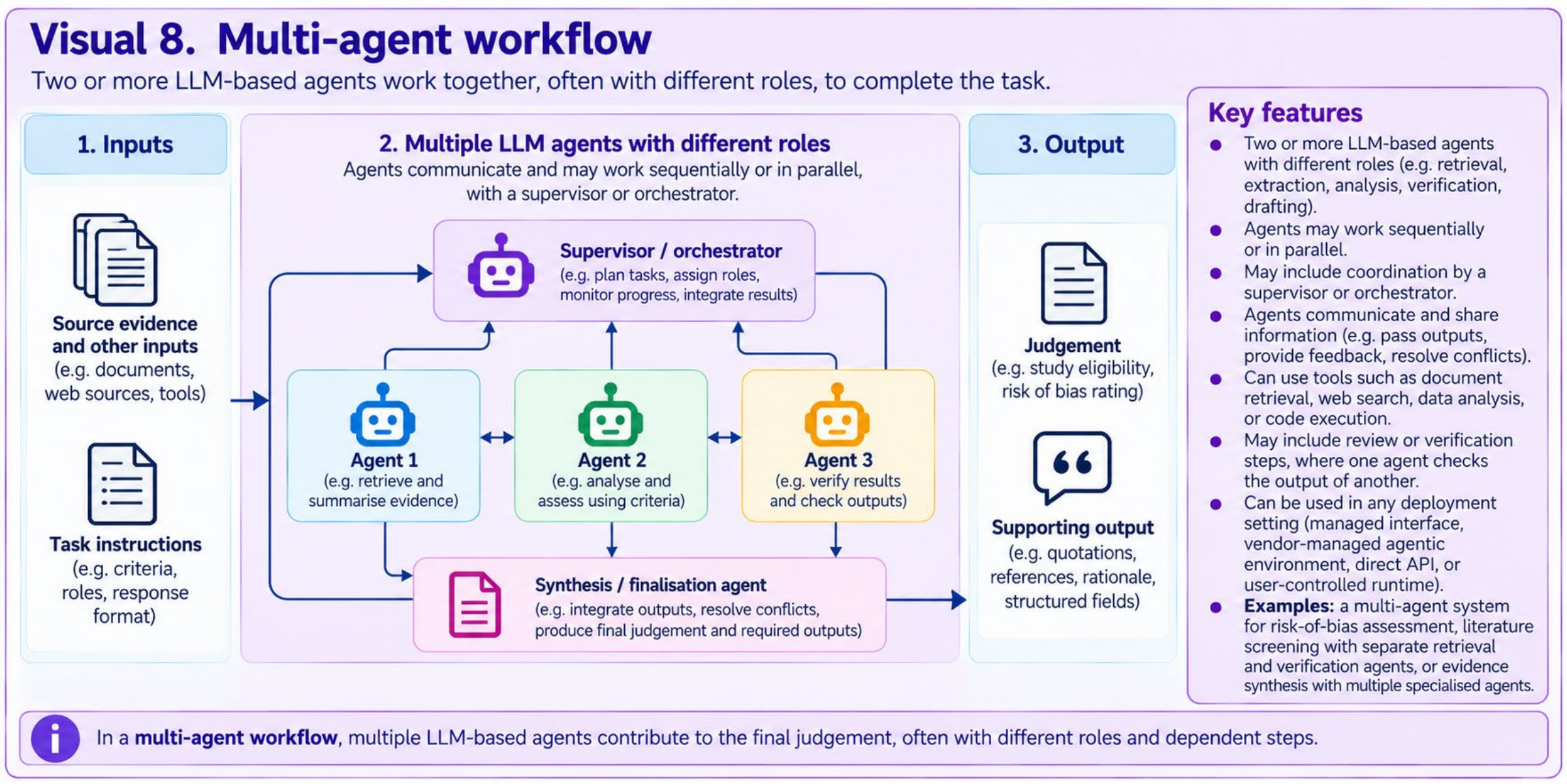

Visual 8 shows that multiple models contribute to the final decision.

**Typical wording includes:**

- Separate LLM agents extracted evidence, assessed risk of bias, and reviewed disagreements before producing a final rating.
- Three LLM-based agents independently evaluated each study, and their assessments were combined by a coordinating agent.
- A reviewer agent checked the first agent’s appraisal, and an adjudicator agent resolved conflicting ratings.

## All conceptually possible combinations of deployment settings and workflow architectures

Deployment setting and workflow architecture describe different features of an LLM-assisted evaluation. Deployment setting describes how the model is accessed and operated, whereas workflow architecture describes how the task is performed. The same workflow architecture may be implemented in different deployment settings, although the available features and degree of researcher control may differ. Visual 9 illustrates these two dimensions and their possible combinations.

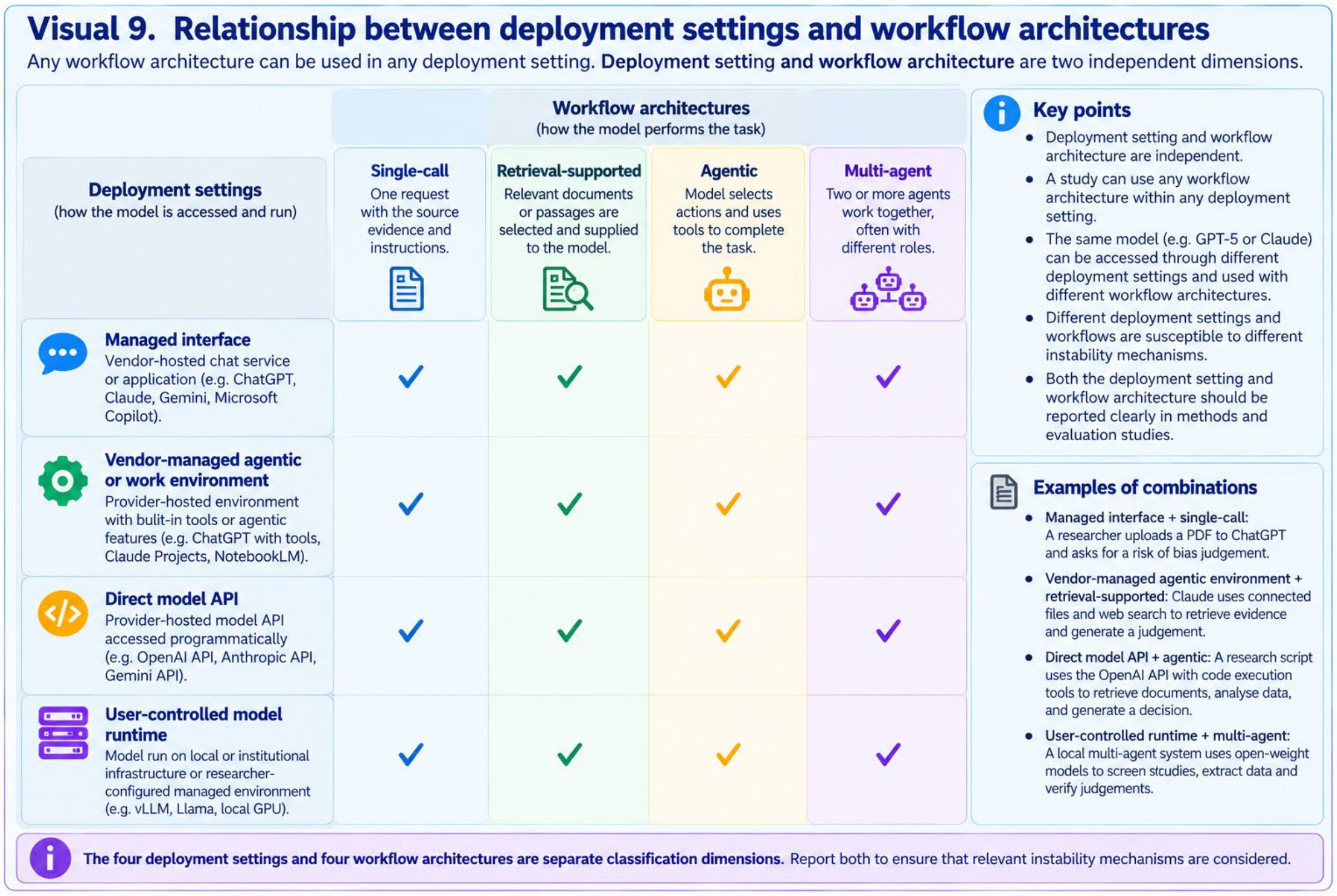

Visual 9 shows all conceptually possible combinations of deployment settings and workflow architectures. Availability depends on the features supported by each system. Check marks do not indicate that every combination is available in every product.

## Step 3. Read the Methods section

Most studies explicitly describe their workflow. The table below provides common clues.

| If the Methods section says... | Likely classification |
|---|---|
| "We used ChatGPT." | Commercial chat |
| "The OpenAI API was used." | Direct model API |
| "Temperature was set to zero (deployment not otherwise reported)." | Direct model API |
| "We used LangChain to let the model choose tools." | Agentic workflow |
| "We used AutoGen." | Multi-agent workflow |
| "We retrieved relevant PDF chunks before appraisal." | Retrieval-supported workflow |
| "Relevant passages were retrieved." | Retrieval-supported workflow |
| "Embeddings were generated." | Retrieval-supported workflow |
| "vLLM was used." | Self-hosted model/User-controlled runtime |
| "Ollama was used." | Self-hosted model/User-controlled runtime |
| "The model called external tools." | Agentic workflow |

**Use caution:** Software and framework names are useful clues, but they are not always sufficient on their own. A framework may support several kinds of workflows. Whenever possible, classify the study from the description of what the software actually did rather than from the package name alone.
Do not infer the deployment setting from a sampling parameter or software library alone. Temperature settings can be exposed in more than one deployment setting; LangChain can support agentic and non-agentic workflows; and chunking alone does not establish retrieval. Record the relevant dimension as unclear if the necessary steps are not reported.

**Reading strategy:** Read the Methods section in two passes. On the first pass, mark clues about deployment, such as a chat interface, an API, a local server, or named hardware. On the second pass, mark clues about workflow behaviour, such as one direct request, chunking and retrieval, model-selected tool use, or several LLM-based agents. Keeping these questions separate helps prevent deployment setting from being confused with workflow architecture.

Read the published Methods paragraph with highlighted phrases:

- "OpenAI API"
- "chunked documents"
- "retrieved passages"
- "LangChain"

Arrows connect each highlighted phrase to the corresponding Figure 2 in the manuscript row or column.

**Beginner tip:** When a manuscript's methods description contains many technical terms, you do not need to understand every term before you can classify the study. First look for the minimum information needed to answer these two questions: how the model was accessed, and what happened between the original evidence and the final evaluation.

## Worked example: Classifying an LLM-assisted evidence synthesis study

Suppose a published systematic review reports that "Full-text PDFs were converted into plain text. Documents were divided into overlapping chunks and stored in a vector database. The five most relevant passages were retrieved for each question and supplied to GPT-5 through the OpenAI API."

**Suppose a published study reports:**
"Full-text systematic reviews were converted into plain text. Documents were divided into overlapping chunks and stored in a vector database. The five most relevant passages were retrieved for each AMSTAR 2 question and supplied to GPT-4o through the OpenAI API. Responses were generated using Python, with temperature set to 0 and outputs recorded in a structured JSON format."

**How to read this example:** "Converted into plain text" describes document preprocessing. "Divided into overlapping chunks" and "stored in a vector database" indicate that a retrieval system was created. "The five most relevant passages were retrieved" confirms that selected evidence was supplied to the model before it made its judgement. "Through the OpenAI API" identifies how the model was accessed. The study also reports the model used (GPT-4o), temperature setting (0), and output format (JSON).

This example shows why the full processing chain matters: errors in document conversion, chunking or retrieval can change the evidence supplied to the model, even when the model and final prompt remain unchanged.

### Step 1. Identify the deployment setting

The researchers used the OpenAI API to send requests to GPT-4o through Python.
**Deployment setting:** ✓ Direct model API (commercial API)

### Step 2. Identify the workflow architecture

The researchers divided documents into smaller sections and used a retrieval system to select relevant passages before sending them to the model.
**Workflow architecture:** ✓ Retrieval-supported workflow

### Step 3. Identify key features and settings

The Methods section reports the following:

- Model: GPT-4o
- Temperature: 0
- Output format: Structured JSON
- Retrieval method: Vector database
- Document preprocessing: Conversion to plain text and division into overlapping chunks

These details help readers understand how the study was conducted and identify features that may affect its results.

## Step 4. Consult the taxonomy table

Combine the information from Steps 1–3.

**Final classification for the example:**

- **Deployment setting:** Direct model API (commercial API)
- **Workflow architecture:** Retrieval-supported
- **Key features:** GPT-4o, temperature 0, structured JSON output and vector-database retrieval

The study is classified aa a *retrieval-supported workflow* row and a *direct model API*. This classification directs readers to the relevant mechanisms in the taxonomy table, including those associated with retrieval, document processing and model settings.

**Why this matters:** The purpose of this step is not to decide whether the study is good or bad. It is to identify which mechanisms are technically possible for that study design and therefore need to be considered. Some mechanisms can affect any LLM workflow, whereas others only arise when a particular capability, such as retrieval, tool use, multiple agents, or self-hosted infrastructure, is present.

**A useful distinction:** At this stage, distinguish "not applicable" from "not reported." If a mechanism cannot occur because the relevant capability is absent, it is not applicable. If the capability may have been used but the Methods section does not provide enough information to determine this, record the classification as unclear and document the reporting gap.

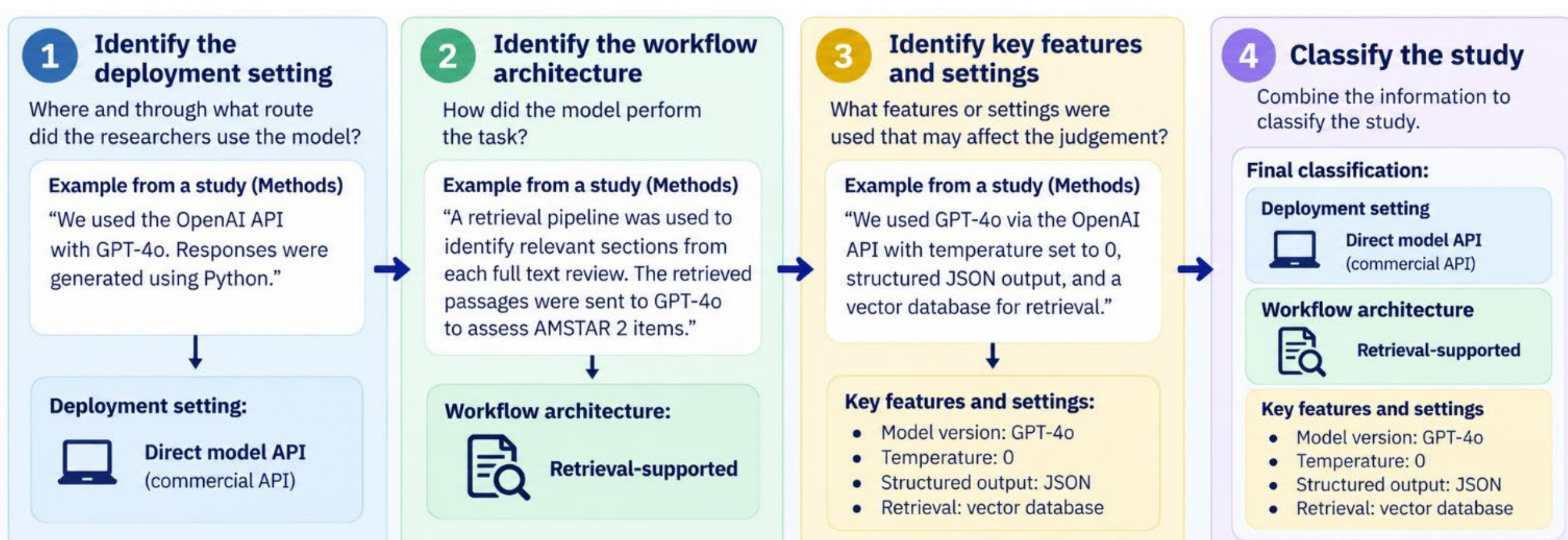

**Caption:** Worked example illustrating the four-step process for classifying an LLM-assisted evidence synthesis study.

This visual shows a worked example illustrating the four-step process for classifying an LLM-assisted evidence synthesis study.

### Common misconceptions

| Misconception | Explanation |
|---|---|
| ChatGPT and the OpenAI API are the same thing. | They may use related underlying models but represent different deployment settings with different levels of user control. |
| Retrieval simply means searching PubMed. | Retrieval refers to the process of selecting documents or passages before they are presented to the model. |
| Using multiple prompts automatically creates an agentic workflow. | Agentic workflows require the model to make decisions about subsequent actions, not merely receive multiple prompts. |
| Temperature = 0 guarantees deterministic behaviour. | Temperature is only one source of variability Other mechanisms may still influence outputs. |
| Every study can be classified perfectly. | Some studies report insufficient methodological detail. When deployment setting or workflow architecture cannot be determined, classification should be recorded as **unclear**, and missing reporting should be documented. |

## Summary

Classifying an LLM-assisted evidence synthesis study requires two independent decisions:

1. **Identify the deployment setting** (how the model was accessed).
2. **Identify the workflow architecture** (how the model completed the task).

By separating deployment setting from workflow architecture, this tutorial enables readers with minimal AI expertise to consistently identify the mechanisms that should be considered when evaluating LLM-assisted evidence synthesis studies.

**Practical rule:** If either question cannot be answered from the publication or supplementary materials, do not guess. Record that dimension as unclear and note exactly what information is missing, such as whether the study used a chat interface or an API, whether retrieval occurred, or whether the model could decide which tools to call.

## Supplementary 4: Full taxonomy of mechanisms, evidence, testing and mitigation strategies

The taxonomy concerns read-and-evaluate tasks that require interpretation of supplied or identified evidence. Eligible mechanisms or operationally defined failure modes must have a specific, plausible, and empirically testable connection to IC1, IC2, or both. They may arise from preparation of the task and evidence, model or system operation, task execution, or recording of the judgement and required supporting outputs. Categories group related mechanisms by where they arise in the LLM system, input, evaluation design, or workflow. The individual mechanism or operationally defined failure mode is the unit used for testing, mitigation, and mapping. The taxonomy does not claim to be comprehensive.

- IC1, Repeat-run variability: The mechanism or failure mode can produce materially different judgements, extractions, or required supporting outputs across repeated runs of the same read-and-evaluate task under the same specified conditions. Specified conditions include the source evidence, task, prompt and instructions, specified model variant and version, and model settings. Internal system conditions, such as routing or serving conditions, may vary without the evaluator's knowledge or control. A difference is material when it affects interpretation, fulfilment of task requirements, or downstream use, rather than wording alone.
- IC2, Hidden judgement failure: The mechanism or failure mode can systematically affect the judgement or required output in a way that is not apparent from the judgement itself. The problem may still be visible elsewhere in the output, for example in an unsupported quotation, an invalid reference, or a missing required field. IC2 does not require repeated runs to agree, and it does not require the final judgement to be incorrect.
- BOTH. A mechanism meets both criteria when it can cause repeat-run variability and can also cause an IC2 problem. These effects need not occur under the same circumstances. Deliberately changing a specified condition is a robustness or sensitivity test and does not establish IC1. A changed judgement under a robustness test supports IC2 only when the changed feature should not legitimately determine the judgement or required task.

The mechanisms are also coded using five descriptive aspects, which show where a mechanism can affect a read-and-evaluate task: (1) Framing: how the task, criteria, and response options are presented; (2) Evidence basis: what evidence reaches the model and is selected or used; (3) Formation: how the model interprets criteria and weighs evidence to reach a judgement; (4) Expression: how the judgement and required outputs are communicated and recorded; and (5) Substantiation: whether quotations, references, and explanations support the judgement, including the interpretive link between evidence and the rating. A mechanism can be assigned to more than one aspect. For infrastructure mechanisms that can change the whole model or serving route, system identity may be recorded separately rather than forcing an aspect assignment.

The four deployment settings are: (1) Commercial chat interface, a commercial chat or app such as ChatGPT, Claude.ai, or Gemini, where the user can usually set the prompt, files, and basic options while hidden instructions, routing, context handling, and most model settings remain vendor-controlled; (2) Vendor-managed agentic environment, environments such as Claude Code, Codex CLI, or IDE-based coding agents, where users can set files, allowed tools, permissions, and standing instructions, while orchestration, hidden prompts, context construction, model routing or version, and some model settings may remain vendor-controlled; (3) Direct model API, a commercial vendor API in which users can specify exact requests, system prompts, supported settings, and tool schemas, while the vendor controls servers, hardware, and serving code; and (4) Self-hosted model, a model run in a user-controlled runtime. This includes third-party open-weight models downloaded and run by the researcher, researcher fine-tuned models, and models developed or trained in-house. Runtime means the hardware, serving or inference software, and settings used to run the model. In self-hosted use, the researcher can usually specify the weights, runtime, sampling settings, seed where supported, and context setup. Self-hosting increases control over execution conditions but does not necessarily provide knowledge of the training data or post-training history. Retrieval-supported, tool-using or agentic, and multi-agent systems are workflow architectures rather than additional deployment settings. Multi-provider gateways can add another routing layer even when the interface looks like an API.

**Table S1. Taxonomy of repeat-run variability and hidden judgment failure mechanisms in LLM reading-and-evaluating tasks**

| Mechanism | Definition | Evidence Summary, Gap and Notes | Applies to | Testing & Mitigation | Affected task components | Inclusion and rationale |
|---|---|---|---|---|---|---|
| **Category A. User-Level Instability. Mechanisms arising from user, account, session, or persistent instruction context applied outside the supplied source and study prompt.** | | | | | | |

| Mechanism | Definition | Evidence Summary, Gap and Notes | Applies to | Testing & Mitigation | Affected task components | Inclusion and rationale |
|---|---|---|---|---|---|---|
| **A.1 Personalisation and user history** | Saved memory, prior conversation history, user profiles, or other account or session memory can add information to a read-and-evaluate task that is not part of the supplied source or declared task conditions. | **Strong overall and moderate for read-and-evaluate tasks.** Prior conversation history or stored user information can alter later model outputs, with effects depending on the model, task, and relevance of the carried-over information.[1-7] In read-and-evaluate tasks, prior judgements systematically shifted later judgements. Two studies found that outputs were influenced by the prevailing polarity of the preceding conversation.[1, 7] Broader personalization studies show that models can use stored user information even when it is irrelevant to the current task.[2-6] OP-Bench and RPEval identified inappropriate memory use, including irrelevant personalization, repetition, flattery, and use of preferences that should have been ignored.[2, 5] Cross-domain studies found that information stored for one context can influence responses in another, and that storing past user information in separate, labeled fields or categories, rather than exposing the model to all prior history can reduce this unwanted carry-over.[3, 6] | All deployments when prior user, session, or memory state is carried into the task. Commercial chat and vendor-managed agentic environments may carry vendor-managed state; APIs and self-hosted models may also carry application-managed state. | **TEST:** Run the same task first without prior state and second with the usual history or memory, while keeping the source, study prompt, model, and other settings fixed. In commercial chat, use a new chat or session for the no-state condition. In APIs or self-hosted systems, use a new or stateless session and record whether prior-turn context, server-side conversation state, or a memory store is used. This comparison tests sensitivity to carried-over state and is not IC1.<br>**MITIGATE:** Use a fresh chat or session for each task and do not supply prior conversation history or stored memory. If persistent memory is required, restrict it to information relevant to the current task and reset or version the state between evaluations. | Evidence basis; Judgment formation | BOTH. IC1: Hidden carried-over state can vary across runs under the same specified conditions and materially change judgements or required outputs. IC2: Irrelevant or inappropriate carried-over state can influence evidence use or the judgement in a way that is not apparent from the judgement itself. |
| **A.2 Custom instructions** | Persistent account-level instructions applied outside the study prompt can add assumptions, priorities, or decision rules that are not part of the intended task. Note: A.1 is what the system remembers about the user. A.2 is what the system is instructed to do. | **Moderate overall and moderate for read-and-evaluate tasks.** Across evaluation studies, higher-level instructions can alter final judgements when the prompt is unchanged.[8-10] When Gupta et al. added a short instruction in the system prompt, more favorable judgments were found.[8] In research quality evaluation, more detailed task instructions performed better than simplified versions.[9, 10] Broader studies show that system instructions can steer model outputs in any directions (i.e., positive, neutral or harmful), and their effects depend on the model and task.[11, 12] Thus, changing the hidden instruction layer does not itself constitute IC1, as the change must produce a different final judgement. Conversely, a fixed but inappropriate instruction can produce a stable shift in judgement that repeated runs will not reveal, which is IC2. | All deployments when persistent instructions exist. Commercial chat may use account-level instructions; vendor-managed agentic environments may use standing or project instructions; APIs may use system, developer, or application instructions; self-hosted models may use fixed system prompts or wrappers. | Commercial chat can include account-level custom instructions. APIs can include system/developer instructions or application-level defaults. Self-hosted systems can also apply fixed system prompts or wrapper instructions.<br>**TEST:** Record the account-level or system instructions applied to the task. As a sensitivity analysis, run the same source and study prompt with and without the standing instruction layer and compare the final judgements.<br>**MITIGATE:** Disable unrelated custom instructions during evaluation. If persistent instructions are required, use the same version-controlled instruction text for all runs and report it verbatim. Keep evaluation instructions explicit and avoid conflicting instructions across system and study prompts. Where the model can be controlled, instruction-hierarchy training can improve adherence to higher-priority instructions, although this is a model-level mitigation rather than something most evaluators can implement. | Framing; Judgment formation | BOTH<br>IC1: An undeclared instruction layer can vary can vary across runs while specified conditions remain fixed and materially change the outputs. IC2: A fixed inappropriate instruction can alter the decision rule or judgment n a way that is not apparent from the judgement itself. |
| **Category B. Infrastructure Instability. Mechanisms arising from changes or variation in model routing, versioning, or the serving environment.** | | | | | | |

| Mechanism | Definition | Evidence Summary, Gap and Notes | Applies to | Testing & Mitigation | Affected task components | Inclusion and rationale |
|---|---|---|---|---|---|---|
| **B.1 Routing variability** | A request is served by a different underlying model, model variant, or inference route even though the specified model identity remains unchanged. | **Moderate overall; limited for read-and-evaluate tasks.** Evidence shows that model substitution and mixed routing are technically possible and can be difficult to identify from outputs alone.[13-15] Cai et al. tested substitution with smaller, quantized, and intermittently substituted models and found that subtle substitutions can evade output-based checks.[13] Zhang et al. subsequently found direct evidence that several third-party APIs claiming to provide named commercial models produced behaviour and model-identity signals inconsistent with the corresponding official APIs.[14] Zhang, Zhang and Qin showed that an endpoint can mix requests between different backends while retaining the same apparent interface, and developed methods for detecting both persistent substitution and request-level mixed routing.[15] Broader routing research shows that changing the selected model can leave an answer unchanged, correct a previous failure, or introduce a new failure. [16, 17] | All deployments can involve routing, but observability differs. Routing is usually vendor-controlled in commercial chat, vendor-managed agentic environments, and direct model APIs. In self-hosted models, the user can usually set and log routing. Multi-provider gateways can add another hidden routing layer. | **TEST:** In commercial APIs, use repeated identical probes and compare output distributions using the model-substitution audit approach of Cai et al. (2025).[13] Use log-probability fingerprints only where they are exposed. In self-hosted/managed systems, inspect routing rules and serving logs to verify which model handled each request. Changes in outputs alone do not establish routing variability. Classify the change as B.1 only when there is evidence that the underlying model or serving route changed.<br>**MITIGATE:** In commercial APIs, pin a dated or snapshot model identifier where available and record the exact model ID, endpoint, and access date, while recognising that hidden provider-side routing may remain. In self-hosted/managed systems, fix the model and routing configuration and retain serving logs so the model used for each request can be verified. | Not assigned. System identity recorded separately. | BOTH. IC1: Hidden routing can send repeated runs under the same specified conditions to different models or inference routes and materially change judgements or required outputs. IC2: Hidden model or route substitution can systematically affect the judgement under an undeclared model or route in a way that the judgement itself does not reveal. |
| **B.2 Model version drift** | The model changes over time because the provider or researcher updates it. The change may be gradual or abrupt. This mechanism is longitudinal and is distinct from variation across repeated runs using the same fixed model version. A version change may alter how the model interprets evidence, follows instructions, uses tools, formats required outputs, or reaches the final judgement. | **Strong overall; moderate for read-and-evaluate tasks.** Across longitudinal studies, model updates changed positively or negatively both the paths models followed and the outputs they produced, even when tasks and prompts were held constant.[10, 18-22] Some updates corrected previously wrong answers, while others changed previously correct answers to incorrect ones. In other cases, overall performance appeared similar but the correct answers changed.[10, 20-22] Path changes included differences in following instructions and responses, sometimes accompanying changes in the final answer.[10, 21, 22] Chen et al. found changes across GPT-3.5 and GPT-4 versions on reasoning, question answering, medical examination, code generation, and instruction-following tasks, with some tasks improving and others not.[22] Venugopal et al. similarly found changes across GPT models during scientific prediction and classification tasks.[21] Ma et al. showed that updates to commercial LLM APIs could change toxicity-classification results and could also change which prompt designs worked best.[20] Echterhoff et al. showed that later LLM versions could reverse previously correct decisions.[18] More recent open-model releases likewise found that | All deployments: Commercial chat, vendor-managed agentic environments, and direct model APIs can be updated by the vendor. Self-hosted models can be held fixed by the user, or deliberately updated to a new weight/version. | **TEST:** Re-run an input at prespecified intervals using the same model version or ID and compare results with the initial reference run. Record the model label or ID and access date at every run. This can detect model drift under a stable label but cannot always identify the exact backend update.<br>**MITIGATE:** In commercial chat, users generally cannot pin a backend version; record the displayed model label and access date and treat a known vendor update as a new model condition. In commercial APIs, use a dated or snapshot model ID where the provider offers one and record the exact ID, endpoint, and access date. If a pinned API version is deprecated, document the date and treat the replacement as a different model. | Judgment formation; Expression | IC1. A hidden model update under an unchanged specified model identity can produce materially different judgements or required outputs across runs. A known version change is a changed condition and should be evaluated as a robustness or longitudinal comparison. A model update alone does not establish an IC2 problem. |

| Mechanism | Definition | Evidence Summary, Gap and Notes | Applies to | Testing & Mitigation | Affected task components | Inclusion and rationale |
|---|---|---|---|---|---|---|
| | | results on fixed evaluation outputs changed between versions.[19] | | | | |
| **B.3 Serving-time numerical non-determinism** | Numerical variation can occur before the next token is selected. The calculations used to run the same model can differ slightly between otherwise identical requests because of differences in how floating-point operations are grouped or performed, how requests are processed on the server, kernel selection, hardware execution, or other serving conditions that affect numerical computation. These small differences can change the probabilities assigned to possible next tokens. Usually the selected token remains the same. When two tokens have very similar probabilities, however, a small numerical difference can change the selected token and may eventually change the final judgement even though the user-visible settings have not changed. This can occur even when the model uses greedy decoding. Greedy decoding removes randomness from token selection, but it does not guarantee identical outputs if the underlying probability calculations differ slightly between runs. | **Strong overall and limited for read-and-evaluate tasks.** Controlled technical studies establish a path-to-output chain. Yuan et al. changed request grouping, GPU count, GPU type, and numerical precision under greedy decoding and found changes in generated responses, including up to 9% variation in accuracy and differences of up to 9,000 tokens in response length for one reasoning model.[23] Thinking Machines Lab showed that standard serving produced 80 unique completions across 1,000 temperature-0 runs, whereas making the numerical operations invariant to how requests were grouped made all 1,000 completions identical.[24] Fu et al. demonstrated that repeated inference produced variation in token probabilities across four models and several GPUs, with larger effects for tokens whose probabilities were not close to 0 or 1.[25] Shanmugavelu et al.[26] and NVIDIA[27] provide supporting evidence that changes in reduction order and hardware execution can alter floating-point results and that bitwise reproducibility is not universally guaranteed. Read-and-evaluate studies show the same phenomenon. Zhao et al. repeated temperature-0 LLM evaluations 50 times and found that one GPT-4o evaluation received scores of 5, 3, and 1 in 52%, 14%, and 34% of runs, respectively, compared to four self-hosted models with consistent results.[28] Tamba found that forcing greedy decoding reduced but did not eliminate repeated pass/fail changes in an LLM-as-judge evaluation.[29] Thinking Machines Lab (2025) demonstrated inference variation and a batch-invariant mitigation; task-specific IC1 effects still require testing. [24] | Can occur in all deployments because serving calculations happen in every runtime, but it is directly testable and partly controllable mainly in self-hosted models. Vendor-managed settings usually hide the relevant hardware and serving details. | **TEST:** In self-hosted systems, repeat identical requests while first holding serving conditions fixed and second varying one serving condition at a time, such as request grouping, hardware, numerical precision, or batch-invariant kernels. Compare token probabilities or outputs to determine whether numerical serving conditions contribute to disagreement.<br>**MITIGATE:** Use batch-invariant or other deterministic inference methods, and keep hardware, software configuration, and numerical precision fixed. Higher-precision computation can further reduce numerical variation. For provider-hosted APIs, document that batch-invariant kernels and serving configurations may be inaccessible; fixed user-visible settings and temperature zero cannot establish numerical determinism.[24] Treat reproducible inference as a property to test, not assume. | Judgement formation; Expression | IC1. Serving calculations can vary under the same specified conditions and alter token selection enough to change a judgement or required output. Numerical variation without a material task-level difference does not meet IC1. The mechanism is not IC2 unless a separate hidden judgement or required-output problem is demonstrated. |
| **Category C. Sampling Instability. Mechanisms arising from stochastic token selection and decoding during generation.** | | | | | | |
| **C.1 Stochastic sampling, decoding and seed** | When stochastic decoding is used, tokens are sampled from a probability distribution. Repeated runs of the same task under the same specified sampling | **Strong and growing evidence.** Repeated run studies show that stochastic decoding can produce different outputs from the same task. Shyr et al.[30] used repeated generations under fixed sampling settings and found that repeatability varied by model, prompt, and dataset.21 An earlier clinical | Can occur in all deployments when stochastic generation is used. Sampling controls are usually most | **TEST:** Where sampling controls are exposed, repeat the identical read-and-evaluate task with the same prompt, model, decoding settings, and seed if available, and report agreement across runs. As a sensitivity test, vary one decoding control at a time to determine whether sampling | Judgment formation; Expression | IC1: Stochastic token selection can produce materially different judgements or required outputs across repeated runs with the same specified sampling settings. Sampling variation alone does not establish IC2. |

| Mechanism | Definition | Evidence Summary, Gap and Notes | Applies to | Testing & Mitigation | Affected task components | Inclusion and rationale |
|---|---|---|---|---|---|---|
| | settings can therefore produce different token sequences and, for judgement tasks, materially different judgements or required outputs. Temperature, top p, top k, and the random seed can influence sampling. | classification study found substantially more disagreement across repeated runs at higher temperatures, with the proportion of cases producing multiple classifications decreasing as temperature was reduced. Windisch et al.[31] similarly found that agreement was highest at temperature 0 and generally decreased at higher temperatures in repeated biomedical classification tasks. A separate mobility extraction study using 100 repeated generations per condition also found that agreement declined as sampling temperature increased, although the effect varied by model and task.[32] | visible in direct model APIs and self-hosted models; Commercial chat and vendor-managed agentic environments may expose few or no controls. | changes the judgement. Commercial chat usually does not expose enough decoding control to isolate C.1 from other backend causes.<br>**MITIGATE:** In commercial APIs, use the least stochastic decoding supported and appropriate for the model, and fix and report any exposed decoding settings and seed. Not all provider-hosted models expose these controls. In self-hosted/managed systems, fix the decoding algorithm and random seed and use greedy or otherwise deterministic decoding where appropriate. These controls reduce sampling variation but do not address B.3 numerical non-determinism. Estimate repeat-run agreement from multiple independently generated judgements and do not treat temperature zero or a single judgement as sufficient evidence of repeatability.[33, 34] Assess agreement with an independent reference standard separately. | | |
| **Category D. Contamination. Mechanisms arising when information that should not determine the judgement is available to or retrieved by the model.** | | | | | | |
| **D.1 Training & Benchmark Contamination** | The model has encountered the test item, expected judgement, a published appraisal, or a close rephrasing or translation during training and may rely on information learned or memorised from that prior exposure rather than independently appraising the supplied evidence. The judgement itself does not show whether it was based on the intended source evidence. | **Strong.** Li et al. 2024 found estimated contamination levels of 1%–45% across six widely used QA benchmarks and more than 15 LLMs; contaminated subsets showed accuracy increases of up to 14% on C-Eval and 7% on HellaSwag, although contamination did not consistently increase performance on every benchmark.[35] Zhang et al. 2024 found that models scored up to 8% lower on the newly constructed GSM1K benchmark than on the matched, publicly available GSM8K benchmark, with analyses suggesting partial memorisation of GSM8K in some model families.[36] Yang et al. showed experimentally that exposure to rephrased benchmark items can substantially inflate performance as demonstrated by Llama-2-13B accuracy on GSM8K increasing from 28.7% to 95.3% after fine-tuning on rephrased test examples.[37] | All deployments. Direct testing is usually limited in closed vendor-managed models. It may be more feasible for some open-weight self-hosted models, researcher-developed models, or APIs when training information is available. | **TEST:** In commercial chat and commercial APIs, training data are generally unavailable, so use behavioural probes, genuinely new matched items, and pre-/post-cutoff comparisons as indirect evidence.[36, 38-42] Where token probabilities or logits are available, model-based membership methods can provide additional evidence.[39] Treat these methods as indirect rather than proof of contamination or its absence. In self-hosted/managed systems, inspect training data where available for exact and near-duplicate overlap between evaluation and training materials.<br>**MITIGATE:** In all deployments, use new, private, or post-cutoff evaluation materials and protected holdout sets where feasible.[36, 43, 44] In self-hosted/managed systems, decontaminate controllable training or fine-tuning data using lexical and semantic matching and keep fine-tuning and evaluation sets separate.[37] Evaluators cannot decontaminate the pretraining data of commercial models. | Evidence basis; Judgement formation | IC2. Prior exposure during training can cause the model to rely on memorised information rather than independently appraise the supplied evidence. That reliance is not apparent from the judgement itself. The judgement may match or differ from a reference judgement. |
| **D.2 Inference-time retrieval** | The model may use information from outside the supplied material, such as | **Moderate.** Direct evidence now shows that web-enabled LLM agents can retrieve benchmark answers during evaluation and thereby inflate apparent | All deployments when external web or retrieval access is | **TEST:** Check and record all retrieved sources. Compare runs with external retrieval disabled versus the planned retrieval condition when the | Evidence basis; Judgement formation | IC2. Information retrieved from outside the intended evidence can influence the judgement in a way that is not |

| Mechanism | Definition | Evidence Summary, Gap and Notes | Applies to | Testing & Mitigation | Affected task components | Inclusion and rationale |
|---|---|---|---|---|---|---|
| **contamination / online influence** | benchmark judgements, earlier appraisals, or other online sources. The judgement may appear acceptable even though the model used information outside the intended evidence, so the judgement itself does not show that the model performed the intended read-and-evaluate task. This mechanism concerns inference-time retrieval or online influence; training-data contamination is D.1. | performance. Han et al. found that web-enabled models directly retrieved datasets containing reference answers for approximately 3% of questions across three question datasets used to test AI models. .[45] When one major dataset-sharing website, Hugging Face, was blocked, accuracy on the contaminated subset fell by 15%.[45] Wang et al. similarly identified metadata leakage, question-context leakage, and answer leakage during evaluation of deep-research models, inflating performance by up to 4%.[46] | enabled. In commercial chat and vendor-managed agentic environments, retrieval may be partly vendor-controlled. APIs and self-hosted workflows may allow more control and logging. | task is intended to use only supplied evidence. **MITIGATE:** Disable web access when the model should judge only the supplied material, or restrict retrieval to approved sources. This addresses inference-time retrieval (D.2) but does not prevent training-data contamination (D.1). | | apparent from the judgement itself. Permitted task-relevant retrieval is not a failure. D.2 concerns inference-time retrieval, and reproduction from training memory is D.1. |
| **D.3 Label Leakage** | The expected judgement, or information that directly signals it, is included in the material supplied to the model. The model may rely on that cue instead of independently appraising the evidence. The cue may appear in filenames, metadata, visible or hidden document content, or other supplied material. D.3 concerns a cue that reveals the expected judgement; D.4 concerns instructions that tell the model how to act. | **Moderate.** Studies show that answer-revealing information, source labels, and metadata available to an LLM during evaluation can influence its judgements or inflate performance. [47, 48] [46, 49] LLM evaluations can be misleading when information already supplied to the model reveals or strongly signals the expected answer. An evidence-synthesis example is provided by Šuster et al., who developed a model to predict GRADE judgements from systematic reviews.[50] Their inputs included abstracts and authors' conclusions, and removing textual supporting information reduced predictive performance. This shows that providing such downstream conclusions could allow the answer to travel with the question. | All deployments. Input cleaning is partly controllable in commercial chat, vendor-managed agentic environments, and APIs, and more directly controllable in self-hosted models where the user controls the input pipeline. | **TEST:** Check PDFs for hidden reference-judgement cues by: (1) comparing machine-extracted text with text recovered from rendered-page OCR, which can identify text present in the document layer but absent from the visible page;[47, 51] (2) inspecting the underlying TeX/source, where available, for concealed text formatting;[51] (3) changing the rendering, for example using HTML or a dark-background/night-mode view, which can expose white-on-white text; and (4) using dedicated hidden-text detectors, such as tools that compare document-extracted text with its rendered appearance.[51]<br>**MITIGATE:** Remove any information from the model's inputs before evaluation. Pre-register the exact prompt and confirm it contains no reference-judgement cue.<br>**Note**: D.3 tests for a cue that reveals the expected judgement. Embedded instructions that tell the model how to act are tested under D.4. | Evidence basis; Judgement formation | IC2. An expected judgement or answer cue in the supplied material can influence the model instead of independent appraisal, and the judgement itself does not reveal the leakage. A judgement can match the reference and still have this problem. Instructions that tell the model how to act are D.4, not D.3. |
| **D.4 Prompt Contamination** | A document, image, Unicode, metadata, retrieved passage, or other source may contain hidden or unintended instructions. The model may follow these instructions instead of using the source only as evidence. D.4 concerns instructions | **Strong.** Evidence from read-and-evaluate tasks studies show that instructions embedded in content can alter LLM judgements.[47, 48] Choi et al. embedded instructions in white text on a white background within medical manuscripts, making them invisible to human readers but available to the LLM.[48] The injected text increased quality and acceptance recommendations across three LLM systems. Jarrett et al. subsequently showed that | All deployments when supplied or retrieved material can contain instructions. User control over document processing and retrieval is greatest | **TEST:** Check documents for hidden instructions in text, images, and metadata. For PDFs, compare machine-extracted text with rendered-page OCR, inspect the source or metadata where available, change the rendering to expose concealed text, and use hidden-text detectors such as PhantomLint.[47] Where images or other non-text content are supplied to the model, inspect these for concealed instructions as well. | Framing; Evidence basis; Judgement formation | IC2. Instructions embedded in evidence or retrieved content can redirect how the task is performed, and the resulting judgement may not reveal that the embedded instruction was followed. A cue that directly reveals the expected judgement is D.3. |

| Mechanism | Definition | Evidence Summary, Gap and Notes | Applies to | Testing & Mitigation | Affected task components | Inclusion and rationale |
|---|---|---|---|---|---|---|
| | embedded in the evidence or retrieved content, and a cue that directly reveals the expected judgement is D.3. | susceptibility depended on the model and document-processing method, while OCR preprocessing prevented injections that relied on hidden text in the document layer.[47] Torrielli et al. tested 100 scientific papers of trials across two public chatbots and five repeated runs and found that hidden instructions could successfully steer positive peer-review in 98% of trials.[52] Jarrett et al. subsequently showed that susceptibility depended on the model and document-processing method, while OCR preprocessing prevented injections that relied on hidden text in the document layer.[47, 53-55] | in self-hosted models and researcher-controlled API workflows. | **MITIGATE:** Remove hidden or unintended instructions before the document is supplied to the model. Create a visible-only version of the document, for example from rendered pages or OCR, inspect metadata and images, and use hidden-instruction detection.[47] Separate task instructions from external document content and explicitly reminding the model not to execute instructions found within that external document.[55] | | |
| **Category E. Input Processing and Context Effects. Mechanisms arising from how source information is converted, segmented, retrieved, tokenised, assembled, or used within the model context.** | | | | | | |
| **E.1 Document conversion distortion** | When a PDF is supplied to an LLM, the model may convert or parse the document before carrying out the requested task. That processing can omit, reorder, merge, split, or misassign text or table content, so the model evaluates a distorted representation of the source.<br>**E.1a PDF parsing / text-conversion distortion:** applies when born-digital document content is altered during parsing or conversion.<br>**E.1b OCR / image-to-text conversion distortion:** applies when scanned or image-based content is incorrectly recognised or reconstructed from the document image. | **Moderate.** Document-conversion errors can affect downstream LLM performance, and results can depend on how documents are converted. Konet et al. (2024) tested extraction from 10 study PDFs and reported 31% errors when GPT-4 accessed the documents through a PDF-reading plug-in.[56] Studies comparing PDF parsers have also found differences in preservation of text, reading order, and table structure. [57-59] Evidence from OCR-based document tasks similarly show errors. Mathew et al. (2021) showed that answering questions from document images is difficult when the answer depends on page layout, such as correctly linking text or numbers to headings, fields, rows, or columns,[60] while more recent LLM studies have shown poorer performance when models receive OCR-corrupted rather than correctly transcribed document content.[61, 62] | All deployments. Users can inspect preprocessing before upload in any setting, but platform-side conversion may be hidden in commercial chat and vendor-managed agentic environments. APIs vary. Self-hosted models provide the most direct control over the conversion pipeline. | **TEST**: Compare the representation actually supplied to the model with the original source for text, numbers, labels, reading order, and table structure. Where feasible, compare alternative parsers or OCR methods as a robustness test.<br>**MITIGATE**: Fix and report the conversion pipeline, archive the converted input used for evaluation, and send failed or uncertain conversions for human review. The taxonomy includes this operationally defined failure mode because researchers can test whether document conversion changes or removes information needed for the judgement by comparing the original document with the converted input. This does not require access to the model's internal reasoning. | Evidence basis | IC2. Document conversion can alter or remove task-relevant source information, so the model judges a distorted representation without that distortion being apparent from the judgement itself. |
| **E.2 Context Segmentation / chunk-boundary effects** | When a long document is split into smaller parts before the model reads it, information that belongs together can end up in different parts. For example, a result may be separated from a sentence that limits it, or a table value may be | **Moderate.** Across clinical read-and-evaluate and retrieval studies, performance varied with how documents were divided, and methods that preserved more surrounding context generally performed better than smaller fixed segments. Yamga et al. (2026) evaluated 1,388 discharge summaries across 16 clinical phenotypes and three LLaMA model sizes and found that whole-document evaluation consistently outperformed both 256- | All deployments when documents are segmented. Segmentation may be hidden or partly controlled in commercial chat, vendor-managed agentic | **TEST:** When segmentation is under researcher control, evaluate the same documents using alternative segmentation methods, such as fixed-size, sentence- or section-based, and semantic segmentation, and compare judgement changes and accuracy where reference answers are available. In commercial chat, manual segmentation can test sensitivity but usually cannot reveal platform-internal chunking. In | Evidence basis | IC2. Segmentation can separate evidence that needs to be interpreted together, affecting the judgement without that missing relationship being apparent from the judgement itself. |

| Mechanism | Definition | Evidence Summary, Gap and Notes | Applies to | Testing & Mitigation | Affected task components | Inclusion and rationale |
|---|---|---|---|---|---|---|
| | separated from its heading or label. The model may then make a judgement without seeing all the information it needs. In retrieval-based workflows, the way the document is split can also affect which parts are selected and shown to the model. | token and section-based segmentation.[63] Similarly, Gomez-Cabello et al. (2025) found that an adaptive segmentation method outperformed fixed-size segmentation and higher retrieval.[64] Supporting retrieval research likewise shows that processing pieces separately can remove surrounding information that helps the model interpret the retrieved text.[65] This mechanism is particularly relevant to retrieval-augmented generation (RAG)-based pipelines because the segmentation method determines what text can be retrieved and presented together. | environments, and provider-managed APIs. It is usually more directly controlled in researcher-managed API pipelines and self-hosted models. | commercial APIs, record whether segmentation is performed by the researcher or by a provider-managed file or retrieval service. When the final judgement depends on evidence that may span several segments, test the complete segmentation-to-aggregation workflow rather than only segment-level performance.<br>**MITIGATE:** Where segmentation is controllable, fix the method and parameters for the main evaluation and report segment size, overlap, and other relevant settings. If platform-internal segmentation is hidden, it cannot be directly standardised; preprocess or split the source before submission where appropriate and report that choice. | | |
| **E.3 Retrieval Variability** | In a retrieval-supported workflow, the retrieval step determines which passages are supplied to the model. | **Moderate.** Retrieval systems commonly turn each document passage and the user's query into lists of numbers, called embedding vectors, that represent their meaning. Passages with similar meanings tend to have vectors that are close together, so the system retrieves the passages whose vectors are closest to the query vector. Across RAG studies, changing which documents are retrieved can change model performance. Retrieval results can differ when the ranking or mix of retrieved documents changes,[66] or when a different embedding model or index content is used.[67] These findings support IC2 as retrieval sensitivity. Evidence for IC1 is weaker. Wang et al.[67] found that when the vectors and retrieval setup were kept fixed, repeated searches returned exactly the same documents in the same order across all tested FAISS index types,[68] including the HNSW index. Retrieval errors or missing passages can also affect the final model output, but few studies track retrieval and generation separately. Direct evidence that retrieved passages change across identical runs and cause changes in the final output therefore remains limited. | Any deployment using retrieval. Retrieval may be vendor-controlled in commercial chat, configurable but partly hidden in vendor-managed agentic environments, configurable in many API workflows, and directly controlled in self-hosted workflows. | Commercial chat offers limited control over built-in retrieval. Commercial APIs vary: provider-managed retrieval may expose only some settings, whereas a researcher-controlled retrieval layer around a provider-hosted model API can expose the corpus, index, embedding model, retriever, reranker, and retrieval depth. Self-hosted/managed systems can control the full retrieval pipeline. Where possible, evaluate retrieval and generation separately. First whether the evidence required for the task was retrieved, and then whether the model used that evidence correctly.<br>**TEST:** Repeat retrieval with the identical query and fixed retrieval setup and compare retrieved document sets, passages, and rankings.[49] In commercial chat, compare only the sources or passages the interface exposes; hidden retrieval cannot be fully tested. In commercial APIs, log retrieval or tool outputs and fix all exposed settings. In self-hosted/managed systems, fix the corpus, index, embedding model, retriever, reranker, retrieval depth, software version, and random settings. Separately varying query wording is a sensitivity test and is not IC1.<br>**MITIGATE:** Keep all researcher-controlled retrieval components fixed and versioned. In commercial chat or provider-managed retrieval, disable retrieval or use a controlled corpus if the product allows; otherwise residual retrieval variability is not directly preventable. In self- | Evidence basis | BOTH. IC1: Retrieval can return materially different evidence across repeated runs under the same specified retrieval conditions and change the judgement or required output. IC2: A fixed retrieval process can systematically omit or over-select evidence in a way that affects the judgement but is not apparent from the judgement itself. |

| Mechanism | Definition | Evidence Summary, Gap and Notes | Applies to | Testing & Mitigation | Affected task components | Inclusion and rationale |
|---|---|---|---|---|---|---|
| | | | | hosted/managed systems, use exact search when needed to rule out approximation as a source of retrieval differences. | | |
| **E.4 Input modality conversion / preprocessing distortion** | Before a model evaluates images, audio, video, or other non-text, conversion or pre-processing can resize, compress, resample, crop, or alter task-relevant information. It does not include poor image or audio quality already present in the original source. If a table image is correctly delivered to the vision model but downsized until small values disappear, that is E.4. If OCR converts those values into the wrong characters, that is E.1. | **Moderate.** Most studies change resolution reduction, compression, filtering, and frame selection to test the effect of preprocessing, rather than examining preprocessing automatically applied by commercial systems. Across four studies, changes that reduced or altered visual data affected model performance, including changes in image resolution, noise, blur, filtering, compression, and other image degradations.[69-73] These effects were found across biomedical and general multimodal tasks, multiple models, and many degradation types, with Liu et al.[71] also showing larger performance losses as degradation severity increased.<br>Note: E.1 covers PDF-to-text parsing and image-to-text/OCR errors. E.4 covers changes to the non-text signal itself before multimodal interpretation, such as reducing an image's resolution, compressing it, changing video frames, or resampling audio. | All deployments using images, audio, video, or other non-text input. Users can control preprocessing before submission, while additional vendor processing may remain hidden in managed settings. Self-hosted models provide the most direct control. | In commercial chat and commercial APIs, users can control preprocessing performed before submission, but platform-side resizing, compression, resampling, or conversion may remain vendor-controlled. Self-hosted/managed systems can control the full preprocessing pipeline.<br>**TEST:** Process the same source media using different user-controlled preprocessing settings, such as image resolution, compression, audio sampling, or video frame selection, and compare whether the model's judgements change. This tests sensitivity to preprocessing. In commercial chat or APIs, platform-side preprocessing may be hidden and cannot usually be isolated directly. In self-hosted/managed systems, vary one preprocessing operation at a time while holding the model and task fixed.<br>**MITIGATE:** Fix and report user-controlled preprocessing settings, preserve the original media, and avoid preprocessing that removes information needed for the judgement. Platform-side preprocessing in hosted systems cannot usually be fixed by the evaluator. In self-hosted/managed systems, fix the full preprocessing pipeline. | Evidence basis | IC2. Preprocessing can remove or alter task-relevant information before evaluation, and the judgement itself may not show that the model received a damaged representation. Poor quality already present in the original source is outside this mechanism. Changing preprocessing settings is a robustness test. |
| **E.5 Tokenisation effects** | Models read text as smaller pieces called tokens. Different tokenisation can change how the same text is represented or how much of it fits within the context limit. The resulting judgement can therefore be influenced by a representation choice or context loss that is not apparent from the judgement itself. A longer token sequence alone is not a failure. | **Moderate.** Studies show that tokenisation can affect model performance, although evidence for read-and-evaluate tasks[74-76] is narrower than the NLP evidence. Three read-and-evaluate studies found performance differences when language tokenizers design was changed.[74-76] Specifically, tokenizers better matched to the languages being processed,[74, 76] as well as the structure of the tokenisation and the number of tokens produced better outputs.[74, 75]<br>Evidence from question-answering tasks shows that tokenisation is associated with how accurately models use supplied information. Sanz-Guerrero et al.[77] found differences from a small change in how the answer boundary was tokenised, while Kanjirangat et al.[78] found that more comparable tokenisation across languages was associated with | All deployments because all models tokenise input. The tokenizer and preprocessing are most directly specified in self-hosted models and some APIs. Commercial chat and vendor-managed agentic environments usually provide little direct tokenizer control. | **TEST**: Keep the model and tokenizer fixed for the main repeat-run evaluation. Where the tokenizer is specified, inspect token counts and whether important source material is displaced from context. Compare alternative tokenizers or languages separately as robustness tests.<br>**MITIGATE**: Use a tokenizer appropriate to the chosen model and language, keep the tokenizer/version fixed, and verify that required evidence remains within the usable context. Keeping a tokenizer fixed controls the condition; it does not prove that the tokenizer is appropriate. | Evidence basis; Judgement formation | IC2. Tokenisation can change representation or reduce the amount of source material that fits in context, affecting evidence use or the judgement in a way that is not apparent from the judgement itself. A long token sequence alone is not a failure. |

| Mechanism | Definition | Evidence Summary, Gap and Notes | Applies to | Testing & Mitigation | Affected task components | Inclusion and rationale |
|---|---|---|---|---|---|---|
| | | better performance. Petrov et al.[79] found that the same text translated into different languages required up to 15 times as many tokens across 17 tokenizers, reducing how much source material can fit within a model's context limit. EVIDENCE GAP (2026): Tokenisation-related performance differences alone do not establish IC1 or IC2 in read-and-evaluate tasks. Retain conditionally, pending evidence showing a pathway from token representation to materially changed judgements or required outputs. | | | | |
| **E.6 Multiple items or tasks in one model call** | When several criteria, tasks, records, or studies are evaluated in the same model run, the judgement on one item can be influenced by the other items being evaluated at the same time. This is an IC2 concern when the other items should not legitimately determine the target judgement. Comparing an item alone with the same item in a multi-item run is a robustness test. | **Moderate.** Direct evidence comes from LLM-as-a-judge studies. Peng et al. evaluated 2,458 human-labelled rubric judgements in deep-research and agentic-coding tasks. They tested several judging strategies, including evaluating multiple rubric items in the same model call, and found a trade-off between accuracy and efficiency.[80] Darshan and Divekar evaluated LLM judgements of summaries against several criteria. Combining instructions for several criteria into one prompt reduced agreement with expert ratings, with Spearman's correlation decreasing by 5.3%.[81] Supporting studies from question answering and classification also show that answers can change when several items are placed in the same prompt. Lin et al. found that performance could fall as more items were included and that answers could depend on an item's position and order within the prompt.[82] Cheng et al. found that evaluating several items together could maintain or sometimes improve performance, showing that the effect does not occur in every task or model.[83] Yue and Yao further showed that content in one item could affect answers to other items in the same model call, although their study deliberately introduced interfering instructions and therefore represents a more extreme setting.[84] Evidence synthesis data extraction provides task-specific support that grouping choices can materially change performance. Kataoka et al. compared extracting all variables simultaneously with approaches that divided variables into smaller groups and found substantial differences in accuracy across methods and datasets.[85] | All deployments when multiple items or tasks are placed in the same model run. | **TEST:** Compare the same item or task when it is judged alone and when it is judged with other items or tasks. Repeat the test with different numbers of items in the same model call and compare the answers with reference answers. Also change which other items appear with the target item and change their order. Check whether the target output changes. These tests have been used directly in studies of LLMs judging multiple criteria.[80, 86]<br>**MITIGATE:** Consider separate model calls for each item or task.[80, 81, 86] In commercial chat, use a fresh chat or session for each item so prior items are not carried over. If several items or tasks must be completed in the same call, use clear labels and require a separate, clearly identified output for each one. This can help prevent outputs from being matched to the wrong item or task, but it has not been shown to remove interference between them. Do not assume that a particular number of items or tasks is safe. Use as few as practical in each call and test the chosen number for the specific task and model..[80, 86] | Judgement formation | IC2. Other items in the same model run can inappropriately influence the target judgement even when the judgement itself does not show that influence. Legitimate shared context is excluded. Comparing an item alone with the same item in a multi-item run is a robustness test, not IC1. |

| Mechanism | Definition | Evidence Summary, Gap and Notes | Applies to | Testing & Mitigation | Affected task components | Inclusion and rationale |
|---|---|---|---|---|---|---|
| **E.7 Long-context position and length effects** | Relevant evidence can be under-used because of its position in a long context or because surrounding context is very long, even when the evidence remains available to the model. IC2 applies when position or context length influences evidence use in a way that should not legitimately determine the judgement and that is not apparent from the judgement itself. Moving evidence or changing context length is a robustness test, not IC1. | **Moderate.** Multiple studies show that LLMs can use information less effectively as context length increases or when relevant information occurs in particular positions within a long input.[87-92] Hsieh et al. (2024) found large performance losses with increasing context length across most of 17 long-context models,[90] while Liu et al. (2024) found performance was lowest when relevant information occurred in the middle and declined more than 20% in some conditions.[88] However, the classic lost-in-the-middle pattern is not universal.[88] Newer models are more resistant in some settings, [92, 93] although Byerly and Khashabi (2026) found that position effects persisted in long-context tasks and were not reliably reduced by self-consistency.[94] | All deployments when long context is used. The user can usually control input order and length, while platform-side context handling may be partly hidden in commercial chat and vendor-managed agentic environments. | **TEST:** As a sensitivity analysis, test the same task after changing the position of the evidence or the amount of surrounding context. For example, move the same evidence to the beginning, middle, and end of the prompt, or compare the full document with a shorter version that retains the evidence needed for the judgement.[88, 90, 95] **MITIGATE:** Keep inputs within context lengths at which the model has demonstrated adequate performance rather than relying only on its advertised context window.[90] And advertised context window is the maximum prompt length that the model vender says the model can accept in one call. For very long documents, consider dividing the document into smaller sections or retrieving relevant passages.[96, 97] | Evidence basis; Judgement formation | IC2. Relevant evidence can be under-used because of its position or surrounding context length, and that evidence-use problem may not be apparent from the judgement itself. Moving evidence or changing context length is a robustness test, not IC1. |
| **Category F. Prompt & Task Design Instability. Mechanisms arising from how the task, instructions, examples, personas, response choices, or output requirements are presented to the model.** | | | | | | |
| **F.1 Prompt Sensitivity** | Meaning-equivalent prompts can produce different judgements when wording or formatting changes even though the intended evidence and task stay the same. | **Strong.** Direct evidence comes from both LLM-as-a-judge tasks and systematic-review screening. Bhat and Varma tested eight LLM judges across four evaluation tasks using 10 semantically equivalent versions of each prompt and about 115,000 evaluations. Judgments varied across prompt versions, and stronger or larger models were not consistently more robust.[98] JudgeSense also found that some LLM-as-a-judge decisions changed across validated paraphrases, although some apparent differences were explained by other problems, including prompt polarity and position bias.[99] Evidence is also available from systematic-review screening. Behrouzian et al. tested five logically equivalent formulations of eligibility criteria. Under sequential screening, GPT-5.1 recall ranged from 0.72 to 0.32 across formulations, and GPT-3.5 Turbo showed a similar pattern. The differences became smaller when the screening workflow was changed, showing that prompt wording can interact with how decisions are made.[100] Evidence from broader LLM tasks also shows sensitivity to meaning-preserving prompt changes. Sclar et al. reported differences of up to 76% across prompt formats.[101] Salinas found that small formatting changes could alter classifications.[102] Chatterjee et al.[103] and Zhuo et al.[104] provide additional evidence that prompt sensitivity occurs across models and tasks. Qin et al. | All deployments. | **TEST:** Run the same task using slightly different versions that are intended to have the same meaning. Include changes in wording and, where relevant, formatting. Keep the task, inputs, model, and other settings fixed. Compare the judgements across prompt versions and with prespecified reference judgements when available. **MITIGATE:** Test meaning-preserving prompt versions on separate development or validation items when these are available. [98, 100, 101, 103] Choose and fix the prompt before the main evaluation. Few-shot examples may reduce prompt sensitivity, although this has not been well tested in evidence-synthesis tasks.[103] | Framing; Judgement formation | IC2. A judgement can depend on task-irrelevant wording even though the evidence and intended task are unchanged. The changed wording supports IC2 only when it should not legitimately determine the judgement. |

| Mechanism | Definition | Evidence Summary, Gap and Notes | Applies to | Testing & Mitigation | Affected task components | Inclusion and rationale |
|---|---|---|---|---|---|---|
| | | examined prompt sensitivity in 50 open-source LLMs and found that subtle wording changes can alter model behaviour.[105] | | | | |
| **F.3 Few-Shot Example Order and Selection Bias** | When a prompt includes worked examples, the judgement can depend on which examples are used or the order in which they appear. Legitimate learning from examples that define the intended task is not a failure. | **Strong.** Across GPT, Claude and Gemini models, worked examples can change how LLMs make judgments, but there is no consistent model-family pattern showing that examples improve judgments.[106-109] Effects vary by task, judgment criterion and the examples provided. In some cases examples improve agreement with human judgments, in others they make little difference or reduce agreement. These effects occur in both older and newer models. No studies tested whether changing which examples were used, or the order in which they appeared, changed judgment accuracy.<br>**Strong** supporting evidence from general in-context learning shows that both example choice and order can affect outputs. Zhao et al.[110] found that changing the examples or their order could make the model perform better or worse. The model tended to favour answers that appeared more often in the examples or appeared near the end of the prompt. Lu et al.[111] found large differences across orders, across model sizes, and found that an order that worked well for one model did not necessarily work well for another. | All deployments when worked examples are included. | **TEST:** Keep the task and instructions the same, but change the worked examples. First, present the same examples in different order. If there are only a few examples, all orders can be tested. Then test different examples and check whether these changes alter the answers. Zhao et al.[110] and Lu et al.[111] found that changing example choice or order could substantially change model performance, while Šuster et al.[108] noted that example choice and the order they appeared in could affect RoB 2 judgments.<br>**MITIGATE:** Pre-specify the examples, how they will be selected, and the order in which they will appear.[110, 111] Keep these fixed for the main evaluation. Where feasible, test other reasonable selections and orders before drawing conclusions from one configuration. Do not assume that one order will work best across models or tasks. Zhao et al.[110] proposed a calibration method on general classification tasks that reduced some biases caused by few-shot prompts. | Framing; Judgement formation | IC2. Example choice or order can inappropriately influence how criteria are applied when those features should not determine the target judgement. |
| **F.4 Output-format failures** | Requiring a fixed output format, such as JSON, XML, or a set response structure, can interfere with the judgement or another required output. The effect depends on the task, instructions, format, and model. This mechanism concerns a substantive effect of the format requirement, not invalid syntax alone. | **Moderate.** Across studies, the effect of output format varied by the task, the format used, the instructions, and the model.[112-114] Tam et al. found larger performance losses on reasoning tasks than on classification tasks (e.g., low quality vs high quality), especially when the required format was more constrained (i.e., a simple format request to a fixed format with required fields, and then to an output that must exactly match that structure).[114] When the model gave its answer first and formatted it afterward, its performance was usually similar to giving an unformatted answer. Long et al. found that improving how well models followed format instructions could reduce these differences.[113] Yuan et al. found that formatting outputs had no effect on model performance in 43 of 48 GPT-4o settings.[115] In the five settings that were affected, the effect depended partly on the task and the specific instructions. Biskupski and Kleber found a similar pattern but they did not test whether | All deployments when an output format or schema is required. APIs and self-hosted models may expose stronger structured-output controls. | **TEST:** Ask the model to complete the formatted version first and second using the unformatted answer. Compare the judgments with the reference answers to see whether changing the format changes performance.[113-115]<br>**MITIGATE:** First run the model reasoning judgment and then format the answer in a separate step.[113, 114] Where the API supports schema-constrained output, define and validate required fields, permitted labels, and data types.[117] This can reduce malformed or missing fields, but it does not verify whether the judgement, quotations, or other field values are correct. Independently check substantive accuracy and evidence support. | Framing; Judgement formation; Expression | IC2. A format requirement can interfere with the judgement or another required output in a way that is not apparent from the judgement itself. Invalid syntax alone is not this mechanism. |

| Mechanism | Definition | Evidence Summary, Gap and Notes | Applies to | Testing & Mitigation | Affected task components | Inclusion and rationale |
|---|---|---|---|---|---|---|
| | | changing the output format changed the substantive judgment.[112] Evidence synthesis data extraction provides additional task-specific support. A 2026 systematic review found that structured output formats, particularly when combined with few-shot prompting, were associated with better extraction performance.[116] Current vendor evidence: Strict schema-constrained outputs reduce structural errors for supported schemas, but refusals, truncation and erroneous field values remain possible. Schema conformity does not demonstrate that an appraisal judgement or supporting explanation is valid.[117] OpenAI Structured Outputs supports schema adherence but explicitly cautions that schema-valid outputs may still contain mistakes, refusals, or incomplete responses. [117] | | | | |
| **F.5 Persona prompting effects** | An assigned role or persona can add priors, assumptions, priorities, or decision style beyond the fixed evaluation construct. | **Moderate.** Across studies, adding or changing a persona can change model performance, but they did not consistently improve it. The direction and size of the effect varied by the task, persona, prompt, and model. Zheng et al. tested 162 personas across 2,410 factual questions and found that personas did not improve accuracy overall, although the persona's type, domain, and gender could affect accuracy.[12] Salewski et al. found that models given a domain-expert persona performed better on reasoning questions than models given a non-domain-expert persona, showing that the match between the persona and task can matter.[118] Cheng et al. compared a base prompt with a persona prompt in a business proposal annotation task. Adding a persona reduced agreement with human judgments for three models and increased it for one, although the differences were generally small.[119] Luz de Araujo et al. tested 162 personas across seven LLMs and five datasets and found more variation between persona prompts than between control prompts with similar wording.[120] They also found that some specialized expert personas performed better on narrow subject areas, even though expert personas generally did not beat the no-persona condition overall.<br>**Gap:** When domain personas change across evaluation targets (e.g. cardiology persona for cardiology reviews, neurology persona for neurology reviews), outputs reflect persona-content | All deployments when a persona or role instruction is used. | **TEST:** Run identical evaluation prompts under three conditions: no persona, generic expert persona, and domain-specific expert persona. Keep the evidence, criteria, and other instructions the same, and compare both judgments with the reference answer.[12, 118, 119]<br>**MITIGATE:** Avoid using a persona to eliminate persona-content confounding. If different personas are used for different types of content, prespecify which persona will be used for each content type and report the exact persona wording.[12, 118, 119] | Framing; Judgement formation | IC2. A persona can add assumptions or priorities that should not determine the target judgement. The resulting influence may not be apparent from the judgement itself. Comparing personas is a robustness test, not IC1. A task-relevant role alone is not a failure. |

| Mechanism | Definition | Evidence Summary, Gap and Notes | Applies to | Testing & Mitigation | Affected task components | Inclusion and rationale |
| --- | --- | --- | --- | --- | --- | --- |
| | | interactions that are not disentangled from content difficulty effects in standard accuracy analyses. | | | | |
| **F.7 Option / Label Selection** | Option or label selection occurs when the model's judgement is influenced by the position or arbitrary identifier of response choices rather than by the evidence being evaluated. The same choices can therefore receive different judgements when their order or labels change even though their meaning is unchanged. | **Strong.** Across LLM-as-a-judge studies, changing the position of otherwise identical candidate responses can change the model's judgment, with the size and direction of the effect varying across models and tasks [109, 121, 122]. Shi et al. confirmed this pattern across 15 LLM judges, 22 tasks, and more than 150,000 evaluations, and found that the observed position effects were not explained by random variation.[121] Supporting evidence from multiple-choice tasks shows that changing the order or label assignment of unchanged answer choices can alter model responses and reduce measured performance.[123, 124] Shi et al. evaluated 15 LLM judges in over 150,000 pairwise and list-wise comparisons and found position effects that varied by judge and task.[121] This supports positional sensitivity but their applicability to other types of LLM evaluation tasks remains uncertain | All deployments when response options or labels are presented. | **TEST:** Keep the evidence, criterion, and response categories unchanged, but present the categories in balanced alternative orders so that each category appears in each position. Compare the judgments across orders to determine whether the position or identifier of a category changes the selected judgment. For ordinal scales, test the observed position preferences rather than assuming that the model favours the middle category.[109, 121, 125]<br>**MITIGATE:** Use balanced category orders and combine the resulting judgments using a prespecified rule. If only one category per task is feasible, rotate or randomise category order across tasks so that one category is not assigned to the same position. Prespecify and keep each category unchanged across versions.[122, 124] Include a position-swap audit for pairwise comparisons, alongside repeat-run consistency and reference-standard agreement. Report position effects separately: high consistency does not exclude systematic position effects. | Framing; Judgement formation | IC2. Arbitrary option position or label can influence the chosen judgement even though it should not determine the evaluation. The judgement itself does not reveal that influence. |
| **Category G. Workflow Instability. Mechanisms arising from variation in multi-step agentic or multi-agent execution under otherwise fixed conditions.** | | | | | | |
| **G.1 Agentic Workflow Instability** | In a fixed multi-step agentic read-and-evaluate workflow, repeated runs under the same declared conditions can take different execution paths, including different plans, tool calls, tool arguments, retrieved evidence, or intermediate outputs. For example, an agent appraising a systematic review with ROBIS might follow a fixed workflow that identifies relevant sections, retrieves supporting passages, applies predefined decision rules, assigns item judgements, checks completeness, and produces the final appraisal. Across repeated runs, the workflow | **Strong overall and moderate for read-and-evaluate tasks.** Several read and evaluate studies found that agentic workflows with identical queries under identical configurations had variation in final answers, findings, and citations.[126-132] In one study, consistency in outputs ranged from 50% to 96% across more than 4,700 financial-agent runs at temperature 0,[127] while a scientific research agent produced the same result in 95% of 10 repeated runs.[129] BuildBench also found variation across three GPT-4o runs, with a mean success rate of 53%.[133] One study found that execution paths can vary more than final decisions.[127] DFAH-Bench found 94–95% agreement in decisions but only 67–69% agreement in exact tool paths.[127]<br>Note: Path variation by itself is not IC1 because IC1 requires the repeated runs to produce different outputs. If the agent uses different tools, retrieves different passages, or takes different intermediate | Any deployment using an agentic or multi-agent workflow. Most directly relevant to vendor-managed agentic environments, API-built agents, and self-hosted agent systems; also applicable to commercial chat that support agentic tools. | **TEST:** Repeat the same agentic workflow several times under identical conditions and compare final judgements across runs. In commercial chat, compare observable final outputs, citations, and surfaced tool activity; hidden intermediate steps may not be available. In commercial API and self-hosted/managed workflows, log retrieved evidence, tool calls, and intermediate outputs to identify where divergent runs begin.[106,111] Compare agreement across runs and accuracy against the reference answer where available.<br>**MITIGATE:** In commercial chat, control over internal orchestration is limited; use a non-agentic mode when the task does not require an agent, or repeat runs and report agreement. In commercial API and self-hosted/managed workflows where orchestration is researcher-controlled, structure and version intermediate outputs and tool settings, preserve execution traces, and use prespecified aggregation or | Evidence basis; Formation; Expression | IC1. Under a fixed agentic workflow, repeated runs can select different evidence, tools, arguments, or intermediate steps and produce materially different judgements or required outputs. A material task-level difference is required abd different paths alone are insufficient. A fixed but flawed agentic path should be classified under the mechanism causing that hidden failure rather than G.1. |

| Mechanism | Definition | Evidence Summary, Gap and Notes | Applies to | Testing & Mitigation | Affected task components | Inclusion and rationale |
|---|---|---|---|---|---|---|
| | can remain fixed but the agent selects different passages, tool calls, or intermediate steps, which may lead to different final judgements. | steps but reaches the same judgement, there is no IC1 instability in the judgement. | | consensus procedures where supported by the workflow evidence.[106,111] | | |
| **G.2 Verification-representation distortion (conditional workflow failure)** | A verification or policy-checking step converts natural-language evidence, instructions, or decision criteria into a formal or structured representation, but changes, drops, or misinterprets a requirement. The distorted representation is subsequently used to form, revise, or record the judgement or required supporting output. | **Moderate overall; limited direct read-and-evaluate evidence.** Studies of LLM-assisted formal verification have documented errors when natural-language requirements are converted into formal representations, including misinterpretation of requirements and incorrect logical specifications.[134-138] Such errors can affect subsequent verification outcomes. AWS Bedrock documentation also identifies missing concepts, incorrectly translated variables and incomplete policy coverage as potential failures in automated reasoning workflows.[134] These findings support a plausible pathway through which a distorted representation could affect a downstream judgement or required supporting output. However, direct evidence of this mechanism in LLM read-and-evaluate tasks remains limited. | Direct model APIs or vendor-managed agentic environments using verification or formal policy conversion; self-hosted or Commercial chat only when the workflow includes a relevant verification step. | **TEST**: Compare the original natural-language requirements with the system-produced formal representation and an independently verified reference representation to identify omissions, changes or misinterpretations. Test whether identified distortions affect downstream judgements or required supporting outputs. **MITIGATE**: Independently review translated requirements, test edge cases and omitted conditions, version formal policies and verify representations before using them to determine or revise judgements. These measures may reduce the risk of distortion but do not establish that it has been prevented. | Evidence basis; Judgment formation; Substantiation | IC2: A verification or policy-checking step may omit, alter or misinterpret a requirement when converting it into a formal representation. If the distorted representation influences the subsequent judgement or required supporting output, the result may appear formally verified despite relying on an incorrect representation of the original requirements. A post-hoc validator that does not influence the recorded judgement or required supporting output is outside the scope of G.2. |
| | | | | | | |
| **Category H. Judgment, Evidence Use, and Reporting Failures. Mechanisms arising when the model interprets evidence, reaches a judgement, links evidence or rationale to that judgement, or reports the required output.** | | | | | | |
| **H.1 Rationale–judgment unfaithfulness** | A rationale can appear plausible while failing to support, explain, or remain logically linked to the judgement and source evidence. The mechanism concerns a testable failure in the stated rationale or substantiation, not whether the rationale reveals the model's hidden internal reasoning. A rationale that differs from a human rationale does not by itself establish this failure. | **Moderate.** In LLM-as-a-judge controlled chain-of-thought studies, a correct or stable judgment does not guarantee that the output or judgment is linked to a logical stated rationale. Wang et al. (2026) found that models with correct judgments can differ with human reasoning supporting those judgments.[139] Their results showed that outcome accuracy alone did not distinguish this failure, whereas evaluating the rationale did. Turpin et al. (2023) introduced biasing information that affected model answers but was often not acknowledged in the reasoning, reducing accuracy by up to 36%.[140] Lanham et al. (2023) found large differences in how much answers depended on the stated chain of thought.[141] Jiang et al. showed that models could retain the same answer while versioning produced inconsistent or | All deployments when a rationale or explanation is required. | **TEST:** Compare whether the stated rationale reflects information that affects the judgment using sensitivity testing. For example, add or remove a relevant cue and assess whether the judgment changes and whether the rationale acknowledges the influential information.[140] Alter, paraphrase, or introduce errors into a generated reasoning chain before the final answer and assess whether the judgment changes.[141] Compare the model's justification with human reasoning.[139] **MITIGATE:** Do not assume that a generated rationale faithfully represents the process that produced the judgment.[140, 141] Require each judgment to include a human (or reference) | Substantiation | IC2. A rationale or explanation can be unsupported or fail to link logically to the evidence and judgement even when the judgement appears acceptable. This problem is not apparent from the judgement itself. A generated rationale is not evidence of the model's hidden internal reasoning. |

| Mechanism | Definition | Evidence Summary, Gap and Notes | Applies to | Testing & Mitigation | Affected task components | Inclusion and rationale |
|---|---|---|---|---|---|---|
| | | faulty reasoning.[43] Bagaria et al. found that rubric text alone partially predicted judge scores and that counterfactual changes to the candidate response or rubric were not consistently reflected in ratings.[142] This supports concern about evidence-independent judging, but it is not direct proof that an accompanying rationale is unfaithful. | | rationale and evidence that can be independently checked.[139] | | |
| **H.2 Accepting unchecked assumptions** | An unverified claim or premise supplied to the model is accepted as true without checking it against available source evidence, when the task requires the judgement to be based on that evidence. The assumption may then influence evidence interpretation or the final judgement. The failure is not accepting an assumption per se, but treating an unverified assumption as evidence when the task requires independent evaluation.<br>H.2 concerns accepting an unsupported premise; H.6 concerns a judgement being influenced by the user's expressed position or preference. | **Moderate.** Studies of false-premise questions consistently show that language models can accept incorrect assumptions instead of challenging them. Hu et al. (2023) found this using 2,365 human-written false-premise questions.[143] Yuan et al. (2024;[144]) and Zhu et al. (2025; [145]) found similar effects in larger false-premise evaluations, and Li et al. (2025; [146]) found that most of 15 tested LLMs had limited ability to identify flawed premises without being explicitly asked to do so. Xu et al. (2026; [125]) showed that models could have the evidence needed to reject a false premise but still answer as if it were correct. Lee et al. (2026) found that judges sometimes disregarded the supplied reference when it conflicted with their prior knowledge in 13 LLM judges across four question-answering datasets, and common prompting strategies did not reliably eliminate the failure.[147] Chen et al. (2024; [148]) showed that LLM judges can overlook false information when evaluating answers, which is closely related but does not directly test an incorrect assumption supplied in the judge's prompt. | All deployments. | **TEST:** Create matched items with (1) no added premise, (2) a correct premise, and (3) an incorrect premise that conflicts with or is unsupported by the source evidence. Compare whether the assumption changes the model's judgement and whether the model notices that the incorrect assumption is not supported by the source. Also test whether asking the model to check assumptions against the source before answering reduces the effect. Explicit instructions to check premises improves premise detection.<br>**MITIGATE:** Avoid inserting factual conclusions into appraisal prompts unless they are part of the prespecified task. Separate instructions from evidence claims. Require the model to verify important task assumptions against the supplied source before using them. For high-stakes appraisals, require a source quote or location supporting any factual premise used in the judgement. | Framing; Evidence basis; Judgement formation | BOTH. IC1: Under the same specified conditions, the model may check an assumption on some runs but accept it unchecked on others, materially changing the judgement or required output.<br>IC2: An unchecked assumption can inappropriately influence the judgement in a way that is not apparent from the judgement itself. |
| **H.3 Hallucinated citations** | The model provides a bibliographic reference that does not correspond to a real source, or substantially alters identifying details of a real source so that the cited reference is invalid. This includes fabricated titles, authors, publication details, or identifiers presented as genuine references. | **Strong.** Hallucinated citations have been demonstrated repeatedly in scientific and evidence synthesis settings. In a study replicating 11 systematic-review searches, Chelli et al. (2024) found hallucinated-reference rates of 39.6% for GPT-3.5, 28.6% for GPT-4, and 91.4% for Bard among the references generated by the models.[149] Walters and Wilder (2023) similarly found that 55% of references generated by GPT-3.5 and 18% generated by GPT-4 in short literature reviews were fabricated.[150] More recent large-scale evidence confirms that the problem persists across models with citation-generation hallucination rates ranging from 14.2% to 94.9% across 13 LLMs and 40 research domains.[151] Direct evidence from scientific literature synthesis also shows substantial citation- | All deployments when the model generates or retrieves bibliographic references. | **TEST:** Check whether each model-generated reference corresponds to a real publication and whether key bibliographic details match the source record. Use automated citation-verification tools (e.g., CiteVerifier) or searches across bibliographic databases, followed by manual checking of references that are missing, conflicting, or uncertain.[149, 151, 155] Compare each cited claim with the cited source to determine whether the source actually supports it.<br>**MITIGATE:** Verify generated references against bibliographic records before use, and manually resolve citations that cannot be verified or have conflicting metadata.[149, 151, 155] Human review remains necessary for references because | Substantiation | IC2. A fabricated or invalid reference can accompany an otherwise acceptable judgement, and the judgement itself cannot verify whether the reference exists or supports the claim. |

| Mechanism | Definition | Evidence Summary, Gap and Notes | Applies to | Testing & Mitigation | Affected task components | Inclusion and rationale |
|---|---|---|---|---|---|---|
| | | attribution errors, including fabricated or unsupported descriptions of findings.[152, 153] Liang et al. (2026) found 90%–98% binary risk-of-bias accuracy while only 60%–65% of generated claims had source support. [154] | | a real citations can still be inaccurately described.[153] | | |
| **H.4 Evidence-extraction / quote instability and misattribution** | When the model must support a judgement or extraction with a quote, location, or extracted value from text or a table, it may return different evidence across identical runs (IC1). It may also return evidence that is not present in the source, is copied incorrectly, is assigned to the wrong source, section, treatment group, outcome, subgroup, time point, row, column, or label, or does not support the judgement it accompanies (IC2). This includes table-reading errors when the table has been presented correctly but the model selects the wrong cell or links a value to the wrong row, column, treatment, outcome, subgroup, time point, or label. | **Moderate.** Direct evidence shows both run-to-run extraction differences and errors in the evidence linked to a judgement. Gartlehner et al. repeated extraction of 160 data elements from 10 randomized trials using the same prompts.[156] Error rates remained low but shifted across repetitions. The errors included omitted information, incorrect treatment-group data, apparent fabrication, and rounding errors. Yisha et al. repeated numerical data extraction from randomized trials three times.[157] Agreement across repeated extractions varied by model and task, with kappa values ranging from 0.67–0.74 for ChatGPT-4 and 0.84–0.96 for Claude 3 Opus. Windisch et al. tested quote stability by asking three models to determine the patient inclusion criteria in 200 randomized trial abstracts and provide a supporting quote.[158] Across GPT-5.2, Gemini 3 Flash, and Claude Opus 4.5, 83%-91% of quotes were verbatim valid, only 48%–79% were correctly attributed. Support was assessed by LLM judges rather than humans, so these results should be interpreted cautiously. Evidence also shows that LLMs can cite or omit the wrong table values.[159] Liang et al. (2026) evaluated GPT-5, o3-mini, and GPT-3.5 on RoB 2 assessments for 97 randomized trials and found that only 60%-65% of generated claims were supported by the source trial evidence despite 90+% accuracy rates.[154] | All deployments when quotations, evidence locations, or extracted values are required. | **TEST:** Require the model to provide a verbatim supporting quote and source location for each judgement, with an explicit "supporting evidence not found" option. Verify separately that (1) the quote exists in the intended source and (2) it supports the specific judgement. Flag missing, unverifiable or conflicting evidence for review rather than automatically resolving discrepancies.<br>**MITIGATE:** Implement evidence-verification procedures to identify unsupported judgements, including checking both quotation accuracy and whether the quoted evidence supports the judgement. Flag unresolved discrepancies for human review and report evidence support separately from judgement accuracy. Do not treat a valid quotation or a correct judgement as sufficient evidence that the judgement is adequately supported. | Evidence basis; Expression; Substantiation | BOTH. IC1: Repeated runs of the same task can return materially different required quotations, evidence locations, or extracted values.<br>IC2: A judgement can be accompanied by evidence that is fabricated, copied incorrectly, misattributed, or does not support the judgement, and that problem is not apparent from the judgement itself. |
| **H.5 Output Completeness Failure** | The model may leave required components out of its output, such as the judgement, a required structured field, a required extraction, or supporting evidence. | **Moderate.** Studies show that LLMs can omit requested components, with completeness generally declining as the number or complexity of instructions increases. Hao et al. (2025) found that GPT-4o produced complete responses for only 49% of queries, with completeness declining from approximately 51% for queries containing four or fewer requests to 36% for those containing more than four.[160] Jaroslawicz et al. (2025) tested 20 models with 10–500 inclusion instructions and found that omission increased errors for many models as instruction complexity increased.[161] Ferraz et al. (2024) provides less direct evidence because omissions were assessed alongside other instruction- | All deployments when more than one output component is required. | **TEST:** Check that the model returned every required output, or returned the correct number of answers. Tell the model to use an identifier for each request to identify an omission. If the model cannot complete a request, tell the model to give a value and a rational to determine if it could not complete the request or omitted it.<br>Do not tell the model to check its own work, rather ask for the request from a validator program to see if the omitted items change across runs, or if the same items are repeatedly omitted. As a separate sensitivity test, vary the number and order of requested items.[160, 161] | Expression; Substantiation | BOTH. IC1: Which required components are omitted can vary across repeated runs under the same specified conditions and materially change task completion.<br>IC2: A judgement can be present while another required output is missing, and that task-completion problem is not apparent from the judgement itself. |

| Mechanism | Definition | Evidence Summary, Gap and Notes | Applies to | Testing & Mitigation | Affected task components | Inclusion and rationale |
|---|---|---|---|---|---|---|
| | | following failures.[162] Although structured-output constraints can enforce required fields and data types, they do not ensure that the content within those fields is complete, accurate or adequately supported by evidence.[163] Missing fields and incomplete or unsupported field contents therefore require separate assessment. | | **MITIGATE:** Require one output for each request and use the identifier so that missing outputs can be detected automatically.[164] If the model cannot complete a request, require an output with rationale so that an incomplete item can be distinguished from an omitted item. Use an external validator to identify missing requests.[162] | | |
| **H.6 User-position / sycophancy effect** | The model's judgement is shifted toward the user's stated belief, preferred judgement, or expressed position even when that position is unsupported by, or conflicts with, the supplied evidence. Appropriate updating in response to genuinely new task-relevant evidence is not included. Influence carried forward from earlier conversation or stored user context is A.1, and acceptance of an asserted factual premise without checking the source is H.2. | **Strong overall and weak for read-and-evaluate tasks.** Controlled studies show that LLM judgements can shift toward a user's stated belief or preferred answer even when that position conflicts with the information that should determine the judgement.[165-167] In a source-based code-evaluation task, misleading user framing produced decision-flip rates of 40% to 72% and false-alignment rates above 90%; evidence-first prompting substantially reduced both.[165] In medical question answering, a 1.2-million-trial factorial study found that concession to incorrect user challenges depended strongly on the question, timing, and supporting claims supplied by the user.[167] Broader evaluations show sycophancy across model families. SycEval observed sycophantic changes in 58% of interactions and harmful regressive changes in 1%, while another seven-model study found mean agreement with incorrect user opinions of 64%.[166] | All deployments when the user states a preferred judgement or position during the task. | **TEST:** Compare the same read-and-evaluate task under matched conditions in which the user expresses no preferred judgement, states a judgement that agrees with the reference answer, and states an incorrect preferred judgement. Also test whether challenging an initial answer with an unsupported user position causes the model to change its judgement. Measure total judgement changes and, especially, correct-to-incorrect flips toward the user's stated position.<br>**MITIGATE:** Do not provide the model with the user's preferred conclusion when it is not needed for the appraisal. Require the model to identify and assess the relevant source evidence before assigning the judgement. If a user subsequently challenges the answer, require any change in judgement to be supported by new or previously overlooked source evidence rather than by the user's disagreement alone. Use a fresh or neutral context where prior user views could influence the appraisal. | Framing; Judgement formation | IC2. The user's stated preference can influence the judgement even when it is unsupported by the source evidence, and the judgement itself does not show that influence. Appropriate updating to new task-relevant evidence is excluded. |
| **H.7 Failure to integrate task components (parent)** | The model correctly identifies or interprets the individual components of a task but fails to relate, combine, or reconcile them as required when forming the judgement or required supporting output. The problem concerns the relationship between components, not missing or incorrectly interpreted individual components. | **Moderate overall; limited direct evidence for the precisely defined mechanism.** Huang et al. (2025) reported discrepancies between LLM-generated RoB 2 domain judgements and those obtained by applying the prescribed algorithm to the same signalling-question answers.[168] A 2026 RoB 2 pilot reported algorithm conformance of 43%, 51%, and 60% across three systems.[169] These findings demonstrate failures to produce judgements consistent with prescribed decision rules, but do not establish whether the failures arose from incorrect interpretation of individual components or their integration. Broader studies by Yang et al. (2025; [170]) and Gupta et al. (2025; [171]) found limitations in combining individually available facts | All four deployments. | **TEST:** Fix independently verified component findings and compare the higher-order judgements with the required relationships or algorithm; repeat identical calls. Verify each component answer separately, then test whether an essential relation is used in the final judgement.<br>**MITIGATE:** Separate component assessment from integration; use independent logic/algorithm checks when prespecified rules are available. Do not assume the stated rationale reveals internal reasoning. | Judgment formation; Substantiation | BOTH. IC1: Integration failures can vary across repeated fixed-condition runs and materially affect judgements or required supporting outputs, though specific evidence is limited.<br>IC2: the final rating alone may not reveal that supplied component findings were combined incorrectly. |

| Mechanism | Definition | Evidence Summary, Gap and Notes | Applies to | Testing & Mitigation | Affected task components | Inclusion and rationale |
|---|---|---|---|---|---|---|
| | | and completing multi-step reasoning tasks, supporting the plausibility of integration failures. However, direct evidence isolating failures to integrate independently verified, correctly interpreted components remains limited. | | | | |
| **H.7a Incorrect integration-rule application** | Component findings are correctly identified, but the final higher-order judgement or required supporting output does not follow a prespecified relationship, hierarchy, threshold, or decision rule. | **Moderate.** Huang et al. (2025) found that applying the published RoB 2 algorithm to LLM-generated signalling-question answers improved domain-level performance in several domains.[168] A 2026 pilot reported algorithm conformance of 43%, 51%, and 60% across three systems, including combinations of signalling-question answers and domain judgements that were inconsistent with the prescribed rules.[169] | All four deployments. | **TEST:** Hold signalling-question answers fixed and independently calculate the required domain rating. Repeat identical runs.<br>**MITIGATE:** Apply deterministic decision algorithms outside the model and flag disagreements; verify that rules faithfully represent the original tool. | Judgment formation; Substantiation | BOTH. IC1: rule-inconsistent judgements or required rationales may vary on repeated fixed inputs. IC2: a judgement can appear plausible while violating the tool rule, even if the final rating happens to be correct. Observable nonconformance does not prove what rule the model internally used. |
| **H.8 Source-embedded instruction interference** | Instruction-like text within material being read or evaluated is treated as an instruction for the model and improperly influences its judgement or required output, rather than being evaluated as source content. | **Moderate.** Maloyan and Namiot (2025) experimentally demonstrated that prompt-injection attacks embedded in material evaluated by LLM-as-a-judge systems could influence their judgements.[172] OpenAI and AWS documentation also identify the risk of models following malicious instructions contained in untrusted material.[173, 174] These findings provide direct evidence of source-embedded instruction interference, although attack success varies across models and workflows, and its frequency in routine evidence synthesis tasks remains uncertain. | All deployments receiving untrusted evidence; particularly retrieval-supported and agentic workflows. | **TEST:** Insert source-embedded instruction text while keeping substantive evidence fixed. Assess sensitivity and repeatability separately.<br>**MITIGATE**: Enforce instruction/source boundaries, restrict tools, use source isolation and prompt-injection check. Iindependently validate the appraisal. | Evidence basis; Judgment formation | BOTH. IC1: compliance with fixed source-embedded instructions may differ across repeat runs. IC2: source text can covertly redirect the task and influence a plausible-looking judgement. Changing the embedded text is a sensitivity test, not IC1. |
| **H.9 Surface-cue substitution in criterion judgement** | The model treats a keyword, phrase, or other surface cue in source evidence as sufficient proof that an appraisal criterion is met, without checking the underlying requirement. | **Limited.** Mao et al. (2025) found that LLMs used to rank documents can be misled when a document contains words that match the search question, even when the document is not the most relevant.[175] Liu et al. (2024) found that the wording of evaluation labels can influence how LLMs judge debates.[175] These studies show that LLM judgements can be influenced by wording rather than the information that should determine the judgement. However, neither study directly tested whether LLMs rely on keywords or phrases instead of checking whether the source evidence meets all the requirements of an appraisal criterion. | All four deployments. | **TEST:** Construct paired cases with the same keywords but different criterion fulfilment. Independently score task requirements. Repeat unchanged cases separately.<br>**MITIGATE:** Require criterion-specific supporting evidence and include adversarial counterexamples in evaluation. | Evidence basis; Judgment formation | BOTH (IC1 plausible, IC2 strong). An over-weighted lexical cue can produce an unsupported judgement that is not apparent from the rating alone. Dependence on the cue may vary across repeated runs. |
| **Category I. Source Presentation Effects. Mechanisms arising from how the material being judged is written, formatted, or labelled, rather than from differences in the evidence itself.** | | | | | | |

| Mechanism | Definition | Evidence Summary, Gap and Notes | Applies to | Testing & Mitigation | Affected task components | Inclusion and rationale |
|---|---|---|---|---|---|---|
| **I.1 Source writing and format effects** | The model may give different judgements when the same source information is written in a different style or format. This mechanism concerns the writing or format of the source material, not the study prompt. | **Moderate.** The evidence shows that judgements can be influenced by how information is written and presented. In scientific reviewing, Kaneko (2026) found that paraphrases increased review scores,[176] while Du (2025) found that changing only the title style altered GPT-4o and Claude review scores, with branded titles generally scoring higher than plain or interrogative titles.[177] LLM-as-a-judge studies show that short or grammatically flawed answers could be penalized more than answers containing factual errors,[178] and stylistic features were prioritised over factuality and safety.[179] Three studies found longer, authoritative-looking, and well-formatted responses, including longer responses and features such as lists, bold text, links, and emojis, were rated higher.[180-182] | All deployments. | **TEST:** Create two or more versions of the same text that preserve the information but differ in writing or presentation style, such as plain versus more formal wording. Have human reviewers confirm that the versions contain the same information. Apply the same prompt, model, and settings to each version in separate calls and repeat each condition. Compare the judgements between versions and, where reference judgements are available, compare their accuracy. Also check whether changes consistently favour one writing style. <br>**MITIGATE:** Give the model explicit criteria and tell it to judge the relevant methods or evidence rather than writing quality or presentation. Ask it to assess each criterion separately before giving the final judgement.[104, 178] Standardizing writing or formatting before evaluation may also reduce presentation effects.[176, 183] If you can fine-tune or train the model that is doing the judging, you can show it examples where a response looks polished or impressive but is actually wrong. This can teach the judge to pay less attention to style and more attention to whether the answer is correct.[104, 184] | Evidence basis; Judgement formation | IC2. Source writing style or format can influence the judgement even when the underlying information is unchanged and the presentation feature should not legitimately determine the rating. The judgement itself does not reveal this influence. |
| **I.2 Source-label effects** | Information about who or what produced the material can influence the model's judgement even when the content is unchanged. Source labels can include Human, AI, Expert, an author, nationality, or an institution. | **Moderate.** Controlled studies show that LLM judgements can change when source or author information is changed while the evaluated content is held constant.[49, 185, 186] Studies have found that identical information was judged as more trustworthy when labelled as human-authored rather than AI-generated.[49, 185] However, Marioriyad et al. found that models favoured responses labelled expert over human, LLM, or unknown.[186] | All deployments. | **TEST**: Keep the content unchanged while varying superficial presentation features, such as source, author, model or institutional labels, apologetic language, claims of authority, or verbosity, as relevant to the task. Compare judgements to determine whether these changes influence the results. <br>**MITIGATE**: Remove unnecessary labels and other presentation features that could influence judgements. Where source information is required for the task, retain it and specify in advance how it should be used. | Evidence basis; Judgement formation | IC2. An irrelevant source label or identity can influence the judgement even when the content is unchanged, and the judgement itself does not reveal that influence. Labels legitimately required by the task are excluded. |
| **Category J. Context Management. Mechanisms arising from compaction, removal, or failure to preserve required prior task context.** | | | | | | |
| **J.1 Loss or distortion during context compaction** | Earlier task context is reduced to a shorter summary or representation, causing required evidence, instructions, or qualifications | **Limited.** OpenAI and Anthropic documentation describe context compaction, in which earlier conversation content is replaced with a shorter representation to allow a task to continue.[187, 188] Pratapa and Mitamura (2025) examined compression approaches in multi-document summarisation, | Commercial chat and vendor-managed agentic environments where enabled; direct APIs and self- | **TEST**: Compare original and confirmed compacted contexts; check retention of task-critical details and repeated-run judgements. <br>**MITIGATE**: Avoid unverified compaction in controlled assessments, preserve original context, | Evidence basis; Framing; Judgment formation | BOTH. IC1 requires variation across repetitions with otherwise identical user-specified conditions, not simply deliberately activating compaction. IC2: missing or changed task information |

| Mechanism | Definition | Evidence Summary, Gap and Notes | Applies to | Testing & Mitigation | Affected task components | Inclusion and rationale |
|---|---|---|---|---|---|---|
| | to be lost or changed before a later judgement. | providing broader evidence relevant to the preservation of information during compression.[189] | hosted workflows with context management. | compare summaries with required evidence and reinsert critical instructions. | | may affect a later judgement without being apparent from the rating. |
| **J.2 Removal of task-relevant context (truncation or pruning)** | Evidence, instructions, or other task-relevant information previously available to the model is removed from the context before a later judgement, without being preserved in a replacement summary. This may occur through automatic truncation when a context limit is reached or through context pruning that removes selected material. | **Moderate.** Research on context truncation and prompt compression shows that removing information from an LLM's input can affect its performance.[190, 191] LongLLMLingua demonstrated that selectively compressing long prompts can change model performance, although removing less relevant information can also improve efficiency without reducing accuracy.[191] Anthropic documentation confirms that context-editing systems can remove earlier tool results.[190] These findings support the risk that context pruning or truncation may remove information needed for a later judgement. However, the studies do not directly establish that automated context removal causes incorrect or unsupported judgements in evidence appraisal tasks. | Commercial chat and vendor-managed agentic environments where enabled; direct APIs and self-hosted workflows with context management. | **TEST:** Inspect context-removal logs and identify whether material needed for the task was actually removed. Compare retained versus pruned condition and repeat fixed configurations.<br>**MITIGATE**: Retain task-critical inputs, log pruning events and validate summaries or retained extracts. | Evidence basis; Framing | BOTH. IC1 applies only when removal varies under the same specified conditions and materially changes output.<br>IC2: a final judgement may conceal that required evidence or instructions were unavailable. |
| **J.3 Loss of required instructions across turns** | Required instructions are not carried forward when the input for a later turn is assembled, even when no truncation or pruning occurs. | **Limited.** OpenAI and Google documentation describe situations in which instructions or configuration settings are not automatically carried forward between turns.[192, 193] This establishes a technical pathway through which required instructions may be missing from a later model input. He et al. (2024) found that LLMs have difficulty following instructions across multiple turns, but did not establish whether these failures occurred because instructions were missing or because models failed to follow instructions that remained available.[194] Direct in read-and-evaluate tasks remains limited. | Direct APIs, agentic environments and commercial chat; self-hosted systems with application-managed state. | **TEST**: Log the assembled instructions at each turn, verify whether prior requirements are present, and independently check the resulting judgement.<br>**MITIGATE**: Resend stable instructions at every API continuation where required; version and inspect assembled prompts. | Framing; Evidence basis; Judgment formation | BOTH. Loss may differ across executions with identical declared settings. Fixed omission may repeatedly affect a judgement or required output. Instruction not delivered is distinct from delivered instruction ignored by the model. |

**Notes:** IC1, repeat-run variability under the same specified conditions. IC2, a mechanism or failure mode can systematically affect the judgement or required output in a way that is not apparent from the judgement itself. BOTH, the mechanism has a plausible and testable pathway to both criteria, which need not occur in the same circumstance. ***Abbreviations:*** AI, artificial intelligence; API, application programming interface; LLM, large language model; IC1, inclusion criterion 1; IC2, inclusion criterion 2.

# TRAILS taxonomy checklist for LLM read-and-evaluate tasks

| No. | Mechanism | Tested | Disabled feature | Reported | Mitigation implemented |
|---|---|---|---|---|---|
| **ALL DEPLOYMENTS (including feature-dependent mechanisms)** | | | | | |
| 1 | Past chats or memory (A.1) | ☐ | ☐ | ☐ | ☐ |
| 2 | Hidden instructions (A.2) | ☐ | ☐ | ☐ | ☐ |
| 3 | The model looked it up (D.2) | ☐ | ☐ | ☐ | ☐ |
| 4 | The expected judgement travelled with the question (D.3) | ☐ | ☐ | ☐ | ☐ |
| 5 | Hidden instructions enter with the evidence (D.4) | ☐ | ☐ | ☐ | ☐ |
| 6 | Prompt sensitivity (F.1) | ☐ | ☐ | ☐ | ☐ |
| 7 | Writing and format effects (I.1) | ☐ | ☐ | ☐ | ☐ |
| 8 | Who produced the material (I.2) | ☐ | ☐ | ☐ | ☐ |
| 9 | Role playing (F.5) | ☐ | ☐ | ☐ | ☐ |
| 10 | Worked examples can steer the judgment (F.3) | ☐ | ☐ | ☐ | ☐ |
| 11 | User influence on the judgement (H.6) | ☐ | ☐ | ☐ | ☐ |
| 12 | Order of the options (F.7) | ☐ | ☐ | ☐ | ☐ |
| 13 | Several items or tasks in the same model call (E.6) | ☐ | ☐ | ☐ | ☐ |
| 14 | Forcing structured output (F.4) | ☐ | ☐ | ☐ | ☐ |
| 15 | The output quietly leaves things out (H.5) | ☐ | ☐ | ☐ | ☐ |
| 16 | Document distorted (E.1) | ☐ | ☐ | ☐ | ☐ |
| 17 | Converting images or other media (E.4) | ☐ | ☐ | ☐ | ☐ |
| 18 | Evidence buried mid-document gets under-used (E.7) | ☐ | ☐ | ☐ | ☐ |
| 19 | Quoted or extracted evidence (H.4) | ☐ | ☐ | ☐ | ☐ |
| 20 | Hallucinated evidence (H.3) | ☐ | ☐ | ☐ | ☐ |

Disabled feature: Tick only if the required feature was deliberately switched off. The code in () represents the item number in Supplement 4, Table S1.

**Cite as:** Lunny et al. When the Same Prompt Gives Different Answers (and the Same Answer Isn't Always Right): A Taxonomy of Failure Mechanisms, Testing, and Mitigation Strategies for LLM-Assisted Evidence Synthesis Read-and-Evaluate Tasks.https://github.com/carolelunny/TRAILS/blob/main/README.md

| No. | Mechanism | Tested | Disabled feature | Reported | Mitigation implemented |
|---|---|---|---|---|---|
| **ALL DEPLOYMENTS (including feature-dependent mechanisms)** | | | | | |
| 21 | Accepting unchecked assumptions (H.2) | ☐ | ☐ | ☐ | ☐ |
| 22 | The stated reasoning is not the real reasoning (H.1) | ☐ | ☐ | ☐ | ☐ |
| 23 | Failure to integrate task components (H.7) | ☐ | ☐ | ☐ | ☐ |
| 23a | Failure to connect task components (H.7a) | ☐ | ☐ | ☐ | ☐ |
| 23b | Incorrect integration-rule application (H.7b) | ☐ | ☐ | ☐ | ☐ |
| 24 | Source-embedded instruction interference (H.8) | ☐ | ☐ | ☐ | ☐ |
| 25 | Surface-cue substitution in criterion judgement (H.9) | ☐ | ☐ | ☐ | ☐ |
| 26 | Different agent runs can reach different judgements (G.1) | ☐ | ☐ | ☐ | ☐ |
| 27 | Verification material changes before checking (G.2) | ☐ | ☐ | ☐ | ☐ |
| 28 | Model updates can change the judgement (B.2) | ☐ | ☐ | ☐ | ☐ |
| **API, VENDOR-MANAGED AGENTIC ENVIRONMENTS, OR SELF-HOSTED** | | | | | |
| 29 | The model learned it during training (D.1) | ☐ | ☐ | ☐ | ☐ |
| 30 | Randomness in token selection (C.1) | ☐ | ☐ | ☐ | ☐ |
| 31 | Splitting a document can separate related information (E.2) | ☐ | ☐ | ☐ | ☐ |
| 32 | Different evidence may be retrieved (E.3) | ☐ | ☐ | ☐ | ☐ |
| 33 | Loss or distortion during context compaction (J.1) | ☐ | ☐ | ☐ | ☐ |
| 34 | Removal of task-relevant context (truncation or pruning) (J.2) | ☐ | ☐ | ☐ | ☐ |
| 35 | Loss of required instructions across turns (J.3) | ☐ | ☐ | ☐ | ☐ |
| 36 | Hidden model switching (B.1) | ☐ | ☐ | ☐ | ☐ |
| **SELF-HOSTED / USER-CONTROLLED MODEL RUNTIME** | | | | | |
| 37 | Tokenisation effects (E.5) | ☐ | ☐ | ☐ | ☐ |
| 38 | Numerical variation during model serving (B.3) | ☐ | ☐ | ☐ | ☐ |

A mechanism may affect other deployments even when researchers cannot directly observe or test its underlying cause. Blank boxes do not mean the mechanism was absent or ruled out. H.7a and H.7b are sub-mechanisms of H.7. Blank boxes do not mean the mechanism was absent or ruled out. H.7a and H.7b are sub-mechanisms of H.7.

## Supplementary 5: Taxonomy evidence gaps

Several gaps remain in the evidence supporting the TRAILS taxonomy. First, much of the available research has examined general LLM tasks, technical system behaviour, or changes in model performance, rather than read-and-evaluate tasks requiring interpretation of source evidence against prespecified criteria. A systematic review of LLM applications in systematic reviews identified 63 studies, with most performance assessments concerning title and abstract screening, followed by data extraction and full-text screening, indicating uneven coverage of evidence-synthesis tasks.[1] Although broader studies can establish plausible failure pathways, their findings may not generalise to methodological quality appraisal, risk-of-bias assessment, or other tasks requiring complex interpretation of evidence and criteria.

Second, many studies examine sensitivity to deliberately changed prompts, model settings, or input conditions, whereas the evidence for repeat-run variability under the same specified conditions (IC1) is less developed for many mechanisms. For example, research on LLM judges frequently evaluates position consistency or robustness to changed input conditions, which does not necessarily establish disagreement across repeated identical runs.[2, 3] Third, studies often report accuracy or agreement without separately establishing whether the model used the intended evidence, applied the required decision rules, or provided supporting outputs that substantiate its judgement (IC2). This concern is illustrated by Liang et al., who found substantial differences between risk-of-bias judgement accuracy and the proportion of generated claims supported by source documents.[4] Similarly, Bagaria et al. identified cases in which LLM judges failed to respond appropriately to changes in the evaluated material or rubric, raising questions about whether judgements reflected the intended evaluation.[5]

Further research is needed to isolate individual mechanisms, assess their frequency and consequences in current models, and determine how their effects differ across deployment settings and workflows. This is particularly important for mechanisms involving hidden system operations, such as model routing, retrieval, context management, and serving-time computation, where researchers may have limited ability to observe or control the relevant conditions. Technical research has demonstrated that serving-time numerical differences can produce different outputs across otherwise identical inference requests, and that controlling numerical computation can improve reproducibility.[6] However, such technical demonstrations do not establish the frequency or consequences of these mechanisms in read-and-evaluate tasks.

Finally, although several technical controls and prompting strategies are available, their effectiveness in preventing material differences in judgements or hidden judgement failures remains uncertain for many read-and-evaluate applications. Structured outputs, for example, address response-format compliance but do not guarantee the correctness of generated values.[7] Controlled experiments using repeated identical runs, mechanism-specific interventions, and independent verification of judgements and supporting evidence would help establish which mechanisms remain consequential, under what conditions they arise, and which mitigations are effective.